\RequirePackage{fix-cm}
\documentclass[smallextended]{svjour3}       
\smartqed  
\usepackage{lineno}

\usepackage{bm}
\usepackage{bbm}
\usepackage{siunitx} 
\usepackage{graphicx} 
\usepackage{amsmath} 
\usepackage[makeroom]{cancel}
\usepackage{mathtools}
\usepackage{algorithm}
\usepackage{booktabs}
\usepackage{amssymb}
\usepackage{url}
\usepackage{rotating}
\usepackage[noend]{algpseudocode}
\usepackage{appendix}
\usepackage{natbib}
\usepackage[export]{adjustbox}
\usepackage{multirow}
\usepackage{pifont}
\usepackage{scrextend}
\usepackage{lineno,hyperref}
\usepackage{caption, subcaption}
\usepackage{mwe}
\usepackage{doi}
\usepackage[labelfont=bf,labelsep=space]{caption}
\usepackage{threeparttable}

\usepackage{xcolor,colortbl}
\modulolinenumbers[5]
\usepackage{commath}

\newcommand{\Da}{D^{a}}
\newcommand{\Di}{D^i}
\newcommand{\Dd}{D^d}

\newcommand{\Dac}{D^{c|i}}

\newcommand{\X}{\mathcal{X}}
\newcommand{\Xmat}{\vec{X}}
\newcommand{\Xmata}{\Xmat^a}

\newcommand{\Xii}{\vec{X}^{i}}

\newcommand{\Xd}{\vec{X}^{d}}

\newcommand{\x}{\vec{x}}
\newcommand{\Xnega}{\Xmat^{\neg a}}

\newcommand{\Ymat}{\vec{Y}}
\newcommand{\Ytwo}{Y^{i|2}}
\newcommand{\Ytwon}{Y^{|2}}

\newcommand{\Y}{\mathcal{Y}}
\newcommand{\Ynega}{Y^{\neg a}}
\newcommand{\Yi}{Y^{i}}
\newcommand{\Yj}{Y_{j}}
\newcommand{\Yij}{Y_{j}^i}
\newcommand{\Ydj}{Y_{j}^d}
\newcommand{\Ymati}{\vec{Y}^{i}}

\newcommand{\Ya}{Y^{a}}
\newcommand{\Ymata}{\Ymat^{a}}
\newcommand{\Yd}{Y^{d}}
\newcommand{\Ymatd}{\vec{Y}^{d}}

\newcommand{\Tij}{\hat{\Theta}_j^i}
\newcommand{\Taj}{\hat{\Theta}_j^a}
\newcommand{\Tdj}{\hat{\Theta}_j^d}
\newcommand{\T}{\hat{\Theta}}
\newcommand{\Td}{\hat{\Theta}^d}
\newcommand{\Ti}{\hat{\Theta}^i}
\newcommand{\Ta}{\hat{\Theta}^a}
\newcommand{\TZ}{\hat{\Theta}^Z}

\newcommand{\pay}{\hat{p}_{y|a}}
\newcommand{\pac}{\hat{p}_{c|a}}
\newcommand{\pdy}{\hat{p}_{y|d}}
\newcommand{\pdc}{\hat{p}_{c|d}}

\newcommand{\hmua}{\hat{\mu}_a}
\newcommand{\hmui}{\hat{\mu}_i}
\newcommand{\hmud}{\hat{\mu}_d}

\newcommand{\hsigi}{\hat{\sigma}_i}
\newcommand{\hsiga}{\hat{\sigma}_a}
\newcommand{\hsigd}{\hat{\sigma}_d}

\newcommand{\M}{\mathcal{M}}

\newcommand{\y}{\vec{y}}

\newcommand{\LN}{L_\mathbb{N}}

\newcommand{\COMP}{\mathcal{C}}

\newcommand{\given}{\mid}

\DeclareMathOperator*{\argmin}{arg\,min}
\DeclareMathOperator*{\argmax}{arg\,max}
\newcommand{\loge}{\log e}

\algnewcommand\algorithmicinput{\textbf{Input:}}
\algnewcommand\INPUT{\item[\algorithmicinput]}
\algnewcommand\algorithmicoutput{\textbf{Output:}}
\algnewcommand\OUTPUT{\item[\algorithmicoutput]}
\newcommand{\ra}[1]{\renewcommand{\arraystretch}{#1}}

\renewcommand{\vec}[1]{\mathbf{#1}}

\newcommand{\cmark}{\ding{51}}%

\newcommand\myeq{\mathrel{\overset{\makebox[0pt]{\mbox{\normalfont\tiny\sffamily i.i.d.}}}{=}}}
\definecolor{Silver}{rgb}{0.85,0.85,0.85}
\definecolor{Gray}{rgb}{0.5,0.5,0.5}

\defcitealias{lavravc2004subgroup}{CN2-SD}
\defcitealias{van2012diverse}{DSSD}
\defcitealias{van2013discovering}{Skylines}
\defcitealias{bosc2018anytime}{MCTS4DM}
\defcitealias{lijffijt2018subjectively}{SISD}
\defcitealias{belfodil2019fssd}{FSSD}

\makeatletter
\newcommand{\orcid}[1]{%
	\begingroup
	\setbox0=\hbox{\href{https://orcid.org/#1}{\includegraphics[height=\f@size pt]{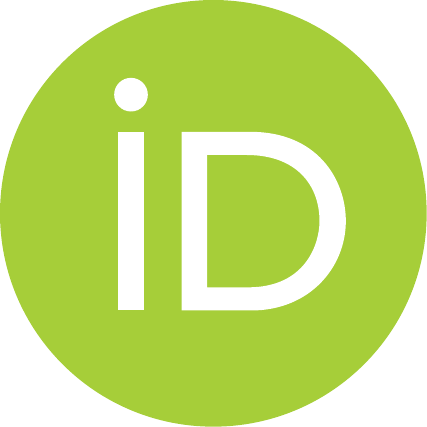}}}%
	\parbox{\wd0}{\box0}
	\endgroup}
\makeatother
\begin{document}
\def\makeheadbox{\relax}

\title{Robust subgroup discovery}

\subtitle{Discovering subgroup lists using MDL}


\author{Hugo M. Proen\c{c}a~\orcid{0000-0001-7315-5925}  
\and Peter Gr\"{u}nwald~\orcid{0000-0001-9832-9936}
\and Thomas B\"{a}ck~\orcid{0000-0001-6768-1478} 
\and Matthijs van Leeuwen~\orcid{0000-0002-0510-3549}
}

\authorrunning{Hugo M. Proen\c{c}a et al.} 

\institute{
    Hugo Manuel Proen\c{c}a  \at
	LIACS, Niels Bohrweg 1, 2333 CA Leiden, Netherlands\\
	\email{h.manuel.proenca@liacs.leidenuniv.nl}           
	\and
	Peter Gr\"{u}nwald \at
	CWI, Science Park 123, 1098 XG Amsterdam\\
	\email{peter.grunwald@cwi.nl}           
	\and	
	Thomas B\"{a}ck \at
	LIACS, Niels Bohrweg 1, 2333 CA Leiden, Netherlands\\
	\email{t.h.w.baeck@liacs.leidenuniv.nl}           
	\and
	Matthijs van Leeuwen \at
	LIACS, Niels Bohrweg 1, 2333 CA Leiden, Netherlands\\
	\email{m.van.leeuwen@liacs.leidenuniv.nl}
}

\date{}

\maketitle

\begin{abstract}
	We introduce the problem of \emph{robust subgroup discovery}, i.e., finding a set of interpretable descriptions of subsets that 1) stand out with respect to one or more target attributes, 2) are statistically robust, and 3) non-redundant. Many attempts have been made to mine either \emph{locally} robust subgroups or to tackle the pattern explosion, but we are the first to address both challenges at the same time from a \emph{global} modelling perspective. 
    First, we formulate the broad model class of subgroup lists, i.e., ordered sets of subgroups, for univariate and multivariate targets that can consist of nominal or numeric variables, including traditional top-$1$ subgroup discovery in its definition. This novel model class allows us to formalise the problem of optimal robust subgroup discovery using the Minimum Description Length (MDL) principle, where we resort to optimal Normalised Maximum Likelihood and Bayesian encodings for nominal and numeric targets, respectively. 
    Second, finding optimal subgroup lists is NP-hard. Therefore, we propose SSD++, a greedy heuristic that finds good subgroup lists and guarantees that the most significant subgroup found according to the MDL criterion is added in each iteration. In fact, the greedy gain is shown to be equivalent to a Bayesian one-sample proportion, multinomial, or t-test between the subgroup and dataset marginal target distributions plus a multiple hypothesis testing penalty. Furthermore, we empirically show on $54$ datasets that SSD++ outperforms previous subgroup discovery methods in terms of quality, generalisation on unseen data, and subgroup list size.
\end{abstract}

\section{Introduction}
\label{sec:intro}

Exploratory Data Analysis (EDA) \citep{tukey1977exploratory} aims at enhancing its practitioner's natural ability to recognise patterns in the data being studied. The more she explores, the more she discovers, but also the higher the risk of finding interesting results arising out of coincidences, e.g., spurious relations between variables that have no connection in the real world. Intuitively this corresponds to testing multiple hypotheses without realising it. This duality of EDA requires a thorough analysis of results and highlights the need for statistically robust techniques that allow us to explore the data in a responsible way. While EDA encompasses all techniques referring to data exploration, \emph{Subgroup Discovery} (SD) \citep{klosgen96sd,atzmueller2015subgroup} is the subfield that is concerned with discovering interpretable descriptions of subsets of the data that stand out with respect to a given target variable, i.e., \emph{subgroups}. This work aims at improving the discovery of subgroup lists, i.e., ordered sets of subsets, that describe different regions of the data while being statistically robust by themselves and against multiple hypothesis testing. Two simple examples of subgroup lists can be found in Figures~\ref{fig:zoo_example} and \ref{fig:car_example}.  \\  

\begin{figure}[!hbt]\centering
	\ra{1.1}\small \begin{tabular}{@{}llrrrrrrrr@{}}																				
		&		&		&	 \multicolumn{7}{r}{$\boldsymbol{\Pr(animal type = \cdots \given s)}$ in $\%$} \\	\cmidrule(l){4-10}
		$s$	&	\textbf{description}	&	$n_s$	&	Mammal	&	Fish	&	 Invert.	&	Bug	&Reptile&	Amph.	&	Bird	\\ \midrule	
		1	&backbone = no&$	18	$&$	0	$&$	0	$&$	56	$&$	44	$&$	0	$&$	0	$&$	0	$	\\ \cmidrule(l){2-10}
		2	&breathes = no&$	14	$&$	0	$&$	93	$&$	0	$&$	0	$&$	7	$&$	0	$&$	0	$	\\ \cmidrule(l){2-10}
		3	&feathers = yes&$	20	$&$	0	$&$	0	$&$	0	$&$	0	$&$	0	$&$	0	$&$	100	$	\\ \cmidrule(l){2-10}
		4	&milk = no&$	8	$&$	0	$&$	0	$&$	0	$&$	0	$&$	50	$&$	50	$&$	0	$	\\ \cmidrule(l){2-10}
		5	&feathers = no &$	41	$&$	100	$&$	0	$&$	0	$&$	0	$&$	0	$&$	0	$&$	0	$	\\ \midrule
		\multicolumn{2}{@{}l}{dataset distribution}&$	0^*	$&$	41	$&$	13	$&$	10	$&$	8	$&$	5	$&$	4	$&$	2	$\\ 	\bottomrule
	\end{tabular}
	\caption{\emph{Zoo} dataset subgroup list obtained by SSD++. \emph{Zoo} contains one nominal target variable with $7$ classes, $101$ instances, and $15$ binary and $1$ numeric variables. $n_s$ refers to the number of instances covered by subgroup `$s$' defined by `description'. $\Pr(animal type = * \given s)$ denotes the estimated probability (in $\%$) of each class label occurring within the subgroup. The bottom row shows the marginal probability distribution of the dataset. $*$ concerns instances not covered by any of the five subgroups. For illustrative purposes, the probabilities displayed correspond to the empirical probabilities in the data, not to the probabilities as would be obtained using the appropriate estimator }\label{fig:zoo_example}																				
\end{figure}

\begin{figure}[hbt!]\centering												
	\ra{1.1}\small \begin{tabular}{@{}llrrr@{}}											
		&		&		&	 \multicolumn{2}{r@{}}{\phantom{a}$\boldsymbol{price \: (\text{K})}$ } \\	\cmidrule(l){4-5}
		$s$\phantom{,}	&	\textbf{description} of automobile	specifications &	$n_s$	&$	\hat{\mu}	$&$	\hat{\sigma}	$\\	\midrule
		1	&	 weight = heavy  \& consumption-city $ \leq 8 $ km/L	&$	11	$&$	35	$&$	8	$\\ 	
		2	&\small	fuel-type = gas \& consumption-city $ \geq 13 $ km/L	&$	45	$&$	7	$&$	1	$\\ 	
		3	&\small	 weight = light \& wheel-base = low	&$	35	$&$	9	$&$	1	$\\ 	
		4	&\small	length = medium \& $13\leq $  consumption-city $\leq 15  $ km/L	&$	27	$&$	10	$&$	2	$\\ 	
		5	&\small	peak-rpm = medium	&$	49	$&$	16	$&$	3	$\\ 	
		6	&\small	engine-size = medium	&$	12	$&$	26	$&$	7	$\\ 	 \midrule
		\multicolumn{2}{l}{dataset overall distribution}			&$	18^*	$&$	13	$&$	8	$\\ 	\bottomrule
	\end{tabular}											
	\caption{\emph{Automobile import 1985} subgroup list obtained with SSD++. The dataset contains \emph{price} as numeric target variable, $197$ examples, and $17$ variables. The dataset was modified, some variables removed and others discretised, for ease of presentation. $n_s$ refers to the number of instances covered by subgroup `$s$' defined by `description', $\hat{\mu}$ and $\hat{\sigma}$ its estimated mean and standard deviation for the target variable in thousands of dollars ($K$). $*$ concerns instances not covered by any of the five subgroups}\label{fig:car_example}											
\end{figure}

\emph{Subgroup discovery} (SD) can be seen as a generalisation of association rule mining or as the exploratory counterpart of rule learning, where the targets/consequent of the rules are fixed and rules are ranked according to quality measures combining subgroup size and deviation of the target variable(s) with respect to the overall distribution in the data. In its traditional form, subgroup discovery is also referred to as top-$k$ subgroup mining \citep{atzmueller2015subgroup}, which entails mining the $k$ top-ranking subgroups according to a \emph{local} quality measure and a number $k$ selected by the user. Since its conception, subgroup discovery has been developed for various types of data and targets, e.g., nominal, numeric~\citep{grosskreutz2019numeric}, and multi-label~\citep{van2010maximal} targets.
SD has been applied in a wide range of different domains \citep{herrera2011overview,atzmueller2015subgroup}, such as identifying the properties of materials \citep{goldsmith2017uncovering}, unusual consumption patterns in smart grids \citep{jin2014subgroup}, identifying the characteristics of delayed flights \citep{proencca2018identifying}, and understanding the influence of pace in long-distance running \citep{de2018effects}.

Although SD appeals to several domains, top-$k$ mining traditionally suffers from three main issues that make it impractical for many applications: $1)$ poor computational efficiency of exhaustive search with the more relevant quality measures \citep{van2012diverse,bosc2018anytime}; $2)$ \emph{redundancy} of mined subgroups, i.e., the fact that subsets with the highest deviation according to a certain \emph{local} quality measure tend to cover the same region of the dataset with slight variations in their description of the subset \citep{van2012diverse}; and  $3)$ lack of generalisation or statistical robustness of mined subgroups \citep{van2016expect}.

In this work, we focus on the last two issues together: reducing \emph{redundancy} by finding small lists of subgroups that describe the differences in the data well; and obtaining \emph{statistically robust} subgroups. First, we define the problem of robust subgroup discovery in its full generality using the minimum description length (MDL) principle. Second, as optimally solving the problem is unfeasible, we focus solely on subgroup lists (not sets) and propose a greedy algorithm that finds good subgroup lists using a \emph{local} objective for binary, nominal, or numeric target variables. Moreover, we show that this MDL-based greedy gain is equivalent to maximising the Bayes factor between each subgroup's distribution and the dataset marginal distribution plus a penalty for multiple hypothesis testing. Since subgroup lists with only one subgroup are equivalent to \emph{top-$1$} mining, all statistical properties developed here can be directly applied to top-$k$ subgroup discovery. For a formal definition of the \emph{Robust Subgroup Discovery} problem and our approximation, please refer to Section~\ref{sec:sec_problem_statement}.

Note that we restrict our focus to finding \emph{subgroup lists} with \emph{the MDL principle} because 1) subgroup lists are one of the core subgroup set models and one of the first that was proposed \citep{lavravc2004subgroup}; 2) they allow for an optimal formulation based on the MDL principle due to their property of unambiguously partitioning the data into non-overlapping parts; 3) the subgroups can be interpreted sequentially, i.e., from most to least relevant discovered subgroup; and 4) the MDL principle is a statistical criterion for model selection that requires the fewest assumptions possible about the model class, data, and (data) generation process.\\

In recent years both \emph{redundancy} and \emph{statistical robustness} issues have been partially addressed, mostly independent of each other. We next briefly discuss recent advances and limitations and refer the reader to Section~\ref{sec:related_work} for an in-depth analysis of related work.

In terms of \emph{redundancy}, the first main limitation of existing works is their focus on one type of target variables, such as binary targets \citep{bosc2018anytime,belfodil2019fssd}, nominal targets \citep{lavravc2004subgroup}, or numeric targets \citep{lijffijt2018subjectively}, where only DSSD focuses on univariate and multivariate nominal and numeric targets \citep{van2012diverse}. The second main limitation is the lack of an optimality criterion for subgroup sets or lists, where the only exception is FSSD \citep{belfodil2019fssd}. It is important to emphasise that some works aim at finding sequential subgroups or subgroup \emph{lists}, while others aim at finding unordered sets or subgroup \emph{sets}. Subgroup lists are akin to rule lists \citep{proencca2020interpretable} in the sense that each subgroup/rule needs to be interpreted sequentially, and thus they should be read and interpreted sequentially and cannot overlap, while subgroup sets are allowed to overlap. In this work, we focus solely on subgroup lists, and although previous works often did not use this term, we retroactively rename those models that are, in fact, subgroup lists.

In terms of \emph{statistical robustness}, the approaches can be divided into post-processing techniques \citep{duivesteijn2011exploiting,van2016expect} and modified quality measures \citep{song2016subgroup,song2017model}. Post-processing techniques first mine the top-$k$ subgroups and then apply a statistical test to find statistical significance. Modified quality measure approaches, which are more closely related to our work, add a statistical test directly to the quality measure definition; however, they do not consider multiple-hypothesis testing correction and only focus on top-k mining.

Next, we show how our contributions address both issues in a unified way.





\paragraph{\textbf{Contributions.}} We aim to bridge the gap in the literature by finding the best non-redundant subgroup list from a \emph{global} dataset perspective while guaranteeing the \emph{local} quality of the found subgroups, making the approach statistically robust from both perspectives. Two examples of subgroup lists for nominal and numeric targets can be seen in Figures~\ref{fig:zoo_example} and \ref{fig:car_example}. To solve this problem, we propose a formal definition of a \emph{subgroup list} and employ the Minimum Description Length (MDL) principle~\citep{rissanen78} to define an optimal subgroup list from a global perspective. We provide this formalisation for univariate and multivariate nominal and numeric targets. Notably, the subgroup that minimises the MDL-optimal formulation for a subgroup list with one subgroup is the same subgroup that would be found by top-$1$ subgroup discovery with Weighted Kullback-Leibler divergence (WKL) as a quality measure. This makes our proposal the first \emph{global} formulation of subgroup set discovery that is a direct generalisation of traditional subgroup discovery. Thus, all statistical properties developed in this work also apply to top-$k$ subgroup discovery.

As finding optimal subgroup lists is NP-Hard, we propose SSD++, a heuristic algorithm that finds ``good'' subgroup lists. SSD++ combines beam search---to find individual subgroups---with greedy search---to iteratively add the best-found subgroup to the subgroup list. Maximising the MDL criterion in each iteration guarantees that each subgroup added to the list adheres to a \emph{local} statistical test equivalent to Bayesian proportions, multinomial, or t-test (for binary, nominal and numeric targets, respectively) plus a penalty to compensate for multiple hypothesis testing. 

\paragraph{Previous work.} This work is an extension of \cite{proencca2020discovering} and builds on some of the results of \cite{proencca2020interpretable}. The former introduces MDL-based subgroup lists for univariate numeric target variables and SSD++, a heuristic algorithm for finding such subgroup lists. The current manuscript significantly extends our previous work by generalising the MDL data encoding to three new target variable types: multivariate numeric targets, and univariate and multivariate nominal targets. To find and prove its efficacy, the algorithm and empirical results are also extended to those new target variables. Moreover, the current work provides a new interpretation of the MDL encoding and of the greedy gain as an MDL equivalent to Bayesian testing.

\cite{proencca2020interpretable} introduced MDL-based rule lists for classification; however, compared to the current work, it is limited to Boolean explanatory and nominal target variables, its model and data encoding are less optimal, it has no local statistical guarantees, and the algorithm was less flexible in accepting user-defined hyperparameters.

\paragraph{Summary of contributions.} To summarise, the \emph{primary} contributions presented in this work---including the contributions originally from \cite{proencca2020discovering}, which we indicate with a * below---are:
\begin{enumerate}
	\item \textbf{Subgroup list model class} -- We define the subgroup list model class* over a tabular dataset in general (Section~\ref{sec:model_class}), providing a \emph{global} formulation for the problem of sequential subgroup mining, and in particular for univariate and multivariate nominal targets (Section~\ref{sec:nominal_data}), and univariate numeric* and multivariate numeric targets (Section~\ref{sec:numeric_data}). 
	\item \textbf{Robust subgroup discovery using MDL} -- We define the optimal problem of robust subgroup discovery in Section~\ref{sec:problem_statement} for the case of subgroup lists using the MDL principle. We show the relation of model encoding and multiple-hypothesis testing (Section~\ref{sec:MDL}), and resort to the optimal Normalised Maximum Likelihood (NML) encoding for nominal targets (Section~\ref{sec:nominal_data}) and the Bayesian encoding with non-informative priors for numeric targets* (Section~\ref{sec:numeric_data}). Notably, we show that this problem formalisation is equivalent to the standard definition of top-$1$ subgroup discovery with WKL as a quality measure for the case of a subgroup list with one subgroup (Section~\ref{sec:sd_proof}).
	\item \textbf{Greedy MDL algorithms maximise local statistical test} -- We show that the greedy gain commonly used in the MDL for pattern mining literature can be interpreted as an MDL equivalent to a \emph{local} Bayesian hypothesis test, a.k.a. Bayes factor, on the likelihood of the data being better fitted by the greedy extended model versus the current model plus a penalty for the extra model complexity (Section~\ref{sec:mdl_statistic}). In the case of our specific algorithm, SSD++, the greedy objective is equivalent to one-sample Bayes proportions, multinomial, and t-test (for binary, nominal and numeric* targets, respectively) plus a penalty to compensate for multiple hypothesis testing (Section~\ref{sec:mdl_statistic}).
\end{enumerate}

Moreover, this work includes the following \emph{secondary} contributions, the details of which are all included in the appendices for the interested reader:
\begin{enumerate}
    \setcounter{enumi}{3}
	\item \textbf{Normalised Maximum Likelihood for partition models} -- Derivation of the Normalised Maximum Likelihood (NML) optimal encoding, a refined MDL encoding, for model classes that partition the data for nominal target variables---subgroup lists, rule lists, trees, etc. (Appendix~\ref{appendix:NML_derivation}).
	\item \textbf{Bayesian encoding of normal distributions}
-- Derivation of a Bayesian optimal encoding of normal distributions with non-informative priors for numeric targets* (Appendix~\ref{appendix:Bayesian_derivation}). It is shown that for a large number of instances, it converges to the BIC* (Appendix~\ref{appendix:BIC}). Similarly to the NML encoding, it can be used by any model class that unambiguously partitions the data, such as subgroup lists, rule lists, trees, etc.
	\item \textbf{Subgroups discovery versus rule-based prediction} -- We demonstrate the difference between the formal objectives for subgroup discovery and predictive rule models, such as classification rule lists, from the perspective of our MDL-based approach (Appendix~\ref{appendix:proof_sd_vs_prediction}).
\end{enumerate}

\paragraph{\textbf{Structure of the paper.}} Besides the customary introduction, related work, and conclusion, this work contains two preliminary sections---Sections~\ref{sec:sd} and~\ref{sec:back_model_sel}---a problem statement section---Section \ref{sec:sec_problem_statement}---three theoretical sections related with MDL encoding---Sections \ref{sec:MDL} to \ref{sec:mdlsl_wklsd_bayes}---one algorithm section---Section~\ref{sec:SSDpp}---and two empirical results sections---Sections \ref{sec:exps_results} and \ref{sec:case_study}. The preliminary sections introduce the notation and basic concepts, while the problem statement formalises the objective of this work. Together, these three sections---where a reader familiar with the topic at hand can skim through---allow reading each of the following sections independently of each other.

The specific details of each section can be described as follows. Section~\ref{sec:sd} introduces the notation used throughout this work and the preliminaries related to subgroup discovery and subgroup set discovery. After that, Section~\ref{sec:back_model_sel} presents the preliminaries related to model selection in the frequentist, Bayesian, and the Minimum Description Length (MDL) principle branches of statistics. Then, Section~\ref{sec:sec_problem_statement} defines the subgroup list model class, the robust subgroup discovery problem statement, and a novel quality measure for subgroup lists based on the Weighted Kullback-Leibler (WKL) divergence. After that, Section~\ref{sec:MDL} shows the MDL principle model encoding part of the subgroup list and its relation to multiple hypothesis testing. Section~\ref{sec:data_encoding} presents the data encoding for nominal and numeric targets. Then, Section~\ref{sec:mdlsl_wklsd_bayes} demonstrates how our MDL-based formulation of subgroup lists equals WKL-based subgroup discovery and Bayesian testing. After that, Section~\ref{sec:SSDpp} presents SSD++, a heuristic algorithm to mine subgroup lists, as well as its time complexity. Then, in Section~\ref{sec:exps_results} we show the empirical results of our proposed method when compared against the state-of-the-art algorithms for univariate and multivariate nominal and numeric targets over $54$ datasets. After that, in Section~\ref{sec:case_study} we apply robust subgroup discovery to find how descriptions of the socioeconomic background affect the grades of engineering students in Colombia. Then, Section~\ref{sec:related_work} covers the most relevant related work, together with the main differences to our approach. Finally, Section~\ref{sec:conclusion} presents our main conclusions and future work.

%
%
\section{Preliminaries: notation and subgroup discovery}	
\label{sec:sd}

In this section, the mathematical notation used throughout this work is introduced together with all the topics of Subgroup Discovery (SD), Subgroup Set Discovery (SSD), and beam search and separate and conquer algorithms for subgroup discovery.\\

This section is divided as follows. First, Section~\ref{sec:back_data} defines the notation used for data, target variables, and subgroups. Then, in Section~\ref{sec:back_sdqm}, we present the task of subgroup discovery and how to rank the discovered subgroups according to a quality measure. A particular emphasis is given to the Kullback-Leibler divergence as it plays a fundamental role in our definition of MDL-based subgroup lists. After that, Section~\ref{sec:back_ssd} defines the task of subgroup set discovery. Finally, Section~\ref{sec:back_alg_sd} briefly describes beam search and separate and conquer, algorithms that we use to generate the top-$k$ subgroups and add subgroups to the subgroup list. 



\subsection{Data and subgroups}\label{sec:back_data}
Consider a dataset $D = (\Xmat, \Ymat) =\{(\x^1,\y^1),(\x^2,\y^2),...,(\x^n,\y^n)\}$ of $n$ \emph{i.i.d.} instances. Each instance $(\x,\y)$ is composed of a vector of explanatory variable values $\x$ and a vector of target variable values $\y$. 
Each observed explanatory vector has $m$ values $\x = [x_1,...,x_m]$, one for each variable $X_1,...,X_m$. The domain of a variable $X_j$, denoted $\mathcal{X}_{j}$, can be one of two types: nominal or numeric. Similarly, each observed target vector is composed of $t$ values $\y = [y_1,...,y_t]$, one for each target variable $Y_1,...,Y_t$, with associated domains $\Y_j$. The target variables can be of two types: numeric, or nominal. In the numeric case, the domain is $\Y_j = \mathbb{R}$ and in the nominal it is $\Y_j = \{1, \cdot, k\}$, with $\Y_j$ the set of classes/categories of variable $Y_j$. For the complete notation used throughout this work please refer to Table~\ref{table:notation}.

Note that we use subscripts on the dataset variables ($D,\Xmat, \Ymat,X,Y,x,y$) to indicate column indices and superscripts for row indices. In the case of other notation, such as the number of elements $n$ or statistics $\mu,\sigma$ we will not use the superscript as it could be confused with the exponentiation of that value. Also, $X_i$ (resp. $Y_i$) refers to both the properties of the $i^{th}$ explanatory (resp. target) variable and to all the values of this variable for a specific column. \\

Depending on the type and number of targets (one or multiple), the type of problem can be divided into \emph{four} categories: \emph{1) single-nominal}; \emph{2) single-numeric}; \emph{3) multi-nominal}; and \emph{4) multi-numeric}. 
In machine learning, the single-numeric case corresponds to regression, the single-nominal to classification, and in the case of more than one variable their multi-target generalisations, respectively. 


\begin{table}													
	\caption{Notation table.}\label{table:notation}													
	\renewcommand{\arraystretch}{0.84} \begin{tabular}{@{}ll@{}}\toprule													
		Symbol	&	Definition									\\	\midrule
		$	D  = \{\Xmat, \Ymat\}	$&	Labelled dataset.									\\	
		$	\Xmat	$&	Dataset of explanatory variables of $D$.									\\	
		$	X	$&	An explanatory variable of $\Xmat$.									\\	
		$	\mathcal{X}	$&	Domain of $X$.									\\	
		$	\x	$&	A explanatory variables sample of $\Xmat$.									\\	
		$	x	$&	The value of sample $\x$ for variable $X$.									\\	
		$	 \Ymat	$&	Dataset of target variables of $D$.									\\	
		$	Y	$&	An target variable of $\Ymat$.									\\	
		$	\mathcal{Y}	$&	Domain of $Y$.									\\	
		$	\y	$&	A target variables sample of $\Ymat$.									\\	
		$	y	$&	The value of sample $\y$ for variable $Y$.									\\	
		$	|\cdot|	$&	Number of elements in a set, as e.g., $|D|$ for number of samples.									\\	
		$	i	$&	Index for subsetting by row.									\\	
		$	j	$&	Index for subsetting by column.									\\	
		$	v	$&	A generic explanatory variable.									\\	
		$	k	$&	Number of classes of a nominal target variable.									\\	
		$	n	$&	Number of examples in dataset $D$.									\\	
		$	m	$&	Number of explanatory variables.									\\	
		$	t	$&	Number of target variables.									\\	
		$	d	$&	Subscript associated with dataset distribution or default rule.									\\	
		$	M	$&	Subgroup list model (including subgroups S and default rule).									\\	
		$	S	$&	Subgroups in model M.									\\	
		$	\omega	$&	Number of subgroups in $M$.									\\	
		$	s	$&	A subgroup.		\\
		$ \Omega$ & The set of all subgroups \\	
		$	a	$&	Description of a subgroup.									\\	
		$	a_i	$&	Description of the $i^{th}$ subgroup in model M.									\\	
		$	D^{a} =\{\Xmata,\Ymata \}	$&	Samples of dataset $D$ covered by description $a$.									\\	
		$	n_a	$&	Number of samples in $D^{a}$. $n_a = |D^{a}|$.									\\	
		$	D^{i} =\{\Xii,\Ymati \}	$&	Samples of dataset $D$ covered by the $i^{th}$ subgroup in model M. 									\\	
		$	n_i	$&	Number of samples in $D^{i}$. $n_i = |D^{i}|$.									\\	
		$	Dist(\Theta)	$&	Generic probability distribution with parameters $\Theta$.									\\	
		$	\mathcal{N}(\mu; \sigma)	$&	Normal probability distribution with parameters $\mu$ and $\sigma$.									\\	
		$	 Cat(p_{1},\cdots,p_{k})	$&	Categorical probability distribution with $p_i$ probability per category.									\\	
		$	p_{y|c}	$&	Probability of category $y$ given description $a$, i.e., $\Pr(y\given a)$									\\	
		$	\mu	$&	Mean value parameter.									\\	
		$	\sigma	$&	Standard deviation parameter.									\\	
		$	\hat{\theta}	$&	Maximum likelihood estimation of parameter $\theta$.									\\	
		$	q(a)	$&	Subgroup discovery quality measure.									\\	
		$	Q(S)	$&	Subgroup set discovery quality measure.									\\	
		$	f(\Ta,\Td)	$&	Function of differences between distribution $\Ta$ and $\Td$.									\\	
		$	\alpha	$&	Tradeoff between subgroup coverage and distribution difference.									\\	
		$	KL	$&	Kullback-Leibler divergence general form.									\\	
		$	KL_{\mathrm{Cat}}	$&	Kullback-Leibler divergence for categorical distributions.									\\	
		$	KL_{\mu}	$&	Kullback-Leibler divergence for location distributions.									\\	
		$	KL_{\mu,\sigma}	$&	Kullback-Leibler divergence for normal distributions.									\\	
		$	WKL	$&	Weighted Kullback-Leibler divergence general form. 									\\	
		$	\textrm{SWKL}	$&	Sum of Weighted Kullback-Leibler divergences.									\\	
		$	L_\mathbb{N}	$&	Universal code of integers.									\\	
		$	L_{\mathrm{NML}}(Y_j^i)	$&	Normalised Maximum Likelihood length of encoding of data $Y_j^i$.									\\	
		$	\COMP(n_a,k)	$&	Multinomial distribution complexity  with $n_a$ points and $k$ categories.									\\	
		$	L_{\mathrm{Bayes}}	$&	Bayesian length of encoding with improper priors.									\\	
		$	\Ytwo	$&	The two points that make the Bayesian encoding proper.									\\	
		$	L_{\mathrm{Bayes}2.0}	$&	Bayesian length of encoding made proper with first $2$ points.									\\	
		$	\Gamma (n)	$&	Gamma function, the extension of the factorial to real numbers.									\\	
		$	\Delta_{\beta} L(D,M \oplus s)	$&	Compression gain of adding subgroup $s$ to model $M$.									\\
		$	\Xi (\Ti,\Td)	$&	Parametric statistical test between $\Ti$ and $\Td$.									\\	
		$	\beta	$&	Level of normalisation of the compression gain.									\\	
		$	\zeta	$&	Set of all items (possible single conditions) in $\Xmat$.									\\	
		$	d_{max}	$&	Beam search maximum depth of search.									\\	
		$	w_{b}	$&	Beam search beam width.									\\	
		$	n_{cut}	$&	Number of cut points for numeric discretisation.									\\	\bottomrule
	\end{tabular}													
\end{table}

\subsubsection{Subgroups}\label{sec:back_sd}
A subgroup, denoted by $s$, consists of a \emph{description} (also intent) that defines a \emph{cover} (also extent), i.e., a subset of dataset $D$. 

\paragraph{Subgroup description:} A description $a$ is a Boolean function over all explanatory variables $X$. Formally, it is a function $a : \mathcal{X}_{1} \times \cdots \times\mathcal{X}_{m} \mapsto \{false,true\}$. In our case, a description $a$ is a conjunction of conditions on $\Xmat$, each specifying a specific value or interval on a variable. The domain of possible conditions depends on the type of a variable: numeric variables support \emph{greater and less than} $\{\geq, \leq\}$; nominal support \emph{equal to} $\{ =\}$. The size of a description $a$, denoted $|a|$, is the number of conditioned variables it contains.

\paragraph{Example 1: } In Figure~\ref{fig:car_example}, subgroup $1$ has a description of size $|a| = 2$, with one condition on a nominal variable: $\{\mbox{weigth} = \mbox{heavy} \}$; and another on a numeric variable: $\{\mbox{consumption-city} \leq 8 km/L\}$.

\paragraph{Subgroup cover:} The cover is the bag of instances from $D$ where the subgroup description holds true. Formally, it is defined by:
\begin{equation}
D^{a}= \{(\vec{x},\y) \in D \given a \sqsubseteq \vec{x} \} = \{X_1^a,\cdots, X_m^a,Y_1^a,\cdots,Y_t^a \} =  \{\Xmata,\Ymata \},
\end{equation}
where we use $a \sqsubseteq \vec{x}$ to denote $a(\vec{x}) = true$. Further, let $n_a = |\Da|$ denote the coverage of the subgroup, i.e., the number of instances it covers. 

\paragraph{Example 2 (continuation):} In Figure~\ref{fig:car_example}, subgroup $1$ covers $11$ instances in the dataset which can be found by conditions in its description, and thus its coverage is $11$.

\subsubsection{Subgroups as probabilistic rules}\label{sec:back_prob_rules}
As $\Da$ encompasses both the explanatory and target variables, the effect of $a$ on the target variables can be interpreted as a probabilistic rule. Regarding the multiple target variables, we assume that they are \emph{independent}. This simplifies the problem and is a common approach in multi-label classification \citep{herrera2016multilabel}. Thus, the general form of the rule is: 
\begin{equation}\label{eq:prob_rule}
a \mapsto y_1 \sim Dist(\hat{\Theta}_{1}^a),\cdots, y_t \sim Dist(\hat{\Theta}_{t}^a),
\end{equation}
where $y_j$ is a value of variable $Y_j$, $Dist$ is a probability distribution (defined later) and $\hat{\Theta}_{j}^a$ is the shorthand for the maximum likelihood estimation of the parameters of $Dist$ over values $Y_j^a$, i.e., $\hat{\Theta}_{j}^a = \hat{\Theta}_{j}(Y^a)$. Thus, $y_i \sim Dist(\hat{\Theta}_{j}^a)$ tells us that the values of variable $Y_j$ are distributed according to a distribution $Dist$ with parameters $\hat{\Theta}_{j}^a$ estimated over the values $Y_j^a$. The vector of all parameter values of a rule is denoted by $\Theta^a$. In our case, $Dist$ can be a \emph{categorical} or \emph{normal} distribution in the nominal or numeric target case, respectively. \\
In the numeric case the normal distribution is represented as: $\mathcal{N}(\hat{\mu},\hat{\sigma})$. In the nominal case the distribution is $Cat(\hat{p}_{1}, \cdots, \hat{p}_{k})$, where $k$ is the number of classes (or categories) of the corresponding variable and $\hat{p}_c$ the estimated probability for class $c$.

\paragraph{Example 3 (continuation):} Revisiting the \emph{Automobile import} subgroup list in Figure~\ref{fig:car_example}, the description and corresponding statistics for the second subgroup are $a =$ $\{$fuel-type $=$ gas \& consumption-city $ \geq 13 $ km/L $\}$ and $\hat{\Theta}^{a_2} = \{ \hat{\mu} = 7; \hat{\sigma} = 1\}$, respectively, where the units are thousands of dollars ($\text{K}$). This corresponds to the following normal probability distribution:
\begin{equation*}
\mbox{price (\text{K})} \sim \mathcal{N}(\hat{\mu} = 7; \hat{\sigma} = 1 ) \\
\end{equation*}

\paragraph{Example 4 (continuation):} In the case of the \emph{Zoo} subgroup list in Figure~\ref{fig:zoo_example}, the description for the first subgroup is $a = \{$backbone $=$ no$\}$, and its corresponding statistics are  $\hat{\Theta}^{a_1} =$ $\{\hat{p}_{1} = 0; \hat{p}_{2} =0;  \hat{p}_{3} = 0.56 ; \hat{p}_{4} = 0.44; \hat{p}_{5} = 0 ; \hat{p}_{6} = 0 ; \hat{p}_{7} = 0 \}$, where the class labels $1,...,7$ correspond to the animal types in the order of Figure~\ref{fig:zoo_example}. The target variable follows the following categorical distribution:
\begin{equation*}
\mbox{animal\_type} \sim Cat(\hat{p}_{1},\hat{p}_{2},\hat{p}_{5},\hat{p}_{6},\hat{p}_{7} = 0.00; \hat{p}_{3} = 0.56 ;\hat{p}_{4} = 0.44) \\
\end{equation*}

\subsection{Subgroup discovery}\label{sec:back_sdqm}

Subgroup discovery is the data mining task of finding subgroups that stand out with respect to some given target variable(s). The definition of standing out, also known as interestingness, is quantified by a quality measure, which depends on the task at hand \citep{webb1995opus,klosgen96sd}. Generally, these measures quantify quality by how different the target variable distribution of a subgroup is from what is defined as `normal' behaviour in a dataset. In the case of structured data, a subgroup generally takes the form of an association rule, and the `normal' behaviour is usually measured by the average behaviour of the target variable of that dataset \citep{atzmueller2015subgroup}. 

\paragraph{Quality measures.} Thus, depending on the target variable and task, different quality measures can be chosen to assess the quality (or interestingness) of a subgroup description $a$, over is cover $\Da$. In general, quality measures have two components: 1) representativeness of the subgroup in the data, based on coverage $n_a =|\Da|$; and 2) a function of the difference between statistics of the empirical target distribution of the pattern, $\Ta = \hat{\Theta}(\Ymata)$, and the overall empirical target distribution of the dataset, $\Td = \hat{\Theta}(\Ymat)$. The latter corresponds to the statistics estimated over the whole data, e.g., in the case of the \emph{Automobile import} subgroup list of Figure~\ref{fig:car_example} it is $\Td = \{ \hat{\mu} = 13; \hat{\sigma} = 8 \}$ and it is estimated over (all) $197$ instances of the dataset.

The general form of a quality measure to be maximised is
\begin{equation}\label{eq:generalmeasure}
q(a) = (n_a)^\alpha f(\Ta,\Td), \; \alpha \in [0,1],
\end{equation}

where $\alpha$ allows to control the trade-off between coverage and the difference of the distributions, and $f(\Ta,\Td)$ is a function that measures how different the subgroup and dataset distributions are. As an example, the most commonly adopted quality measure for single-numeric targets is Weighted Relative Accuracy (WRAcc) \citep{lavravc1999rule}, with $\alpha=1$ and $f(\hat{\Theta}_a,\hat{\Theta}_d) = \hat{\mu}_a-\hat{\mu}_d$ (the difference between subgroup and dataset averages).

\subsubsection{Weighted Kullback-Leibler divergence}
\label{sec:wkl}

Another commonly adopted measure is the Weighted-Kullback Leibler divergence (WKL) \citep{van2011non}. This is also the measure that we consider throughout this work because of 1) its flexibility in terms of (number and types of) supported target variables; and 2) its relationship to the MDL principle (see Sections~\ref{sec:sd_proof}); and 3) it arises from using the Log-loss for assessing the goodness of fit of the dataset and subgroup distribution, which is a proper scoring rule \citep{song2016subgroup}.

WKL is defined as the Kullback-Leibler (KL) divergence \citep{kullback1951information} between a subgroup's and dataset target distribution $KL(\Ta; \Td)$ linearly weighted by its coverage. Revisiting Eq.~\eqref{eq:generalmeasure} this corresponds to $f(.) = KL(.)$ and $\alpha = 1$. The definition of WKL for a univariate target variable $Y$ is given by:
\begin{equation}\label{eq:wkl}
WKL(\Ta; \Td) = n_a KL(\Ta; \Td),
\end{equation}
where $KL(\Ta; \Td)$ is the Kullback-Leibler divergence between subgroup and dataset for target $Y$. The KL divergence in Eq.~\eqref{eq:wkl} depends on the probabilistic model chosen to describe the target variables. In its general form, the KL divergence can be defined as:
\begin{equation}
KL(\Taj; \Tdj) = \sum_{y \in \Ya} \Pr(y \given \Taj) \log \left( \frac{\Pr(y \given \Taj)}{\Pr(y \given \Tdj)} \right),
\end{equation}
where the logarithm is to the base two (like all logs in this work).
Thus the choice of the distribution used to describe the target is of great importance and should reflect what the analyse would like to find in the data. Now, depending on the type of target we will see show how to compute $WKL(\Ta; \Td)$. It is easy to see that for multivariate targets, we either use a multivariate distribution, e.g., a multivariate normal distribution or assume that they are \emph{independent} target variables, where the total WKL turns out to be just the sum of the WKL for each target variable.\\

We will now provide the definitions of WKL for univariate categorical and normal distributions.

\subsubsection{Weighted Kullback-Leibler for categorical distributions} 
In the case of a univariate \emph{nominal target} $Y$, the distribution can be uniquely described by a categorical distribution with the probability of each category $\Ta = \{ \hat{p}_{1|a},...,\hat{p}_{k|a} \}$, so that the $KL(\Ta; \Td)$ of Eq.~\eqref{eq:wkl} takes the form of:
\begin{equation}
KL_{\mathrm{Cat}}(\Ta; \Td)= \sum_{c \in \Y}  \hat{p}_{c|a} \log \left( \frac{\hat{p}_{c|a}}{\hat{p}_{c}} \right),
\end{equation}
where $\hat{p}_{c|a} = \Pr(c \given a)$ is the maximum likelihood estimate of the conditional probability of the target $c$ given the subgroup $a$, and $\hat{p}_{c}$ is the marginal probability for that category.

\subsubsection{Weighted Kullback-Leibler for normal distributions} 
In the case of a univariate \emph{numeric} target $Y$, many distributions could be used for modelling. We resort to the normal distribution for its robustness and analytical properties, as mentioned before. Nonetheless, still two possibilities remain: a location distribution $\Ta = \{\mu_a\}$ that only accounts for the mean, or a `complete' normal distribution $\Ta = \{\mu_a, \sigma_a \}$ that accounts for the mean and the variance. With the location distribution $KL(\Ta; \Td)$ equals:
\begin{equation}\label{eq:wkl_mu}
KL_{\mu}(s) = \frac{(\hat{\mu}_d-\hat{\mu}_a)^2}{\hat{\sigma}_d},
\end{equation}
while with the normal distribution one obtains:
\begin{equation}\label{eq:wkl_mu_sigma}
KL_{\mu,\sigma}(s) =  \left [ \log\frac{\hsigd}{\hsiga}+ \frac{\hsiga^2+(\hmua-\hmud)^2}{2 \hsigd^2}\loge -\frac{\loge}{2} \right ].
\end{equation}
Note that since $\hat{\sigma}_d$ is a constant for each dataset, there is a strong resemblance between $WKL_{\mu}(s)$ and WRAcc, where the only difference is the square of the difference of the means. Also, notice that $WKL_{\mu,\sigma}$ directly takes penalises subgroups with large variance---a \emph{dispersion-aware} quality measure---while $WKL_{\mu}(s)$ (and also WRAcc) fail to give importance to the dispersion of subgroup values.

\subsection{Subgroup set discovery}\label{sec:back_ssd}

Subgroup set discovery (SSD) \citep{van2012diverse} is the task of finding a set of high-quality, non-redundant subgroups that together describe all substantial deviations in the target distribution. It can be seen as an instantiation of the LeGo (from \textbf{L}ocal Patt\textbf{e}rns to \textbf{G}lobal M\textbf{o}dels) framework, which describes the steps to pass from local descriptions of the data to a global model \citep{knobbe2008local}. LeGo identifies three phases of this process for SSD: 1) mining local candidate subgroups; 2) finding a compact set of the subgroups from the candidates found in 1; and 3) combining the interesting subgroups identified in 2 in one global model. Of course, not all phases need to happen in this order, and some stages can be combined.\\

Here, we are interested in phase 3: how to aggregate the subgroups in a global model.

There are three most common aggregation models for subgroups: 1) top-$k$ subgroups, which are just the best $k$ ranking subgroups according to a \emph{local} quality measure (sometimes called traditional SD); 2) subgroup list, a sequentially ordered set of subgroups; and 3) subgroup set, an unordered set of subgroups. Top-$k$ is still a local paradigm as it does not consider how those $k$ subgroups describe different regions of the data. Contrastingly, subgroup lists and sets are true global models as they consider both the subgroups' local coverage and their global coverage as sets.

Over the years, works for SSD focus got divided between lists \citep{lavravc2004subgroup,belfodil2019fssd} and sets \citep{lavravc2004subgroup, van2012diverse,bosc2018anytime,lijffijt2018subjectively}. For a detailed comparison between all these methods, please refer to Section~\ref{sec:work_ssd}. \\ 

\subsubsection{Defining the task of Subgroup Set Discovery}
Now, we will formally define the task of SSD from a global dataset perspective. SSD can be defined as, given a quality function $Q$ for subgroup sets and the set of all possible subgroup sets $\mathcal{S}$, the task is to find the subgroup set $S^* = \{s_1, \ldots ,s_k \}$ given by $S^* = \argmax_{S \in \mathcal{S}} Q(S)$. Note that $Q$ should take into account the individual quality of subgroups $q(a)$ and the overlap of their coverages $\Da$ and quantify the contribution of each instance only once. As opposed to top-$k$ mining where only their individual qualities are considered, i.e., $Q (S) = \sum q(a)$.

Ideally, a quality measure for subgroup sets $Q$ should: $1$) \emph{be global}, i.e., for a given dataset it should be possible to compare subgroup set qualities regardless of subgroup set size or coverage; $2$) \emph{maximise the individual qualities} of the subgroups; and $3$) \emph{minimise redundancy} of the subgroup set, i.e., the subgroups covers should overlap as little as possible while ensuring the previous point. Next, we formulate the subgroup list model class and propose a new global measure for subgroup lists.

\subsection{Beam search and separate and conquer algorithms in subgroup discovery}\label{sec:back_alg_sd}

To find good subgroup lists, we propose the SSD++ algorithm in Section~\ref{sec:SSDpp}. SSD++ is a heuristic based on the Separate-and-Conquer (SaC) \citep{furnkranz1999separate} strategy of iteratively adding the local best subgroup to the list, combined with beam search for candidate subgroup generation at each iteration level. 

For that reason, we now present the beam search algorithm in subgroup discovery and the SaC algorithm usually used in SSD. For an in-depth analysis of algorithms in SD and SSD check Section~\ref{sec:work_sd} and \ref{sec:work_ssd}, respectively.

Greedy approaches are often employed in SSD in general, and subgroup list discovery in particular, as the task of finding the optimal unordered or ordered set of patterns is NP-Hard. 

\paragraph{Beam search} is arguably the most common heuristic in subgroup discovery \citep{lavravc2004subgroup,meeng2011flexible,van2012diverse,meeng2020forreal}. It is a greedy hill-climbing approach that starts with candidate subgroups of size one and iteratively refine a subset of those to subgroups to a larger length by adding one more condition per iteration. Specifically, beam-search has three main hyperparameters: 1) the beam-width $w_b$; 2) the maximum search depth $d_{max}$; and 3) a quality measure. In its standard form, the process starts by finding all the best $w_b$ subgroups of size one, i.e., their description only includes one condition such as  $ x_1<5$ or $ x_2 = category$, according to the quality measure. Then, it refines all the $w_b$ size one subgroups by adding one more condition, selecting the $w_b$ best refinements, and discarding the rest according to the quality measure. The process continues until the maximum number of conditions $d_{max}$ is achieved, and the subgroup that maximises the quality measure is returned. There are variations such as the one used by \cite{van2012diverse} and \cite{meeng2020forreal} where the numeric explanatory variables are discretised in each refinement, also known as dynamic discretisation.

\paragraph{Separate and Conquer.} Most algorithms for Subgroup Set Discovery find their global models---subgroup lists or sets---sequentially by adding one subgroup at the time \citep{lavravc2004subgroup,van2012diverse,bosc2018anytime,lijffijt2018subjectively,belfodil2019fssd}, and mostly vary on how they remove the data or generate their candidate subgroups. Thus, all SSD approaches that sequentially add subgroups to a model can be seen as a variation of the traditional Separate-and-Conquer (SaC) rule learning strategy, defined by: 1) adding local best rule/subgroup to the model; 2) remove (traditional SaC) or re-weight (deviation from traditional SaC) the data covered by it; and 3) repeat process 1 and 2 until there is no data left to cover. Thus, using SaC for finding subgroup lists is the obvious choice. Depending if they find list or sets, in step 2 they either remove or re-weight, respectively. For details on each specific method please refer to Section~\ref{sec:work_ssd}.

\section{Preliminaries: Model selection in frequentist, Bayesian, and MDL perspective}\label{sec:back_model_sel}

This section briefly revisits model selection in different branches of statistics: classic statistics, also known as frequentist; Bayesian statistics; and using the Minimum Description Length (MDL) principle. For an in-depth comparison of these methods, please refer to Chapter $7$ of \cite{friedman2001elements}.

The objective of model selection is to find the best point hypothesis, i.e., choosing the best model (and its parameters set to specific values) for a dataset from a class of possible models. In our case, this translates to selecting the best subgroup list, such as in Figure~\ref{fig:zoo_example}, out of all the possible subgroup lists that can be constructed for that dataset. Furthermore, the best model should describe the data well while not overfitting, i.e., it should generalise its findings beyond the (training) data used to estimate its parameters. 

Many methods are reformulations of the principle often called Occam's Razor, i.e., select the simplest model that fits the data well. However, depending on the branch of statistics, the assumptions and notation differ, making it hard to compare the methods directly; thus, we attempt to present them in a more unified way. The reason for this is twofold: first, we aim to provide a gentle introduction to the MDL principle for the unacquainted reader by starting from more known branches of statistics; second, our proposed MDL formulation of robust subgroup discovery is related to concepts from other branches, such as model comparison with Bayesian factors or multiple hypothesis testing. \\

Note that we do not delve into the related Akaike Information Criterion (AIC) \citep{akaike1998information}. Although it is usually advantageous in predictive settings \citep{grunwald2019minimum}, the AIC has a higher rate of false positives and a bias towards more complex models than the Bayesion Information Criterion (BIC) \citep{rouder2009bayesian} and, consequently, our MDL formulation (as it asymptotically converges to BIC up to a constant; see Appendix~\ref{appendix:BIC}). We want to avoid these properties when mining statistically robust subgroups.\\

This section is divided as follows. First, in Section~\ref{sec:back_frequentist}, we present frequentist approaches as they are the most commonly employed. Then, Section~\ref{sec:back_bayes} discusses Bayesian hypothesis testing. After that, Section~\ref{sec:back_mdl} introduces the basic principles of MDL for model selection. Finally, Section~\ref{sec:back_multiple_hyp} looks into the concept of multiple hypothesis testing from the three different perspectives.

\subsection{Frequentist approach to model selection}\label{sec:back_frequentist}
From a classical statistics perspective, one can use several methods to select the best model from a set of models. These can be broadly divided into out-data and in-data methods, corresponding to testing the models on an external or the same (internal) data source on which they were estimated (trained), respectively. To the first category belong methods commonly used in machine learning, such as cross-validation \citep{friedman2001elements}. The second category, which concerns us most, corresponds to structural measures, which have additional terms to penalise the complexity of the model, i.e., how general is a model class constrained to a certain number of parameters and dataset size. Model complexity is sometimes also interpreted as the effective number of parameters. For example, imagine two classes of subgroup lists: subgroup lists with one subgroup; and subgroup lists with up to three subgroups (same number of conditions and dataset); it is easy to see that the second subgroup list includes the first class and can potentially divide the data in more ways. Common examples of model complexity measures are the L1 and L2 norms and the Vapnik–Chervonenkis (VC) dimension \citep{vapnik2015uniform}.

We do not directly mention common `performance' measures such as accuracy or maximum likelihood. In fact, from an in-data testing perspective and for a nested model class such as subgroup lists, e.g., the class of all subgroup lists with two subgroups includes the class of all subgroup lists with one subgroup, these measures will overfit on the data \citep{grunwald2007minimum}. 

\paragraph{Structural measures.} Ideally, the method chosen should give guarantees on the model performance on unseen data. The risk minimisation principle guarantees those by using the VC dimension for model complexity. The main idea is to select the model that maximises a performance measure, e.g. accuracy, while having the smallest VC dimension. Informally, the VC dimension of a model is given by the largest set of data examples it (e.g., the class of subgroup list with $3$ subgroups of one condition) can separate. 
In practice, computing the VC dimension can be impractical for certain model classes, and one resorts to more straightforward model complexities, such as the L1 or L2 norm.

An example of an L1 norm structural measure for finding decision lists was described by \cite{angelino2017learning} as
\begin{equation}
    Q(D,M) = Acc.(Y \given \Xmat, M) + \lambda |M|,
\end{equation}
where $Q(D,M)$ is the objective function used to select the model, $Acc.(Y \given \Xmat, M)$ is the accuracy, $\lambda$ is an adjustable parameter, and $|M|$ is the number of rules in the model. For completeness, the method should include an extra term to penalise the number of conditions in each rule. 

Bayesian statistics and the MDL principle can be seen as probabilistic structural measures with mathematically rigorous foundations, similarly to the structural risk minimisation, that take into account all model parameters to quantify model complexity \citep{grunwald2019minimum}. The main difference between these branches of statistics is that frequentist statistics usually focus on the probability of the data given the model $\Pr(D \given M)$. In contrast, Bayes and MDL focus on $\Pr(M \given D)$ by making additional assumptions about the probability of the model before seeing any data $\Pr(M)$ (prior)---similar, albeit different, to the additional assumptions made by the VC dimension.

\subsection{Bayes hypothesis testing and Bayes factor}\label{sec:back_bayes}

Bayesian hypothesis testing was introduced by Jeffreys \citep{jeffreys1935some,jeffreys1998theory}, and focuses on comparing the hypothesis/models based on their probability of occurrence given the data, i.e., based on the posterior probability of each model $\Pr(M \given D)$. To compare both models, one computes the ratio of their posterior distributions such as \citep{kass1995bayes,rouder2009bayesian}:
\begin{equation}
 \Xi =  \frac{\Pr (M_1 \given D)}{\Pr (M_2 \given D)} = \frac{\Pr (D \given M_1)}{\Pr (D \given M_2)}  \times \frac{\Pr (M_1)}{\Pr (M_2)} = K_{1,2} \times \frac{\Pr (M_1)}{\Pr (M_2)},
\end{equation}
where the transition from the second to the third expression is made using the Bayes rule and removing the terms $\Pr(D)$ from the expression, and $K_{1,2}$ is the Bayes factor between the models. Indeed, depending on how large or how small the ratio $K_{1,2}$ is, we can interpret it---similar to p-values in frequentist statistics---as more evidence in favour of hypothesis $1$ or $2$, respectively \citep{kass1995bayes}.  It is interesting to notice that, in the case of a large number of instances and a smooth prior, the Bayes factor approximates the Bayes Information Criterion (BIC) up to a constant \citep{schwarz1978estimating,raftery1995bayesian}.

As the two hypotheses being tested are over the same data, the ratio $\Xi$ can be rewritten as $\Pr(M_1,D)/\Pr(M_2,D)$ using the chain rule. Also, as we deal with a supervised setting, $\Pr (D \given M)$ becomes $\Pr (\Ymat \given \Xmat, M)$, taking the same form used in the next section for the MDL principle. When this comparison is extended to a whole model class $\mathcal{M}$, several methods can be used to select the best model; however, the one most similar to our MDL approach corresponds to choosing the model with the highest probability. i.e., the mode of the distribution.

\subsection{MDL-based model selection}\label{sec:back_mdl}

The Minimum Description Length (MDL) principle originates from the ideas of information theory. It allows for model selection by comparing code lengths for different models and selecting the model that compresses the data best\citep{rissanen78,grunwald2007minimum,grunwald2019minimum}. In our specific case, the goal is to find the best subgroup list model $M$ from the class of all possible subgroup list models $\mathcal{M}$, as defined in Section~\ref{sec:model_class}. As we want to find the best point-hypothesis, such as in Figure~\ref{fig:car_example}, the model selection problem should be formalised using a two-part code \citep{grunwald2007minimum}, i.e., 
\begin{equation}\label{eq:LengthTotal}
\begin{split}
M^* = \argmin_{M \in \M} L(D,M) &= \argmin_{M \in \M} \left[ L(D \given M)  + L(M) \right]\\
&=\argmin_{M \in \M} \left[ L(\Ymat \given \Xmat,M)  + L(M) \right],
\end{split}
\end{equation}
where $L(\Ymat \given \Xmat,M)$ is the encoded length, in bits\footnote{To obtain code lengths in bits, all logarithms in this paper are to the base 2.}, of target variables data  $\Ymat$ given explanatory data $\Xmat$ and model $M$, $L(M)$ is the encoded length, in bits, of the model, and $L(D,M)$ is the total encoded length and the sum of both terms. Note that the data encoding changes from $L(D  \given M)$ to $L(\Ymat \given \Xmat,M)$ to reflect our supervised setting and how we are only concerned with encoding the target variables $\Ymat$ in the goodness of fit part of MDL. Intuitively, the best model $M^*$ is the model that results in the best trade-off between how well the model compresses the target data and the complexity of that model---thus minimising redundancy and automatically selecting the best subgroup list size. 

\paragraph{MDL and probabilities.} Although the MDL principle measures the length of encodings in bits, every encoding can be translated to probabilities by the Shannon-Fano code \citep{shannon1948mathematical}: 
\begin{equation}\label{eq:shannonfano}
    L(A) = - \log \Pr(A),
\end{equation}
where $A$ is an event---in our case, it can be the model $M$, the target variables $\Ymat$, or any of their subparts defined in Section~\ref{sec:MDL} and \ref{sec:data_encoding}---and $\Pr(A)$ its probability. Thus, each code length in MDL can be directly interpreted as the negative logarithm of a probability. Consequently, the model with the smallest total encoded length $L(D,M)$ is that one having the largest probability $\Pr(D,M)$.

\subsection{Multiple-hypothesis testing}\label{sec:back_multiple_hyp}
Multiple-hypothesis testing is the task of testing more than one hypothesis\footnote{In this manuscript, each subgroup list model forms a hypothesis.} on the same data \citep{shaffer1995multiple}. To avoid increasing the Type I error rate, i.e., selecting models that are false discoveries, we should compensate for the fact that we test multiple hypotheses. For clarity, note that our focus here is on multiple-hypothesis testing to select a single best model from a class of models for one dataset, not selecting the best algorithm over multiple datasets. For an in-depth explanation of the latter, please refer to \cite{demvsar2006statistical}.\\

In \emph{frequentist} statistics, one usually accepts a model based on a p-value and an assigned significance level $\alpha$. Suppose you have chosen $\alpha$ equal to $0.05$ and you are testing only two hypotheses; then, the $\alpha$ value tells us that by chance, you have a $0.05$ probability of having evidence in favour of one model when it is false and should be rejected. If we would instead test $100$ different models (and these are considered independently), it is easy to see that we have a chance of at least $5$ models being acceptable when they should not have been. One way to counteract this effect is by adjusting the significance level to accommodate the number of hypotheses tested. Several methods exist to that end, and the conceptually more simple is the Bonferroni correction, where one divides $\alpha$ by the number of hypotheses/models being tested.

In the case of \emph{Bayesian statistics} and \emph{the MDL principle}, this translates to using $\Pr(M)$ and $L(M)$, respectively, to account for all possible models that are tested as hypotheses. In Section~\ref{sec:model_encoding} we will show how to interpret $L(M)$ as a correction for multiple-hypothesis testing in the context of subgroup lists.

%
%

\section{Problem statement: robust subgroup discovery}\label{sec:sec_problem_statement}

This section formally introduces the problem we propose to solve: robust subgroup discovery. Informally, the problem can be described as:

\begin{quote}
Find the \emph{globally} optimal set or list (i.e., an ordered set) of non-redundant and statistically robust subgroups that together explain relevant \emph{local} deviations in the data with respect to specified target variables.
\end{quote}

As this is a broad problem, we need to narrow it down: we only deal with \emph{subgroup lists}, define the optimal list using the MDL principle, and propose an iterative greedy algorithm that guarantees that each subgroup added to the list is statistically robust. Our MDL-based formulation of subgroup lists includes the \emph{MDL-based top-$k$ subgroup discovery} problem (Section~\ref{sec:sd_proof}), which gives a direct relationship to subgroup discovery as it was originally introduced. \\

To formally state our problem, we first need to introduce the subgroup list model class, which we will do in Section~\ref{sec:model_class}. Then, based on this model, we provide our problem statement in Section~\ref{sec:problem_statement}. Finally, in Section~\ref{sec:back_swkl} we propose a quality measure that quantifies the goodness of fit of subgroup lists.  

\subsection{Subgroup list model class}\label{sec:model_class}


\emph{Subgroup lists} are a sequentially ordered set of subgroups; see Figure~\ref{fig:rule_list}. Given its ordered format, a subgroup list always partitions the data, i.e., each instance of data is covered by one and only one subgroup (or the default rule). For example, if a subgroup list contains $4$ subgroups, the dataset will be partitioned into $4+1$ parts, one for each subgroup plus one for the dataset/default rule. A subgroup down the list, such as the second subgroup, should be interpreted as: the second subgroup is active only when its description is active and the description of the first subgroup \emph{is not} active.

More specifically, as we are only interested in finding subgroups for which the target deviates from the overall distribution, we assume $\Ymat$ values distributed according to $\Td$ by default (last line in Figure~\ref{fig:rule_list}). Thus, for each subset in the data where the target distribution deviates from $\Td$  and a description exists, a subgroup specifying a different distribution $\Ta$ could be added to the list. 
Ordering the rules formed by subgroups $S = \{s_1,\cdots, s_\omega\}$ and adding the dataset rule at the end (default rule) leads to a subgroup list $M$ of the form of Figure~\ref{fig:rule_list}.

Regarding the possible distributions $Dist$, we use \emph{categorical} distributions for the nominal targets, i.e., $Dist ~ Cat(\hat{p}_{1}, \cdots, \hat{p}_{k})$, or \emph{normal} distributions for the numeric target case, i.e.,  $Dist ~ \mathcal{N}(\hat{\mu}_i,\hat{\sigma}_i)$.

The categorical distribution is a natural choice for describing the probabilities of classes \citep{letham2015interpretable}. For numeric targets, several distributions can be selected; however, the \emph{normal} distribution captures two properties of interest in numeric variables, i.e., centre and spread, while being robust to cases where the data violates the normality assumption. Also, it allows for a closed-form solution from a Bayesian \citep{jeffreys1998theory} and MDL \citep{grunwald2007minimum} perspective. 
For an analysis of the direct use of the numeric empirical distribution in subgroup discover, please refer to \cite{meeng2020uni}.

\begin{figure}[b!]
	\centering
	\ra{1.1}\normalsize
	\setlength{\tabcolsep}{5.1pt}
	\begin{tabular}{@{}llrrrrr@{}}
		$s_1$:	&	IF	&	$a_1 \sqsubseteq  \x$		&	THEN	&	$y_1 \sim Dist(\hat{\Theta}_{1}^{1})$	&$\cdots$&	$y_t \sim Dist(\hat{\Theta}_{t}^{1}) $	\\
		&	 	&	$\vdots $		&		&	&		&		\\
		$s_\omega$:	&	ELSE IF	&	$a_\omega \sqsubseteq  \x$		&	THEN	&	$y_1 \sim Dist(\hat{\Theta}_{1}^{\omega}) $	&$\cdots$&	$y_t \sim Dist(\hat{\Theta}_{t}^{\omega})$	\\
		dataset:	&	ELSE	&			&		&	$y_1 \sim Dist(\hat{\Theta}_{1}^{d}) $	&$\cdots$&	$y_t \sim Dist(\hat{\Theta}_{t}^{d})$	\\
	\end{tabular}\vspace{0.3cm}\caption{Generic subgroup list model $M$ with $\omega$ subgroups $S= \{s_1,...,s_\omega\}$ and $t$ (number of target variables) distributions per subgroup}\label{fig:rule_list}	
\end{figure}	

\subsubsection{Subgroup lists versus subgroup sets} 
While we formulate our theory solely for subgroup lists, each global model has advantages and disadvantages. On the one hand, subgroup lists allow for a sequential interpretation of the subgroups, generally in decreasing order of their importance. Moreover, each instance in the data is associated with only one subgroup. On the other hand, subgroup sets allow for a semi-independent interpretation of each subgroup and can be considered a more general framework. These properties tend to make sets more interpretable when looking at all the subgroups. At the same time, lists are usually more interpretable from an instance perspective---as each instance is only covered once---and on the contribution of each subgroup to the global model. 

However, as the number of subgroups in a model increases, both types of models become harder to interpret. In the case of subgroup lists, one must inspect the covering subgroup and all the preceding ones for each instance. In the case of subgroup sets, there can be a considerable overlap for each instance, making it hard to assess the individual contribution of each subgroup.
 
In addition to being one of the first subgroup set discovery models, the main advantage of selecting subgroup lists, in our case, is their property of unambiguously partitioning the data into non-overlapping parts. This property allows us to use the MDL principle to formulate the robust subgroup discovery problem for subgroup lists optimally.
 

\subsubsection{Difference between subgroup lists and (predictive) rule lists} 
A subgroup list defined above corresponds to a probabilistic rule list with $\omega=|S|$ rules and a last (default) rule fixed to the overall empirical distributions for each target variable \citep{proencca2020interpretable}. Fixing this last `rule' distribution is crucial and differentiates a subgroup list from a rule list as used in classification and/or regression \citep{manuel2021robust}, as this enforces the discovery of a set of subgroups whose individual target distributions all substantially deviate from the overall target distribution (dataset rule). It is shown in Section~\ref{sec:sd_proof} that the objective of finding a subgroup list with this format is equivalent to top-$k$ subgroup discovery when finding subgroup lists with just one subgroup. A theoretical comparison of the difference between the objectives of predictive rule lists and subgroup lists from an MDL-based perspective is given in Appendix~\ref{appendix:proof_sd_vs_prediction}.

\subsection{Formal problem statement}\label{sec:problem_statement}

Let $D$ be a dataset consisting of explanatory variables data $\Xmat$ and target variables data $\Ymat$, i.e., $D = \{\Xmat, \Ymat \}$. 
Let $M \in \mathcal{M}(D)$ be all possible subgroup lists for $D$ formed by an ordered set of subgroups $S$ and a dataset rule, as in Figure~\ref{fig:rule_list}. Let $s\in \Omega(D)$ be all possible subgroups in $D$ with respect to all possible descriptions in $\Xmat$, which are formed by conjunctions of conditions (pattern language) on the possible values $\X$ of the explanatory variables $X \in \Xmat$. 
The conditions vary by variable type $X$ and can be an interval over the reals $\mathbb{R}$ for numeric, e.g., $consumption-city \in [0,8]$ in Figure~\ref{fig:car_example}, or equality for nominal or Boolean, e.g. $weight = light$ in Figure~\ref{fig:car_example}. The target description of each subgroup is restricted to Categorical or Normal distributions.

Given the definition of all possible subgroup lists in a dataset, the objective is to return the subgroup list $M$ that minimises the MDL two-part code of Eq.~\eqref{eq:LengthTotal}, i.e., the objective is to find

\begin{equation}\label{eq:problem_statement}
\begin{split}
M^* =& \argmin_{M \in \M} \left[ L(\Ymat \given \Xmat,M)  + L(M) \right], \\
& s.t. \;  \forall_{s_i \in M} \Xi (\Ti,\Td) > 0,
\end{split}
\end{equation}

where the first part concerns the \emph{global} optimality of the subgroup list, on the complete dataset, while the constraint $\Xi (\Ti,\Td)$ states that there should be more evidence in favour of having each subgroup in the list than for using the overall dataset distribution (for the motivation on using the MDL principle refer to Section~\ref{sec:motivation_mdl}). 

To operationalise this, $L(\Ymat \given \Xmat,M)$ and $L(M) $ need to be defined. Thus, we propose a model encoding $L(M)$ in Section~\ref{sec:model_encoding} and a data encoding in Section~\ref{sec:data_encoding}. Then, Section~\ref{sec:mdlsl_wklsd_bayes} shows that the data encoding equals WKL-based subgroup discovery and Bayesian testing, reflecting the statistical robustness of each subgroup in the list. 

Finally, as finding the optimal subgroup list according to this formulation is unfeasible for most real-world problems, Section~\ref{sec:SSDpp} proposes SSD++, a heuristic algorithm (for NP-Hard intuition, check Section~\ref{sec:nphard}). Moreover, this algorithm approximates the MDL minimisation by restricting the search space of possible subgroups and subgroup lists. Nonetheless, it guarantees that the most statistically robust subgroup found by beam search is added to the subgroup list. Indeed, the greedily adding a subgroup automatically accounts for the statistical constraint of Eq~\eqref{eq:problem_statement}, i.e., $\Delta_{\beta} L(D,M \oplus s) = \Xi (\Ti,\Td) > 0$. \\

\subsubsection{Motivation for the MDL principle}\label{sec:motivation_mdl}
The MDL principle is used due to its statistical robustness and objectivity when compared to other approaches (for a short introduction to model selection, please refer to Section~\ref{sec:back_model_sel}). First, it does not assume that the model that generated the data belongs to the model class used. Second, it allows the data encoding, i.e., $L(\Ymat \given \Ymat,M) = -\log \Pr(\Ymat \given \Ymat,M)$ to be computed with several optimal methods, including the one in Bayesian statistics---Bayes updating rule \citep{grunwald2019minimum}. Third, from a model encoding perspective, i.e., $L(M)= -\log \Pr(M)$, which is the most subjective part of the encoding, the MDL principle recommends choosing an encoding that uses the minimum number of assumptions, similarly to the max entropy principle \citep{jaynes1957information}.\\

\subsubsection{Finding optimal subgroup lists is NP-Hard}\label{sec:nphard}
The general task of finding an ordered set of patterns \citep{mielikainen2003pattern} and that of finding the smallest decision list for a dataset $D$ \citep{rivest1987learning} are both NP-hard problems. Thus, it is trivial to see that the problem of finding a subgroup list,  which is a probabilistic generalisation of the decision list, is NP-hard. \\

\subsection{A new measure for subgroup lists: the sum of WKL divergences}\label{sec:back_swkl}

This section extends the WKL divergence to subgroup lists, allowing us to compare the quality of different algorithms that mine subgroup lists. Also, it is shown in Section~\ref{sec:sd_proof} to correspond to one part of the data encoding of our MDL formulation.

Following the introduction of quality measures in Section~\ref{sec:back_sdqm} and subgroup lists in the previous sections, we can extend the KL-based measure of Eq.~\eqref{eq:wkl} for individual subgroups to measure subgroup lists. That is, we propose the \emph{Sum of Weighted Kullback-Leibler divergences} (SWKL), which can be interpreted as the sum of weighted KL divergences for the individual subgroups:
\begin{equation}\label{eq:swkl}
\textrm{SWKL} (S)= \frac{\sum_{i = 1 }^{\omega} n_i KL(\Tij;\Tdj)}{|D|},
\end{equation}

where $i$ is the subgroup index in a subgroup list, $\omega$ is the number of subgroups in $S$, and $|D|$ is the number of instances in $D$. The latter is used to normalise the measure and compare values across datasets. In the case of multiple target variables, the normalisation could also include the number of targets, but we do not use this in this work. 
The SWKL measure assumes that the data is partitioned per subgroup and is based on the assumption that subgroups can be interpreted sequentially as a list. 

An advantage of the SWKL measure is that it can be used for any target variable(s), as long as probabilistic models are used. Note that computing SWKL is straightforward for subgroup lists, but not for subgroup \emph{sets} as multiple subgroups can cover an instance. For subgroup sets, it would be necessary to explicitly define the type of probabilistic overlap, e.g., additive or multiplicative mixtures of the individual subgroup models.

\paragraph{Overfitting.} It should be noted that this measure only quantifies how well a list of subgroups captures the deviations in a given dataset and is prone to overfitting: the higher the number of subgroups, the easier it is to obtain a higher value as there is no penalty for the number of subgroups (or their complexities, for that matter). As such, SWKL can be seen as a measure of `goodness of fit' for subgroup lists. This is not an issue for our approach as our MDL-based criterion naturally penalises for multiple hypothesis testing and the complexity of the individual subgroups, which is empirically validated by the statistical robustness analysis Section~\ref{sec:statistica_robustness}. Moreover, overfitting does not seem to be an issue in our empirical comparisons with other algorithms of Section~\ref{sec:empirical_nominal} and \ref{sec:empirical_numeric}, as the number of subgroups found was similar for most algorithms, rendering the subgroup lists comparable based on SWKL. 
%
%
%
%
%
%

%
%

\section{Model encoding of subgroup lists}\label{sec:MDL}

We presented the MDL principle in its generality in Section~\ref{sec:back_mdl} and the specific problem statement of finding optimal subgroup lists in Section~\ref{sec:problem_statement}. In this section, we define the model encoding $L(M)$ of subgroup lists and its relationship to multiple hypothesis testing.

\subsection{Model Encoding}\label{sec:model_encoding}

Following the MDL principle \citep{grunwald2007minimum}, we need to ensure that 1) all models in the model class, i.e., all subgroup lists for a given dataset, can be distinguished; and 2) larger code lengths are assigned to more complex models. To accomplish the former we encode all elements of a model that can change, while for the latter we resort to two different codes: when a larger value represents a larger complexity we use the universal code for integers \citep{rissanen1983universal}, denoted\footnote{$\LN(i)= \log k_0 + \log^{\ast} i $, where $\log^{\ast} i = \log i + \log \log i + \ldots$ and $ k_0 \approx 2.865064$.} $\LN$, and when we have no prior knowledge but need to encode an element from a set we choose the uniform code.

Specifically, the encoded length of a model $M$ over variables in $\Xmat$ is given by 
\begin{equation} \label{eq:LModel}
L(M) = L_\mathbb{N}(|S|) + \sum_{a_i \in S} \left[L_\mathbb{N}(|a_i|) +\log \binom{m}{|a_i|} + \sum_{v \in a_i} L(v)\right] ,
\end{equation}
where we first encode the number of subgroups $|S|$ using the universal code for integers, and then encode each subgroup description individually. For each description, first the number $|a_i|$ of variables used is encoded, then the set of variables using a uniform code over the set of all possible combinations of $|a_i|$ from all explanatory variables, and finally the specific condition for a given variable. As we allow variables of two types, the latter is further specified by

\begin{equation}
L(v) =\left\{\begin{matrix}
\log |\mathcal{X}_v| & \text{if } v \text{ is nominal}\\ 
L_{\mathbb{N}|2}(|n_{op}|) + \log N(n_{op},n_{cut}) & \text{if } v \text{ is numeric}
\end{matrix}\right.
\end{equation}

where the code for each variable type assigns code lengths proportional to the number of possible parts the variable's domain can partition the dataset. Note that this seems justified, as more parts imply more potential spurious associations with the target that we would like to avoid. For \texttt{nominal} variables this is given by the size of the domain, i.e., the number of categories in a nominal variable. For \texttt{numeric} variables it equals the number of operators used $L_{\mathbb{N}|2}(|n_{op}|)$\footnote{$L_{\mathbb{N}|2}$ is the universal code for integers with codes restricted to $n=1$ or $2$. This can be obtained by applying the maximum entropy principle to $L_{\mathbb{N}}$ when it is known that it cannot take values of $n > 2$.} plus the possible number of outcomes $N(n_{op}, n_{cut})$ given the operators and $n_{cut}$ cut points. The number of operators for numeric variables can be one or two, as there can be conditions with one (e.g., $ x \leq 2$) or two operators (e.g., $1 \leq x \leq 2$), which is a function of the number of possible subsets generated by $n_{cut}$ cut points. Note that we here assume that equal frequency binning is used, which means that knowing $X$ and $n_{cut}$ is sufficient to determine the cut points.

In the case of \emph{top-$k$ subgroup discovery}, i.e., only be interested in the top individual subgroups, the model becomes a subgroup list with only one subgroup. Thus, the model encoding of Eq.~\eqref{eq:LModel} remains the same except for the first term ($L_\mathbb{N}(|S|)$), which should be removed. This is because it is not required to account for a subgroup list with more than one subgroup.

\paragraph{Example 5 (continuation):} Let us assume that the subgroup list of the \emph{Automobile} example of Figure~\ref{fig:car_example} is composed of only the first subgroup. In that case the list only has one subgroup with description: \{weight = heavy  \& consumption-city $ \leq 8 $ km/L \}. Taking into account that the dataset has $17$ variables, $|\mathcal{X}_{weight}| = 3$ and only $3$ cut-points were used for numeric attributes, the expression of the model length is given by:
\begin{equation*}\begin{split}
L(M) &= L_\mathbb{N}(1) + L_\mathbb{N}(2) +\log \binom{17}{2} + \log |\mathcal{X}_{weight}| + \left[L_{\mathbb{N}|2}(1) + \log 2n_{cut} \right]\\
&= 1.52 + 2.52 + 7.09 + 1.59 + 0.77 + 2.59 \\
& = 16.08 \: \text{bits}
\end{split} 
\end{equation*}

It is important to note that the length of the model can (and should) be a real number, as we are only concerned with the idea of compression, not with materialising and transmitting the actually encoded data \citep{grunwald2007minimum}.

\subsection{Multiple-hypothesis testing and model length.}\label{sec:model_multiple}
As presented in the previous section, the model encoding $L(M)$ must be able to distinguish all possible models that could be learned for a dataset $D$. In fact, we are counting all possible models and then giving them a probability of occurring---with a smaller probability to models with fewer terms. 
Looking at Eq.~\eqref{eq:LModel} in particular, we can see that each term counts different parts of the model: 1) $L_\mathbb{N}(|S|)$ counts the possible number of subgroups in subgroup lists; 2) $L_\mathbb{N}(|a_i|)$ counts the possible length of a subgroup description; 3) $\log \binom{m}{|a_i|}$ counts the possible pairs of descriptions of size $|a_i|$; and 4) $L(v)$ counts the possible values a variable can have. Together, these four terms count all possible subgroup lists. In some cases, we are being extra conservative and counting more models that do not exist in a dataset, e.g., by using $L_\mathbb{N}(|S|)$ for the number of subgroups we allow for subgroup lists with infinite subgroups---above the number of instances in the data. Thus, the MDL principle is actively trying to avoid false positive or Type I error models by penalising all the models that could be learned and compared to each other. 

%
%

\section{Data encoding of target variables}\label{sec:data_encoding}

When the model is defined, what remains is to define the length function of the target data given the explanatory data and model, $L(\Ymat \given \Xmat,M)$. In this section, we show how to encode the target data $\Ymat$ by dividing it into smaller subsets that can be encoded individually and then summed together, and why there are different types of data encoding for each of the subsets. The specifics of encoding nominal and numeric targets are described in Sections~\ref{sec:nominal_data} and \ref{sec:numeric_data}, respectively.

\paragraph{Cover of a subgroup in a subgroup list. }First, we observe that for any given subgroup list of the form of Figure~\ref{fig:rule_list}, \emph{any individual instance $(\x^i,\y^i)$ can only be `covered' by one subgroup}. That is, the cover of a subgroup $a_i$, denoted $\Da$, depends on the order of the list and is given by the instances where its description occurs minus those instances covered by previous subgroups:
\begin{equation}\label{eq:coversubgroup}
\Di =\{\Xii,\Ymati \} = \{(\x,\y) \in D \given a_i \sqsubseteq \x \wedge \left ( \bigwedge_{\forall_{i'<i}}  a_{i'} \not \sqsubseteq \vec{x}   \right ) \}.
\end{equation}

Next, let $n_i =|\Di|$ be the number of instances covered by a subgroup (also known as \emph{usage}). In case an instance $(\x^i,\y^i)$ is not covered by any subgroup $s \in S$ \emph{then it is `covered' by the default rule}. The instances covered by the default rule $\Dd$ are the ones not covered by any subgroup (hence the name default rule) and formally defined as:

\begin{equation}\label{eq:coverdefault}
\Dd = \{\Xd,\Ymatd \} = \{(\x,\y) \in D \given \forall_{a_i \in M} a_i \not \sqsubseteq \x \}.
\end{equation}

Now, given that the subsets for each subgroup or default rule and each target variable are well-defined, one can---for each of the rules and targets---estimate the parameters of its probabilistic distribution using the maximum likelihood estimator.

Note that this shows us that a subgroup $s_i \in M$ is fully defined by its description $a_i$ in a dataset $D$, and we will interchangeably refer to the subgroup by its description and to its elements (statistics, parameters, distributions, etc.) by its index $i$ when obvious from context. 

As the subgroup list induces a \emph{partition of the data}, the total length of the encoded data can be given by the sum of its \emph{non-overlapping parts}:
\begin{equation}\label{eq:encode_parition}
L(\Ymat \given \Xmat, M) = L(\Ymatd  \given \boldsymbol{\Theta}^d ) + \sum_{s_i \in S} L(\Ymati), 
\end{equation}
where $\boldsymbol{\Theta}^d $ is the vector of parameters for each variable $ \Theta^d_1,\ldots,\Theta_t^d $. Observe that we dropped $\Xmata$ as these are not necessary to encode $\Ymata$ but only to generate the partition of the data, and also dropped the parameters $\boldsymbol{\Theta}^i$ of the subgroups as we do not know what are their parameters until we see the data. This last part will be clarified in the next paragraphs, where we describe how to encode subsets without knowing the parameters.

As a side-note, note that Eq.~\eqref{eq:encode_parition} concerns the encoding of any supervised partition of the data, which allows to directly quantify the quality of any tree learning method---each such tree induces a partition of the data.

\paragraph{Encoding data of $t$ (assumed) independent target variables. } As each target variable is assumed independent from each other the encoding of target data is given by the sum of their individual encodings: 
\begin{equation}\label{eq:encode_independent}
L(\Ymat \given \Xmat, M) = - \log \left( \prod_{j=1}^{t} \Pr (\Yj \given \Xmat, M) \right) = \sum_{j=1}^{t} L(Y_j \given \Xmat, M).
\end{equation}

Integrating \eqref{eq:encode_parition} and \eqref{eq:encode_independent}, one obtains:
\begin{equation}\label{eq:data_encoding}
L(\Ymat \given \Xmat, M) = \sum_{j=1}^{t} \left(  L(\Ydj  \given \Theta^d_j ) + \sum_{s_i \in S} L(\Yij) \right)
\end{equation}

\paragraph{Two types of data encoding:} data encoding can be separated in two different categories: $1$) with \emph{known parameters}; and $2$) with \emph{unknown parameters}. \\

\emph{$1$) Known parameters}: when the parameters of a distribution are \emph{known}, one can encode the data points directly using the probability for those points given by the distribution with the known parameters. Thus, the encoding of points $Y_j^i$ ($j^{th}$ variable and $i^{th}$ subgroup) is equal to the negative logarithm of their probability given by known parameters $\Tij$: 
\begin{equation}\label{eq:known_encoding}
L(\Yij \given \Tij) = \sum_{y \in \Yij} - \log \Pr(y \given  \Tij).
\end{equation}
This type of code is used in the case of the default rule of a subgroup list, as the parameters $\Tdj$ are equal to the marginal distribution of variable $\Yj$ and are constant for each dataset. Note that this is the \emph{key difference between a subgroup list and a predictive rule list}: the last rule of a subgroup list is fixed to the marginal distribution, while in the (predictive) rule list its parameters are unknown and depend on the subset $D^d$.\\

\emph{$2$) Unknown parameters}: when the parameters are \emph{unknown} we need to encode both the parameter values and the data points. We have two possibilities: 1) crude MDL, i.e., encoding the probabilities using a suboptimal probability distribution and then applying the Shannon-Fano code, i.e., the logarithm of the empirical probability \citep{shannon1948mathematical}; or 2) employ an optimal encoding of both parameters of the distribution and data points together \citep{grunwald2019minimum}. In this work, we employ optimal encoding of parameters, as it guarantees optimality in the sense that the encoding is the best possible in the worst-case scenario, i.e., in case the sample of the data is not representative of the population. For our problem, three main types of optimal encodings exist, which are, in increasing order of optimality guarantees: $1$) \emph{prequential plug-in}; $2$) \emph{Bayesian}; $3$) \emph{Normalised Maximum Likelihood (NML)}. While the first two are asymptotically optimal, the NML encoding is optimal for fixed sample sizes. 

Depending on the target type, we employ the best encoding possible while being computationally feasible, i.e., we require adequate run-time for our algorithm. 
For nominal targets, we present an NML encoding for both the probabilities of each class and the data points in Section~\ref{sec:nominal_data}, which is a theoretical improvement over the prequential plug-in code that was recently proposed for classification rule lists by \cite{proencca2020interpretable}. For numeric targets, we resort to a Bayesian encoding, as recently proposed by \cite{proencca2020discovering}, as the NML code is not computationally feasible for that case.


\subsection{\textbf{Data encoding: nominal target variables}}\label{sec:nominal_data}

When the data have one or more nominal targets, the distributions in the probabilistic rules \eqref{eq:prob_rule} are categorical distributions $Cat(\Theta)$, each with a set of parameters $\Theta = \{ p_1,\cdots, p_k \} $ representing the $k$ classes:
\begin{equation}
\Pr( y = c \given p_1,\cdots, p_k ) = p_c, \text{ subject to } \sum_{c=1}^{k} p_c = 1.
\end{equation}
This implies a subgroup of the form:
\begin{equation*}\label{eq:prob_rule_nominal}
a \mapsto y_1 \sim Cat(p_1,\cdots, p_k ), \cdots, y_t \sim Cat(p_{1'},\cdots, p_{k'}),
\end{equation*}
where $k$ and $k'$ are the number of classes $Y_1$ and $Y_t$, respectively. To simplify the introduction of concepts we will assume we only have one target variable in $\Ymat$, and then generalise the results to multiple variables at the end. Thus, throughout this section $\Ymat$ becomes $Y$, and the parameters of each subgroup $s_i$ become $\Ti = \{ p_{1|i},\cdots, p_{k|i} \}$ as there is only one variable with $k$ classes, where $p_{1|i}$ is the probability of class $1$ for subgroup $i$, i.e., $\Pr(c=1\given a_i)$. The general form of a subgroup list with one nominal target takes the form of Figure~\ref{fig:subgroup_list_nominal}.\\
\begin{figure}[ht]															
	\centering															
	\ra{1.1}\normalsize														
	\begin{tabular}{@{}rlrrr@{}}
		$s_1$:	&	IF	&	$a_1 \sqsubseteq  \x$		&	THEN	&	$y \sim Cat(\hat{p}_{1|1},\cdots, \hat{p}_{k|1})$					\\
	&	 	&	$\vdots  $		&		&				\\
		$s_\omega$:	&	ELSE IF	&	$a_\omega \sqsubseteq  \x$		&	THEN	&	$y \sim Cat(\hat{p}_{1|\omega},\cdots, \hat{p}_{k|\omega})$					\\ 
		dataset:	&	ELSE	&			&		&	$y \sim Cat(\hat{p}_{1|d},\cdots, \hat{p}_{k|d})$					\\
	\end{tabular}\vspace{0.3cm}\caption{Generic subgroup list model $M$ with $\omega$ subgroups $S= \{s_1,...,s_\omega\}$ and a single nominal target $Y$ with $k$ categories}\label{fig:subgroup_list_nominal}															
\end{figure}																

In the following sections, we will derive the data encoding for subgroup lists with categorical distributions. First, in Section~\ref{sec:ML_nominal} we introduce the maximum likelihood estimators that will be needed to derive the MDL encodings. Then, in Section~\ref{sec:known_nominal}, it is shown how to encode a categorical distribution when its parameters are known, which is the case for the default rule of a subgroup list. Finally, in Section~\ref{sec:unknown_nominal} it is shown how to encode a categorical distribution when the parameters of the distribution are unknown.

\subsubsection{Maximum Likelihood (ML) estimation of the parameters }\label{sec:ML_nominal}
Each description $a_i$ uniquely defines a subset $\Di$ given by its cover Eq.~\eqref{eq:coversubgroup}. However in the nominal case for each class label $c$, we also need to find its subset of the data $\Dac$, formally given by:
\begin{equation}
\Dac  = \{(\x,y) \in \Di \given y = c\}.
\end{equation} 
which allows us to compute the usage over each class $n_{c|i} = |\Dac|$. Now, we are in a position to use the maximum likelihood estimator for the parameters $\Ti$ of each categorical distribution as:
\begin{equation}
\hat{p}_{c|i} = \frac{n_{c|i}}{n_i}.
\end{equation}


We can show how to encode each subset of target values with the \emph{known} parameters of the distribution---the default rule of a subgroup list---and \emph{unknown} parameters---all the subgroups. 

\subsubsection{Encoding categorical distributions with \emph{known} parameters}\label{sec:known_nominal}
To encode target values with \emph{known parameters}---as is the case for the default rule of a subgroup list---we can directly use Eq.~\eqref{eq:known_encoding} with given parameter estimates $\Td = \hat{p}_{1|d}, \cdots, \hat{p}_{k|d}$ (marginal distribution over the whole dataset): 
\begin{equation}\label{eq:encode_known_nominal}
L(Y^d \given \hat{p}_{1|d}, \cdots, \hat{p}_{k|d}) =  \sum_{c \in \mathcal{Y}} - n_{c|d}\log \hat{p}_{c|d} = - \ell(\Td \given Y^d),
\end{equation}
where $\ell(\Td \given Y^d)$ is the log-likelihood of the parameter set $\Td$, and $n_{c|d}$ denotes the number of points associated with each class $c$ covered by default rule $\Yd$.

\subsubsection{Encoding categorical distributions with \emph{unknown} parameters}\label{sec:unknown_nominal}

When the parameters are \emph{unknown}---as is the case for each individual subgroup distribution---we will employ the Normalised Maximum Likelihood (NML) code, as it ``is optimal in the sense that it achieves the minimax optimal code length regret'' \citep{grunwald2007minimum}. 

Although the expression of the NML code can be daunting, its intuition is very clear \citep{kontkanen20051}, i.e., the NML code is equivalent to first encoding all maximum likelihood estimates of sequences $Z$ of $n_i$ points based on their likelihoods, and then encoding data $\Yi$ with its maximum likelihood estimate $\Ti$ as in Eq.~\eqref{eq:encode_known_nominal}. 
Formally, the NML code length of the subset $Y^i$ is given by\footnote{For details on the derivation of Eq.~\ref{eq:NML}, please see Appendix~\ref{appendix:NML_derivation}.}:
\begin{equation}\label{eq:NML}
\begin{split}
L_{\mathrm{NML}}(\Yi) &= -\log \frac{\prod_{y \in \Yi} \Pr (y \given \Ti)}{\sum_{Z \in \Y^{n_i}} \prod_{z \in Z} \Pr (z \given \TZ)} \\
&=   \sum_{c \in \mathcal{Y}} - n_{c|i}\log \hat{p}_{c|i} 
+\log  \sum_{Z \in \Y^{n_i}} \prod_{z \in Z} \Pr (z \given \TZ)  \\
&=  -\ell(\Ti \given \Yi) + \COMP(n_i,k)
\end{split}
\end{equation}

where $\Y^{n_i}$ is the space of all possible sequences of $n_i$ points with cardinality $k=|\Y|$ (possible values per point), $\TZ$ is the maximum likelihood estimate over $Z$, $\COMP(n_i,k)$ is the complexity---as it is called in MDL literature \citep{grunwald2007minimum}---of the multinomial distribution over $n_i$ points and $k$ categories. Note that this term can be efficiently computed in sub-linear time $ \mathcal{O}(\sqrt{dn_i}+k) $ if approximated by a finite floating-point precision of $d$ digits \citep{mononen2008computing}. \\


Finally, inserting \eqref{eq:encode_known_nominal} and \eqref{eq:NML} in \eqref{eq:data_encoding} we obtain, for the total data encoding of a \textbf{subgroup list}:
\begin{equation}\label{eq:length_data_subgroup_list_nominal}
L(\Ymat \given \Xmat, M) = \sum_{j=1}^{t} \left(  L(Y_j^d  \given \boldsymbol{\Theta}^d ) + \sum_{s_i \in S} L_{\mathrm{NML}}(Y_j^i) \right).
\end{equation}

\paragraph{Example 6 (continuation):} Let us revisit the \emph{Zoo} subgroup list example of Figure~\ref{fig:zoo_example} and compute the length of NML encoding of the first subgroup. To compute it we just need to get the probabilities associated with each category ($\{0;0;0.56;0.44;0;0;0\}$), the number of samples covered by each of them ($\{0;0;10;8;0;0;0\}$), and the total number of categories $k = |\Y| = 7$. Given these, the length of encoding of the data $Y^1$ is given by:
\begin{equation*}
\begin{split}
L_{\mathrm{NML}}(Y^1) &= \left(-10\log0.56 -8\log0.44  \right) + \COMP(18,7)\\
&= 17.84 + 10.42\\
& = 28.26 \: \text{bits}.
\end{split}
\end{equation*}

\subsection{ \textbf{Data encoding: numeric target variables}}\label{sec:numeric_data}

When we have one or more numeric target variables, the consequent of probabilistic rules as in Eq.~\eqref{eq:prob_rule} are now normal distributions $\mathcal{N}(\Theta)$ with parameters $\Theta = \{ \mu, \sigma \} $, and take the following form:
\begin{equation*}\label{eq:prob_rule_numeric}
\Pr(y \given \mu, \sigma ) = \frac{1}{\sqrt{2\pi \sigma^2}} \exp{  \left(-\frac{(y- \mu)^2}{2 \sigma^2}\right) },
\end{equation*}
where we use $\Pr(y \given \mu, \sigma )$ to denote the probability density function (pdf), which is a slight abuse of notation that we admit to unify the whole work.

This translates to a probabilistic rule of the form:
\begin{equation}
a \mapsto y_1 \sim \mathcal{N}(\hat{\mu}_{a1}, \hat{\sigma}_{a1}), \cdots, y_t \sim \mathcal{N}(\hat{\mu}_{at}, \hat{\sigma}_{at})
\end{equation}

To simplify the introduction of concepts, we will again assume we have only one target variable in $\Ymat$, and generalise the results to multiple variables at the end. Thus, throughout this section $\Ymat$ becomes $Y$, and the parameters of each subgroup $s_i$ become $\Theta^i = \{ \mu_i, \sigma_i\}$ as there is only one variable. The general form of a subgroup list with normal target distribution is given in Figure~\ref{fig:subgroup_list_numeric}.

\begin{figure}[ht]															
	\centering															
	\ra{1.1}\normalsize														
	\begin{tabular}{@{}rlrrr@{}}
		$s_1$:	&	IF	&	$a_1 \sqsubseteq  \x$		&	THEN	&	$y \sim \mathcal{N}(\hat{\mu}_{1}, \hat{\sigma}_{1})$					\\
		&	 	&	$\vdots  $		&		&				\\
		$s_{\omega}$:	&	ELSE IF	&	$a_\omega \sqsubseteq  \x$		&	THEN	&	$y \sim \mathcal{N}(\hat{\mu}_{\omega}, \hat{\sigma}_{\omega})$					\\ 
		dataset:	&	ELSE	&			&		&	$y \sim \mathcal{N}(\hat{\mu}_{d}, \hat{\sigma}_{d})$					\\
	\end{tabular}\vspace{0.3cm}\caption{Generic subgroup list model $M$ with $\omega$ subgroups $S= \{s_1,...,s_\omega\}$ and a single numeric target $Y$}\label{fig:subgroup_list_numeric}															
\end{figure}

In the following subsections, we will derive the data encoding for subgroup lists with normal distributions. First, in Section~\ref{sec:ML_numeric} we introduce the maximum likelihood estimators that will be needed to derive the MDL encodings. Then, in Section~\ref{sec:known_numeric} we show how to encode a normal distribution when its parameters $\mu$ and $\sigma$ are known, such as is the case for the default rule of a subgroup list. Finally, in Section~\ref{sec:unknown_numeric} we show how to encode a normal distribution using an uninformative prior when the parameters of the distribution are unknown. 

\subsubsection{Maximum Likelihood (ML) estimation of the parameters}\label{sec:ML_numeric} 
Each description $a_i$ uniquely defines a subset $\Di$ given by its cover \eqref{eq:coversubgroup}, which allows to estimate the parameters of each normal distribution using the maximum likelihood estimate over $Y^i$: 

\begin{equation}\label{eq:mean}
\hmui= \frac{1}{n_i}\sum_{y \in \Yi} y,
\end{equation}

\begin{equation}\label{eq:std}
\hsigi^2= \frac{1}{n_i}\sum_{y \in \Yi} (y-\hat{\mu}_i)^2,
\end{equation}

where $\hsigi^2$ is the biased estimator such that the estimate times $n_i$ equals the Residual Sum of Squares, i.e., $n_i \hsigi^2 = \sum_{y \in \Ya} (y-\hmui)^2 = RSS_i $. Note that the parameters of the default rule of Figure~\ref{fig:subgroup_list_numeric} are fixed for a dataset and thus correspond to estimates $\hmud$ and $\hsigd$ over all target values $Y$.

As each subgroup list defines a partition of the data, we can encode each target value part, $\Yi$ or $\Yd$, separately and sum them to obtain the total encoding of $Y$. In the case of subgroup lists, the last rule--i.e., default rule---has fixed parameters equal to the overall dataset distribution, while the subgroups parameters are not known in advance and have thus to be encoded together with the data points. 

We start by showing how to encode the subset of target values with the default `rule'---known parameters of the distribution---and then show how to encode each subgroup subset---unknown parameters of the distribution.

\subsubsection{Encoding normal distributions with \emph{known} parameters} \label{sec:known_numeric}

The target values not covered by any subgroup $\Yd $, as defined in \eqref{eq:coverdefault}, are covered by the default dataset `rule' and distribution at the end of a subgroup list. As the statistics $\hat{\Theta}_{d} = \{ \hmud, \hat{\sigma}_d \}$ are known and constant for a given dataset, one can simply encode the instances using this (normal) distribution, resulting in encoded length
\begin{equation}\label{eq:encode_known_numeric}
\begin{split}
L(\Yd\given \hmud, \hat{\sigma}_d)  &=  -\log \left[ \prod_{y \in \Yd} \frac{1}{\sqrt{2\pi \hsigd^2}} \exp \left( \frac{(y-\hmud)^2}{2 \hat{\sigma}_d^2}  \right) \right]\\
&=\frac{n_d}{2} \log 2\pi + \frac{n_d}{2} \log \hat{\sigma}_d^2 +  \left( \frac{1}{2 \hat{\sigma}_d^2} \sum_{y \in \Yd} (y-\hmud)^2   \right) \loge.
\end{split}
\end{equation}

The first two terms are normalising terms of a normal distribution, while the last term represents the Residual Sum of Squares (RSS) normalised by the variance of the data. Note that when $Y_d = Y$, i.e., the whole dataset target, RSS is equal to $n_d \sigma_d $ and the last term reduces to $n_d/2 \loge $.

\subsubsection{Encoding normal distributions with \emph{unknown} parameters}\label{sec:unknown_numeric}
In contrast to the previous case, here \emph{we do not know a priori the statistics defining the probability distribution corresponding to the subgroup}, i.e., $\hat{\mu}$ and $\hat{\sigma}$ are not given by the model and thus both need to be encoded. For this, we resort to the Bayesian encoding of a normal distribution with mean $\mu$ and standard deviation $\sigma$ unknown, which was shown to be asymptotically optimal \citep{grunwald2007minimum}. The optimal code length is given by the negative logarithm of a probability, and the optimal Bayesian probability for $\Ya$ is given by
\begin{equation}\label{eq:bayes}
\begin{split}
&L_{\mathrm{Bayes}} (\Yi) =  \\
&- \log \int_{-\infty}^{+\infty} \int_{0}^{+\infty} (2\pi \sigma)^{-\frac{n_i}{2}}\exp{ \left( - \frac{1}{2 \sigma^2}\sum_{y \in \Yi} (y - \mu)^2 \right) }  w(\mu,\sigma) \dif \mu \dif \sigma,
\end{split}
\end{equation}

where $w(\mu,\sigma)$ is the prior on the parameters, which needs to be chosen.

\paragraph{Choosing the prior.} The MDL principle requires the encoding to be as unbiased as possible for any values of the parameters, which leads to the use of uninformative priors. The most uninformative prior is  Jeffrey's prior, which is $1/\sigma^2$ and therefore constant for any value of $\mu$ and $\sigma$, but unfortunately its integral is undefined, i.e., $ \int \int \sigma^{-2} \dif \sigma \dif \mu = \infty$. Thus, we need to make the integral finite, which we will do next.

It should be noted that when using normal distributions with Bayes factors---Bayesian equivalent to traditional statistical testing---the authors tend to also add a normal prior on the effect size, as e.g., $\delta = \mu/\sigma \sim \mathcal{N} (0,\tau)$ \citep{jeffreys1998theory,gonen2005bayesian,rouder2009bayesian}. Nonetheless, this prior gives a higher probability to values of $\mu$ closer to zero, which is a bias that we do not want to impose. Thus we only use Jeffrey's prior, which converges\footnote{See proof in Appendix~\ref{appendix:BIC}.} to the Bayes Information Criterion (BIC) for large $n$.\\

Now, given the our prior $ w(\mu,\sigma) = \frac{1}{\sigma^2\sqrt{2\pi}}$---where $\sqrt{2\pi}$ was added for normalisation reasons---the remaining question is how we can make the integral finite. The most common solution, which we also employ, is to use $u$ data points from $\Yi$, denoted $Y^{i|u}$, to create a proper conditional prior $w(\mu,\sigma \given Y^{i|u})$. As there are only two unknown parameters, we only need two points hence $u=2$ \citep{grunwald2007minimum}; for more on the interpretation of such ``priors conditional on initial data points", see \cite{grunwald2019minimum}. Consequently, we first encode $\Ytwo$ with a non-optimal code that is readily available---i.e., the dataset distribution of Eq.~\eqref{eq:encode_known_numeric}---and then use the Bayesian rule to derive the total encoded length of $\Yi$ as

\begin{equation}
\begin{split}
L_{\mathrm{Bayes}2.0}(\Yi) &= -\log \frac{P_{\mathrm{Bayes}}(\Yi)}{P_{\mathrm{Bayes}}(\Ytwo)}P(\Ytwo \given \mu_d,\sigma_d) \\
&= L_{\mathrm{Bayes}}(\Yi)+L_{cost}(\Ytwo),
\end{split}
\end{equation} 
where $L_{cost}(\Ytwo) = L(\Ytwo \given \mu_d,\sigma_d) - L_{\mathrm{Bayes}}(\Ytwo)$ is the extra cost incurred by encoding two points non-optimally. After some re-writing\footnote{The full derivation of the Bayesian encoding and an in-depth explanation are given in Appendix~\ref{appendix:Bayesian_derivation}.} we obtain the encoded length of the $y$ values covered by a subgroup $\Yi$ as
\begin{equation}\label{eq:lengthsubgroup}
\begin{split}
&L_{\mathrm{Bayes}2.0}(\Yi)= L_{\mathrm{Bayes}}(\Yi) + L_{cost}(\Ytwo)\\ 
&=1 +\frac{n_i}{2} \log \pi- \log \Gamma \left( \frac{n_i}{2} \right) + \frac{1}{2} \log (n_i) +\frac{n_i}{2} \log  n_i \hat{\sigma}_{i}^2  + L_{cost}(\Ytwo), \\
\end{split}
\end{equation}
where $\Gamma$ is the Gamma function that extends the factorial to the real numbers ($\Gamma (n) = (n-1)!$ for integer $n$) and $\hmui$ and $\hsigi$ are the statistics of Eqs.~\eqref{eq:mean} and \eqref{eq:std}, respectively. Note that for $\Ytwo$ any two unequal values (otherwise $\hat{\sigma}_2 =0$ and $L_{\mathrm{Bayes}}(\Ytwo) = \infty$) can be chosen from $\Ya$, thus we choose them such that they minimise $L_{cost}(\Ytwo)$.  
Finally, inserting \eqref{eq:encode_known_numeric} and \eqref{eq:lengthsubgroup} in \eqref{eq:data_encoding} we obtain for the total data encoding for a \textbf{subgroup list}:
\begin{equation*}\label{eq:length_data_subgroup_list_numeric}
L(\Ymat \given \Xmat, M) = \sum_{j=1}^{t} \left(  L(Y_j^d  \given \boldsymbol{\Theta}^d ) + \sum_{s_i \in S} L_{\mathrm{Bayes}2.0}(Y_j^i) \right).
\end{equation*}

\paragraph{Example 7 (continuation):} We revisit the \emph{Automobile} subgroup list of Figure~\ref{fig:car_example} and find the length of the $Bayes2.0$ encoding (Eq.~\eqref{eq:lengthsubgroup}) of the first subgroup. To compute it we need to get the statistics of the subgroup ($\hat{\Theta}^1= \{\hat{\mu}_1 =35; \hat{\sigma}_1=8\}$), the number of samples it covers ($n_1 =11$), the dataset statistics ($\hat{\Theta}^d= \{\hat{\mu}_d =13; \hat{\sigma}_d=8\}$), and the two points closest to the dataset mean $Y^{1|2} =\{ 14; 31\}$ that makes the encoding proper (and which are not available in the example information). Assuming that $L_{cost}(\Ytwo) = 0.69 \: \text{bits}$ for simplicity, the length of the encoding of $Y^1$ is given by:
\begin{equation*}
\begin{split}
L_{\mathrm{Bayes}2.0}(Y^1) =& 1 +\frac{11}{2} \log \pi- \log \Gamma \left( \frac{11}{2} \right) + \frac{1}{2} \log (11) +\frac{11}{2} \log  11 \cdot 8^2  \\
& + L_{cost}(\Ytwo)\\
= &58.06 +0.69 \\
= &58.75 \: \text{bits}. \\
\end{split}
\end{equation*} 

%
%
\section{MDL-optimal subgroup lists relation to WKL-based SD and Bayesian testing}\label{sec:mdlsl_wklsd_bayes}

This section investigates the equivalence of our MDL-optimal formulation of subgroup lists to top-$k$ subgroup discovery with WKL and Bayesian testing with multiple hypotheses. First, Section~\ref{sec:sd_proof} shows that when our subgroup lists only contain one subgroup, they correspond top-$1$ subgroup discovery with WKL as a quality measure. Second, Section~\ref{sec:mdl_statistic} shows that adding one subgroup to the list---MDL greedy gain---corresponds to Bayesian proportions, multinominal, and t-test for binary, nominal, and numeric targets, respectively, plus a term for multiple-hypothesis testing.

\subsection{Relationship of MDL-optimal subgroup lists to WKL-based SD}\label{sec:sd_proof}

We now investigate the relationship between finding an MDL-optimal subgroup list and WKL-based top-$k$ subgroup discovery. Remember that WKL is a subgroup discovery measure based on information-theory~\citep{van2010maximal} and takes the form of Eq.~\eqref{eq:generalmeasure} for a general probability distribution; we describe it in more detail in Subsection~\ref{sec:wkl}.\\

Now, assume that we have a single target variable ($Y$ instead of $\Ymat$) and a subgroup list consisting of just one subgroup $s$ with description $a$ (and the default rule). Next, let us turn the MDL minimisation problem into a maximisation problem by multiplying Eq.~\eqref{eq:LengthTotal} by minus one and adding a constant (for each dataset) $L(Y \given \boldsymbol{\Theta}^d )$ to obtain:
\begin{equation*}\label{eq:maximize}
s^* = \argmax_{s \in \M} \left[ L(\Yd  \given \boldsymbol{\Theta}^d ) - L(Y \given \Xmat,M)  - L(M) \right].
\end{equation*}

In the case of a subgroup list with \emph{one subgroup} and one target, the data encoding of Eq.~\eqref{eq:length_data_subgroup_list_nominal} for nominal targets or Eq.~\eqref{eq:length_data_subgroup_list_numeric} can be substituted by
\begin{equation*}
L(Y \given \Xmat, M) =  L(\Yd  \given \boldsymbol{\Theta}^d ) +  L_{*}(\Ya)
\end{equation*}
where $L_{*}$ stands for $L_{\mathrm{NML}}$ or $L_{\mathrm{Bayes}2.0}$ for nominal and numeric targets, respectively. Also, note that $\Yd$ is given by all the points not covered by the subgroup description $a$, i.e., $Y^{\neg a}$. Thus, we can further develop the maximisation problem to\footnote{The derivation for categorical and normal distributions is shown in Appendix~\ref{appendix:wkl_derivation}}:
\begin{equation}\label{eq:KLproof} 
\begin{split}
L(Y \given  \Td)&-L(Y \given \Xmat, M)- L(M) \simeq \\ 
&\simeq n_a KL(\Ta; \Td) -  \texttt{COMP}(n_a, Dist)- L(M),
\end{split}
\end{equation}
where $n_a KL(\Ta; \Td)$ is the Weighted Kulback-Leibler divergence from $\Ta$ to $\Td$, and $\texttt{COMP}(n_a, Dist)$ is the complexity associated with the target probability distribution---$\COMP(n_a,k)$ for categorical and $\log n_a$ for normal. The equality of the expression holds for categorical and is asymptotically equal for the normal. \emph{This result shows that finding the MDL-optimal subgroup is equivalent to finding the subgroup that maximises WKL, plus two extra terms: one that defines the complexity of the distribution $\COMP(n_a,k)$, and another that defines the complexity of the subgroup $L(M)$}.

\paragraph{Dispersion-correction quality measure for numeric targets.} Importantly, we can observe from Eq.~\eqref{eq:KLproof} that the measure based on the Kullback-Leibler divergence of normal distributions is part of the family of \emph{dispersion-corrected} subgroup quality measures, as it takes into account both the centrality and the spread of the target values \citep{boley2017identifying}.

When we consider subgroup lists having more than one subgroup, Eq.~\eqref{eq:KLproof} simply expands to:
\begin{equation*}
\begin{split}
L(Y \given  \Td)&-L(Y \given \Xmat, M)- L(M)= \\
&\simeq\sum_{a_i \in S} n_i KL(\Ti; \Td) -  \sum_{a_i \in S} \texttt{COMP}(n_a, Dist) - L(M) \\
&= \textrm{SWKL}(S) -  \sum_{a_i \in S} \texttt{COMP}(n_a, Dist) - L(M),
\end{split}
\end{equation*}
where $\textrm{SWKL}(S) $ is the measure for subgroup set quality that we proposed in Section~\ref{sec:back_swkl}, and the other terms penalise the complexity of the subgroup list.
This demonstrates that the MDL-based objective for the optimal subgroup corresponds to a subgroup set quality minus two terms for model complexity---multiple=hypothesis testing.

\subsection{Relationship of MDL-optimal subgroup lists to Bayesian testing}
\label{sec:mdl_statistic}

We will now show how our MDL criterion relates to Bayesian and multiple hypothesis testing. The Bayesian alternative to statistical testing is the Bayesian factor presented in Section~\ref{sec:back_bayes}, which compares the best model by comparing the likelihoods of the data given each model.
Now imagine we are comparing two models $M_1$ and $M_2$ based on their MDL quality as defined by Eq.~\eqref{eq:LengthTotal}:
\begin{equation*}
\begin{split}
L(D, M_1)-L(D, M_2) &= - \log \left(\frac{\Pr(D, M_1)}{\Pr(D, M_2)}\right) \\
&= -\log \left(\frac{\Pr(Y \given \Xmat, M)}{\Pr(Y \given  \Td)} \times \frac{\Pr(M_1)}{\Pr(M_2)} \right) \\
&= - \log K_{1,2} -\log \left( \frac{\Pr(M_1)}{\Pr(M_2)} \right),
\end{split}
\end{equation*}
where we use the Shannon-Fano code of Eq.~\eqref{eq:shannonfano} to transform code length in bits $L(\cdots)$ to probabilities $\Pr(\cdots)$, and $K_{1,2}$ is the Bayes factor between model $1$ and $2$ (as presented in Section~\ref{sec:back_bayes}). In practice, taking into account $\Pr(M)$ (or $L(M)$) is equivalent to using the posterior distributions instead of just the ``Bayes'' factor, and in our case, amounts to a penalty for multiple hypothesis testing as described in Section~\ref{sec:model_multiple}. This is the general equivalence between our formulation and Bayesian testing but now let us look at the specific case of adding only one subgroup at the end of the list, i.e., the \emph{greedy gain}.\\

\paragraph{Greedy gain.} Adding one subgroup to the list is of special interest to us because it relates with both the greedy gain of our algorithm (Section~\ref{sec:gain}) and to Bayesian statistical testing against the dataset distribution $\Td$. 

Formally, the greedy gain of adding a subgroup to a model $M$, where $M_1=M$ is any subgroup list and $M_2= M \oplus s$, i.e., $M_2$ is $M$ plus one more subgroup $s$ at the end of the list. Observe that the subgroups in $M$ cover the dataset exactly these as the same subgroups $M \oplus s$ and that the default rule is fixed. Thus, this model comparison only depends on $\Td$ from $M$ and $\Ta$ from $M \oplus s$:
\begin{equation}\label{eq:statistica_greedy}
\begin{split}
L(D, M)-L(D, M \oplus s) \simeq&  n_a KL(\Ta; \Td) -  \texttt{COMP}(n_a, Dist)\\
&+ L(M)-L(M \oplus s), \\
\end{split}
\end{equation}
where we obtained the expression on WKL divergence again. Thus, the hypothesis we are testing here is that the data non-covered by $M$ is better described by the dataset distribution $\Td$ or the subgroup distribution $\Ta$. This tells us that adding one subgroup that minimises the MDL expression to the subgroup list is equivalent to Bayesian testing \cite{rouder2009bayesian}.
Specifically, the first two terms are equivalent to a Bayesian proportions test (with a binary target), a multinomial test (with a nominal target), or a t-test (with a numeric target described by a normal). $L(M)-L(M \oplus s)$ accounts for multiple hypothesis testing by penalising for all the possible subgroups that could be added to the model $M$, as explained in Section~\ref{sec:model_encoding}.
Indeed, in the nominal case, the first two terms are similar to one of the quality measures proposed by \cite{song2016subgroup}---we use the NML encoding and they use a Bayesian one---for taking into account the uncertainty of the class distribution for nominal targets, assuming that the subgroup and dataset are different. However, they do not consider our additional term for multiple hypothesis testing, which is necessary to have a low number of false positives.

%
%

\section{The SSD++ Algorithm}\label{sec:SSDpp}

This section proposes SSD++, a heuristic algorithm to find good subgroup lists based on the proposed MDL formulation of Section~\ref{sec:MDL} and \ref{sec:data_encoding}. Our proposed algorithm combines the Separate-and-Conquer (SaC) \citep{furnkranz1999separate} strategy of iteratively adding the local best subgroup to the list with beam search for candidate subgroup generation (for a short introduction on these algorithms, please refer to Section~\ref{sec:back_alg_sd}).
We use a double greedy approach because the problem of finding the optimal subgroup list is NP-hard (Section~\ref{sec:nphard}), and the algorithm can be extended to different types of target variables. 

Further, in subgroup discovery, beam search was empirically shown to be competitive in terms of quality compared to a complete search while offering a considerable speed-up \citep{meeng2020forreal}. Greedy heuristic approaches are a common practice in MDL-based pattern mining \citep{vreeken2011krimp,proencca2020interpretable} and rule-based learning \citep{furnkranz2012foundations}.

Lastly, our approach of greedy search and adding one subgroup at a time is computationally interpretable to the user, as it adds at each iteration the locally best and most statistically significant subgroup found by beam search. \\



This section is divided as follows. First, Section~\ref{sec:gain} presents the greedy gain---compression gain---of adding one subgroup to the list and its equivalence to WKL-based SD. Then, Section~\ref{sec:greedysearch} describes the SSD++ algorithm. Finally, Section~\ref{sec:timespace_complexity} shows the time and space complexity of SSD++.

\subsection{Compression gain}\label{sec:gain}

To quantify the quality of annexing a subgroup $s$ at the end (after all the other subgroups and before the default rule) of model $M$, denoted $M \oplus s$, we employ the \emph{compression gain}:
\begin{equation}\label{eq:greedy_gain}
s = \argmax_{s \in \mathcal{s}} \Delta_{\beta} L(D,M \oplus s) = \argmax_{s \in \mathcal{s}} \left[\frac{L(D,M) - L(D,M\oplus s)}{(n_s)^{\beta}} \right], \; \beta \in [0,1]
\end{equation} 
where $\beta$ weighs the normalisation level, and $\Delta_{\beta} L(D,M \oplus s)$ should be greater than zero for a decrease in the encoded length from $L(D,M)$ to $L(D,M \oplus s)$, i.e., a favourable statistical test for adding a subgroup (Section~\ref{sec:mdl_statistic}). Considering the extremes, with $\beta = 1$ we have the \emph{normalised gain} first introduced for the classification setting by \cite{proencca2020interpretable}, and for $\beta = 0$ we have the \emph{absolute gain} which is just the regular gain used in the greedy search of previous MDL-based pattern mining \citep{vreeken2011krimp}. 

Developing Eq.~\eqref{eq:greedy_gain} further shows that the compression gain only depends on the added subgroup $s$ with description $a$, as in the specific case of a subgroup list the default rule is fixed and it is the same for $M$ and $M \oplus s$:

\begin{equation}\label{eq:gain_kld}
\begin{split}
\Delta_{\beta} L(D,M \oplus s)&= \frac{L (\Ymat \given \Xmat, M) - L(\Ymat \given \Xmat, M\oplus s)}{(n_a)^{\beta}} + \frac{L (M) - L(M\oplus s)}{(n_a)^{\beta}}  \\
&= \Delta_{\beta} L (\Ymat \given \Xmat, M\oplus s)+ \Delta_{\beta} L(M\oplus s), \\
&\simeq  \frac{1}{(n_a)^{\beta}}  \left ( n_a KL(\hat{\Theta}_s;\hat{\Theta}_d) -\texttt{COMP}(n_a) + \Delta L(M \oplus s) \right ) \\
\end{split}
\end{equation}

where $\Delta_{\beta} L (\Ymat \given \Xmat, M\oplus s)$ and $\Delta_{\beta} L(M\oplus s)$ are the data and model compression gain, respectively. The last expression shows the equivalence of the compression gain to statistical testing with multiple-hypothesis as shown in Eq.~\eqref{eq:statistica_greedy}. $\texttt{COMP}(n_a, Dist)$ is the uncertainty associated with the probability distribution for $n_a$ points, which for categorical distributions is given by $\COMP(n_a,k)$ and for normal is $\log n_a$

\paragraph{Interpretation of hyperparameter $\beta$.} The hyperparameter $\beta$ represents a tradeoff between finding many subgroups that cover few instances or few subgroups that cover many instances\footnote{For details on the empirical analysis of different $\beta$ values, please refer to Appendix~\ref{appendix:empiricalabsvsnorm}}. In the general form of a subgroup quality measure of Eq.~\eqref{eq:generalmeasure}, $\beta$ is just given by $\beta = 1- \alpha$. Later, we empirically show that the \emph{normalised gain} ($\beta= 1$) usually achieves a better MDL score than other $\beta$ values; this was already known for other measures from rule learning theory \citep{furnkranz2012foundations}. Nonetheless, the main objective of subgroup discovery is to \emph{locally} describe regions in the data that strongly deviate from a certain target. Thus, the user can specify what she is looking for in the data: either a more granular and detailed perspective ($\beta$ close to one) or a more general and high-level one ($\beta$ close to zero). Note that, for comparison to other algorithms \emph{we will always use the normalised gain} ($\beta= 1$) except when explicitly stated.

\subsection{SSD++ algorithm}\label{sec:greedysearch}

We propose SSD++\footnote{Our implementation can be found on: \url{https://github.com/HMProenca/RuleList}}, a heuristic algorithm with two main components: 1) a SaC iterative loop that adds the best-found subgroup at the end of the subgroup list; and 2) a beam search to find high-quality subgroups at each SaC iteration based on the compression gain of Eq.~\ref{eq:gain_kld}. 

More specifically, the greedy search algorithm starts from an empty list, with just a default rule equal to the priors in the data, and adds subgroups according to the well-known \emph{separate-and-conquer} strategy \citep{furnkranz2012foundations}: 1) iteratively find and add the subgroup that gives the most considerable improvement in compression; 2) remove the data covered by that rule; and 3) repeat steps 1-2 until compression cannot be improved. This implies that \emph{we always add subgroups at the end of the list, but before the default rule}. Beam search is used for candidate generation at each iteration to find the best candidate to add. 
In terms of the LeGo framework \citep{knobbe2008local}, our work can be seen as first mining all the possible local patterns (subgroups)---technically we could use beam-search only once, but using it at each iteration results in better candidates---and then using the SaC algorithm to sequentially add the best subgroup to a list.\\

Note that this algorithm extends that of \cite{proencca2020discovering} for univariate numeric targets to three extra target types---univariate and multivariate nominal, and multivariate numeric---with the added normalisation hyperparameter $\beta$.

\subsubsection{Algorithm description}\label{sec:algorithm_highlevel}

Algorithm~\ref{alg:SSD} presents SSD++. The algorithm starts by taking as input a dataset $D$ and the beam search parameters, namely the number of cut points $n_{cut}$, the width of the beam $w_b$, and the maximum depth of search $d_{max}$. It initialises the rule list with the default rule, based on the dataset empirical distribution (Ln~\ref{alg:initialise}). Then, while the beam search algorithm returns subgroups that improve compression (Ln~\ref{alg:loop}), it keeps iterating over two steps: 1) finding the best subgroup from all candidates generated in the beam search (Ln~\ref{alg:beam}); and 2) adding that subgroup to the end of the model, i.e., after all the existing subgroups in the model (Ln~\ref{alg:addrule}). The beam search (explained further in the next paragraph) returns the best subgroup on the data not covered by any subgroup already in model $M$. When no subgroup improves compression (non-positive gain), the loop stops and returns the subgroup list. Beam search is used at each iteration to generate the best candidates at each SaC iteration, instead of only once at the beginning, as it could yield local optima and get stuck on the top-$k$ subgroups. 

\paragraph{Beam search.} Given a beam width $w_b$ and maximum search depth $d_{max}$ it consists of: $1)$ find all items, i.e., all conditioned variables such as $ x_1<5$ (see next paragraph for the numeric discretisation with cut points) or $ x_2 = category$, and add the best $w_b$ items according to compression gain (Eq.~\eqref{eq:gain_kld}) as subgroups of size $1$ to the beam; $2)$ refine all subgroups in the beam with all items and add the best $w_b$ to a new empty beam; $3)$ repeat $2$ and $3$ until the maximum depth $d_{max}$ of the beam is reached and return the best subgroup---according to the compression score---found in all iterations.   \\

\paragraph{Numeric discretisation.} Suppose a numeric variable $X_j$, and a number of cut points $n_{cut}$. The $items$ generated from this numeric variable are all valid subsets (they must cover at least one instance) given by equal frequency discretisation with open and closed intervals for $n_{cut}$ cut points. Open intervals require one operator ($\geq$ or $\leq$), while closed intervals require two ($\geq$ and $\leq$). As an example, in the case of a generic variable $X_j$ and $n_{cut}=2$, with $cut\_point_1 = 10$ and $cut\_point_2 = 20$ it generates four $items$ with one operator, i.e., $items_{1op}= \{$ $x_j \geq 10$, $x_j \leq 10$,$x_j \geq 20$, $x_j \leq 20 \}$, and one $item$ with two operators, i.e., $items_{2op}= \{ 10 \leq x_j \leq 20\}$.

\begin{algorithm}[!h]
	\centering
	\caption{SSD++ algorithm}\label{alg:SSD}
	\begin{algorithmic}[1]
		\INPUT Dataset $D$, number of cut points $n_{cut}$, beam width $w_b$, depth max. $d_{max}$ and normalisation $\beta$		
		\OUTPUT Subgroup list $S$
		\State $M \gets [\Theta_d(Y)]$ \label{alg:initialise}
		\State $subgroup \gets BeamSearch(M,D,w_b,n_{cut},d_{max})$
		\While{$\Delta_{\beta} L(D,M \oplus subgroup) > 0$} \label{alg:loop}
		\State $subgroup \gets BeamSearch(M,D,w_b,n_{cut},d_{max})$ \label{alg:beam}		
		\State $M  \gets M \oplus subgroup$	\label{alg:addrule}	
		\EndWhile 
		\State \textbf{return} $S \in M$
	\end{algorithmic}
\end{algorithm} 

\subsection{Time and Space Complexity}\label{sec:timespace_complexity}

In this section, we analyse the time and space complexity of SSD++ as given in  Algorithm~\ref{alg:SSD} in Section~\ref{sec:time_complexity} and \ref{sec:space_complexity}, respectively.

\subsubsection{Time Complexity}\label{sec:time_complexity}

The algorithm can be divided into three parts: $1$) preprocessing of the data; $2)$ the Separate and Conquer (SaC) algorithm; and $3)$ the beam search. In addition, there are different complexities depending on the target type, as each statistic requires different computations.

\paragraph{1) Preprocessing phase.} In the preprocessing phase, all the coverage bitsets of the items are generated, i.e., the indexes of the instances covered by each item generated from numerical and nominal variables. The set of all items is $\zeta$ and its size $|\zeta|$. Thus, we go over the data a maximum of $|\zeta|$ times, obtaining a time complexity of $\mathcal{O}(|\zeta|n)$, and the results are stored in a dictionary for $\mathcal{O}(1)$ access. Also, some constants are cached for a fixed amount the first time they are computed, such as the universal code of integers $\LN(i)$, and $\Gamma(i)$ for the numeric target case, and $\COMP(i)$ in the categorical case.

\paragraph{2) SaC phase.}For the SaC phase, it is clear that the algorithm runs the beam search $|S|$ times and will thus multiply the time complexity of the beam search by $|S|$. 

\paragraph{3) Beam search phase.} For the last $d_{max}-1$ iterations of the loop, each of $w_b$ candidates in the beam is refined with all $|\zeta|$ items, which gives a time complexity by itself of $\mathcal{O}( d_{max}w_b|\zeta|)$. Then, for each refinement, the algorithm computes its coverage, statistics, and score, where the last two depend on the number and type of target. 

The \emph{coverage} of the refinement is the logical conjunction of two bitsets, i.e., the bitset of the candidate $b_{cand}$ and that of the item $b_{item}$. The computation of this new coverage has a time complexity of $\mathcal{O}(|b_{cand}|+|b_{item}|)$, which in a worst-case equals a run over the dataset $\mathcal{O}(n+n) = \mathcal{O}(n)$. Thus the time complexity of the algorithm is given by
\begin{equation*}
\mathcal{O}\left( |S|d_{max}w_b|\zeta| stats \right),
\end{equation*}
where $stats$ is the time complexity associated with computing the statistics for one candidate. Now, we will analyse the specific $stats$ complexity depending on the type of target.

\paragraph{Nominal target variables.} The \emph{statistics for categorical distributions} require the computation of the usage for each class for each target of each subgroup rule and the new default rule. Assuming a maximum number of classes $k$ (for all target variables) and $t$ target variables, then the worst case for the coverage gives $\mathcal{O}(tnk)$ from which the likelihood can be directly computed.

The \emph{nominal score} requires the computation of the data and model encoding, from which the data encoding dominates. The data encoding entails the computation of the NML complexity and likelihood for each refinement. In general, the values of the NML complexity are just computed once and then cached; thus, in a worst-case where one requires to compute $n$ values for $\COMP(n_i), \forall_{n_i=1,...,n}$. Using the approximation of \cite{mononen2008computing} for its computation, with $\mathcal{O}(\sqrt{10n_i}+k)$, gives a worst-case complexity of $\mathcal{O}(tn(\sqrt{n}+k))$. This does not depend on the parameters of the beam, as the lookup of these values is $\mathcal{O}(1)$. The likelihood generally dominates over this term as it is computed for each refinement.

Thus the total time complexity for \textbf{nominal targets} is given by:
\begin{equation*}\label{eq:time_complexity_nominal}
\mathcal{O}\left( |S|d_{max}w_b|\zeta|tnk  + tn(\sqrt{n}+k) \right)
\end{equation*}

\paragraph{Numeric target variables.} The \emph{statistics for normal distributions} require the computation of the mean and variance (or residual sum of squares) for the refined subgroup and for the default rule. The mean can be computed in $\mathcal{O}(n)$ and given its value the variance can also be computed in $\mathcal{O}(n)$. Thus, for all the targets, one obtains $\mathcal{O}(tn)$.

The \emph{numeric score} requires the computation of the data and model encoding, from which the data encoding dominates. The data encoding entails calculating the gamma function and the direct use of the statistics. Similar to the NML complexity, we compute the values of the gamma function as needed and cache them afterwards. In general, the computation of the gamma function is dominated by the other terms as we only compute it at most $n$ times.

Thus the total time complexity for \textbf{numeric targets} is given by:
\begin{equation*}\label{eq:time_complexity_numeric}
\mathcal{O}\left( |S|d_{max}w_b|\zeta|tn  \right).
\end{equation*}
Notice that this represents a worst-case scenario. In practice, the direct use of bitsets to compute the class usages in the nominal case makes it faster than its numeric counterpart for the same dataset size.

\subsubsection{Space Complexity}\label{sec:space_complexity}

The main memory consumption resources of the algorithm are: $1)$ the storage of items $\zeta$; $2)$ the beam; and  $3)$ the cached constants. The item storage requires at most the storage of $|\zeta|$ bitsets, with each bitset taking $\mathcal{O}(n)$, thus it totals $\mathcal{O}(|\zeta|n)$. The beam saves $w_b$ bitsets at a time, thus having a space complexity of $\mathcal{O}(w_bn)$. The cached values make up a total of $n$ values being dominated by the items or beam part.
Thus, depending on which part dominates, the space complexity of the algorithm is 
\begin{equation*}
\mathcal{O}(w_bn + |\zeta|n).
\end{equation*} 

\section{Empirical evaluation}\label{sec:exps_results}

In this section, we will empirically validate our proposed problem formulation and the SSD++\footnote{Our implementation can be found on: \url{https://github.com/HMProenca/RuleList}; and for replication of the experiments, please refer to \url{https://github.com/HMProenca/RobustSubgroupDiscovery}} algorithm. To do this, we will test how varying the hyperparameters of SSD++ affects the subgroups found, and then we will compare SSD++ against state-of-the-art algorithms in subgroup set discovery. \\

This section is divided as follows. In Section~\ref{sec:empirical_hyperparameters} we evaluate the effect of changing the different hyperparameters of SSD++. Then, in Section~\ref{sec:empirical_setup} we present the setup for validating our approach based on algorithms compared against, datasets, and measures used to evaluate them. After that, in Section~\ref{sec:empirical_nominal}, the results for univariate and multivariate nominal targets compared with state-of-the-art algorithms are presented. Then, in Section~\ref{sec:empirical_numeric} the results for univariate and multivariate numeric targets compared with state-of-the-art algorithms are shown. After that, in Section~\ref{sec:statistica_robustness} the statistical robustness of our formulation and algorithm are tested based on the generalisation to unseen data. Finally, in Section~\ref{sec:empirical_time} the runtimes of the algorithms are compared.

\subsection{Influence of SSD++ hyperparameters}\label{sec:empirical_hyperparameters}

Here we study the effect of SSD++ hyperparameters on the discovered subgroup lists. To not overfit our hyperparameters to the datasets and for this reason obtain a better performance than other methods, the values of SSD++ hyperparameters for the remaining of the experiments (besides this section) are fixed at the standard values of the DSSD implementation for the beam search, i.e., beam width $w_b=100$, number of cut points $n_{cut} = 5$, and maximum search depth $d_{max} =5$, and to the compression gain normalisation term $\beta = 1$ (normalised gain). These values are assumed to be enough to achieve convergence and to obtain good subgroup lists and are thus taken as the \emph{standard values} of SSD++.

Now, to evaluate hyperparameter influence, we vary one hyperparameter value at a time while others remain fixed at their \emph{standard values}. The results of varying the compression gain normalisation hyperparameter $\beta$ can be seen in Appendix~\ref{appendix:empiricalabsvsnorm}; the results of varying the beam search hyperparameters $w_b$, $n_{cut}$, and $d_{max}$ can be found in Appendix~\ref{appendix:empiricalbeamsearch}.

\paragraph{Normalisation term $\beta$.} The results are evaluated in terms of compression ratio, SWKL, and the number of rules. For compression gain, the results (as shown in Appendix~\ref{appendix:empiricalabsvsnorm}) are similar for a small number of samples, but $\beta =1$ and $0.5$ obtain better results for larger datasets. In terms of SWKL, normalised gain ($\beta =1$) is better. On the other hand, in terms of the number of rules $\beta =1$ can obtain one order of magnitude more rules than the others, especially for larger datasets.

\paragraph{Beam search hyperparameters $w_b$, $d_{max}$, and $n_{cut}$.} The results are evaluated in terms of compression ratio and the average number of conditions per subgroup (for $d_{max}$). In general, increasing any of the three values results in better models according to relative compression. It is also interesting to note that for maximum depths above $5$ it is rare to have an average number of conditions above $4$, backing up our decision for the standard value $d_{max} =5$.

\subsection{Setup of the subgroup quality performance comparisons}\label{sec:empirical_setup}

In this section, we evaluate the quality of our proposed method by comparing it to the state-of-the-art approaches in subgroup set discovery, which may vary depending on the type of target variable(s). The comparison takes three dimensions: $1)$ the \emph{algorithms} used to compare against; $2)$ \emph{measures} used to evaluate the quality of the subgroups found by each algorithm; $3)$ the \emph{datasets} in which the algorithms are evaluated. We now discuss the details of each dimension. 

\subsubsection{Algorithms} 
The algorithms we compared and their relevant characteristic are listed in Table~\ref{table:setup_algorithms}. A short description of each is as follows: 
\begin{enumerate}
\item top-$k$\footnote{top-$k$, seq-cover, and DSSD are available in the implementation of the DSSD algorithm \url{http://www.patternsthatmatter.org/software.php\#dssd/} \label{footnote:dssd}} - standard subgroup discovery miner used as a benchmark.
\item seq-cover\footref{footnote:dssd} - sequential covering as implemented in the DSSD implementation.
\item CN2-SD\footnote{Available in the Orange data mining toolkit \url{https://orangedatamining.com/}} - the classical sequential covering subgroup discovery algorithm, which is only implemented for nominal targets, and only removes the examples of the class of interest already covered (not all examples covered, as seq-cover does).
\item Diverse Subgroup Set Discovery (DSSD)\footref{footnote:dssd} - diverse beam search for diverse sets of subgroups \citep{van2012diverse}.
\item Monte Carlo Tree Search for Data Mining (MCTS4DM) - an approach to improve on beam search to find better subgroups without getting stuck in local optima \citep{bosc2018anytime}.
\item FSSD - a sequential approach for subgroup set discovery that defines a set as a disjunction of subgroups \citep{belfodil2019fssd}.
\end{enumerate}

As can be seen in Table~\ref{table:setup_algorithms} most algorithms can only be applied to single-target binary problems, and besides SSD++ only top-$k$, seq-cover and CN2-SD support the use of Sum of Weighted Kullback-Leibler (SWKL) divergence to measure the quality of the found subgroup set. Thus we only compare against seq-cover and CN2-SD, algorithms that output a subgroup list and can be applied to many target types, and with top-$k$ as a reference of a non-diverse subgroup discovery algorithm. The algorithms that output sets do not have a stopping criterion or global formulation, and underperform in terms of SWKL; thus those comparisons are relegated to Appendix~\ref{appendix:non-sequential}. As an example, DSSD can indeed be applied to all types of target variables, but the fact that it uses weighted sequential covering makes it unsuitable to use the SWKL, making it unfairly underperform and unsuitable for a fair comparison (as shown in the Appendix). Also, note that we do not compare with machine learning algorithms that generate rules for classification or regression, such as RIPPER or CART, as the rules generated aim at making the best prediction possible and not the highest difference from the dataset distribution, as shown theoretically in Appendix~\ref{appendix:proof_sd_vs_prediction}. 

\begin{table}[t!]\centering																\begin{threeparttable}[b]
	
	\caption{Algorithms included in the comparison and their functionalities. \emph{Quality} represents the quality measure used to evaluate one single subgroup,  \emph{search} is the type of search algorithm supported, \emph{swkl} shows if it supports SWKL to measure the quality of a subgroup set, \emph{output} tells if the subgroups discovered form a list or a set, and `\cmark' and `$-$' represent if that type of target variable(s) is supported. MCTS stands for Monte Carlo Tree Search.}\label{table:setup_algorithms}																				
	\ra{1.0} \begin{tabular}{@{}llllcccccc@{}}\toprule																				
		&		&		&		&		&	\multicolumn{3}{r@{}}{nominal}					&	\multicolumn{2}{r@{}}{numeric}			\\	\cmidrule(l){6-8} \cmidrule(l){9-10}
		Algorithm	&	quality	&	search	&	output	&	swkl	&	bin.	&	nom.	&	multi	&	single	&	multi	\\ \midrule	
		SSD++	&$	WKL	$&	beam	&	list	&	\cmark	&	\cmark	&	\cmark	&	\cmark	&	\cmark	&	\cmark	\\	
		top-$k$ 	&$	WKL_{\mu}\tnote{a}	$&	beam	&	set	&	\cmark	&	\cmark	&	\cmark	&	\cmark	&	\cmark	&	\cmark	\\	
		seq-cover	&$	WKL_{\mu}\tnote{a}	$&	beam	&	list	&	\cmark	&	\cmark	&	\cmark	&	\cmark	&	\cmark	&	\cmark	\\	
		CN2-SD	&	entropy	&	beam	&	list	&	\cmark	&	\cmark	&	\cmark	&	-	&	-	&	-	\\	
		DSSD	&$	WKL_{\mu}\tnote{a}	$&	beam	&	set	&	-	&	\cmark	&	\cmark	&	\cmark	&	\cmark	&	\cmark	\\	
		MCTS4DM	&$	WKL_{\mu}\tnote{a}	$&	MCTS	&	set	&	-	&	\cmark	&	-	&	-	&	-	&	-	\\	
		FSSD	&$	WRAcc	$&	DFS	&	list	&	\cmark	&	\cmark	&	-	&	-	&	-	&	-	\\	
		\bottomrule																				
	\end{tabular}
	\begin{tablenotes}
\item [a] The algorithms only support $WKL_{\mu}$ for numeric targets (Eq.~\eqref{eq:wkl_mu}), i.e., a Weighted Kullback-Leibler divergency that only takes into account the mean, contrary to the one used by SSD++ that also uses the variance (Eq.~\eqref{eq:wkl_mu_sigma}). For the nominal target case there is only one WKL (the different WKL measures are explained in Section~\ref{sec:wkl}).
\end{tablenotes}
	
\end{threeparttable}
		
\end{table}

\textbf{Quality measures.} As the quality of a set is measured using the SWKL, the most appropriate measure to use is the Weighted Kullback-Leibler (WKL) for the algorithms that support it. CN2-SD supports entropy which is related to WKL. FSSD only supports WRAcc at the moment. Note that for the case of numeric targets, except SSD++, all use a WKL that only takes into account the mean, given by $WKL_{\mu}(s) = n_s/\hat{\sigma}_d (\hat{\mu_d}-\hat{\mu_s})^2$, in contrast to the deviation-aware measure of SSD++ in Eq.~\ref{eq:wkl_mu_sigma}.

\textbf{Hyperparameters.} Most algorithms use beam search, thus only have three main hyperparameters:  the maximum depth of search $d_{max}$; the width of the beam $w_b$; and the number of cut points to discretise numeric explanatory variables $n_{cut}$. The larger the values, the better the performance, but the slower the algorithms become, as time complexity is linear to each of them. To be fair and not over-search the hyperparameters, we selected the default values of the DSSD and seq-cover implementation for all beam-search algorithms: $d_{max} = 5$, $w_b = 100$, $n_{cut} =5$. For the case of MCTS4DM, which requires a larger set of hyperparameters, only the number of iterations is set, $n_{iter}= 50\,000$, to ensure good convergence, and the rest were set as default. FSSD only requires the maximum depth, which was set at $5$.

\subsubsection{Measures} 
To compare the quality of the subgroup sets obtained by different algorithms, we use three different measures. The first is our proposal to measure the overall quality of an ordered set of subgroups, the \emph{Sum of Weighted Kullback-Leibler} (SWKL), as defined in Eq.~\eqref{eq:swkl}. The other two are the \emph{number of subgroups} $|S|$ and the \emph{average number of conditions} per subgroup $|a|$, two commonly used measures for the interpretability/complexity of a set of rules. These two measures follow the law of parsimony and assume that fewer subgroups with fewer conditions are easier to understand by humans, which can be an invalid assumption in some situations. Nonetheless, it is widely used and its simple understanding typically makes for a good proxy~\citep{doshi2018considerations}. 

\paragraph{Generalisation.} In machine learning, algorithms are evaluated based on their generalisation to unseen data (e.g., cross-validation). This is not common practice in subgroup discovery and other algorithm implementations cannot run on unseen data. For this reason, we test against other algorithms in the same dataset. In terms of generalisation we compare SSD++ in its proposed format, versus SSD++ with KL and WKL divergence as quality measures instead of the greedy MDL gain (in Ln~\ref{alg:loop} of the SSD++), i.e., our formulation without distribution and model complexity, $\texttt{COMP}(n_a)$ and $L(M)$, respectively, in Eq~\eqref{eq:gain_kld}.


\subsubsection{Datasets}
For a thorough analysis we use a total of $54$ datasets---$10$-univariate binary; $10$ univariate nominal; $9$ multivariate nominal; $15$ univariate numeric; and $9$ multivariate numeric---that are listed in Tables~\ref{table:data_nominal} and~\ref{table:data_numeric} of Appendix~\ref{appendix:datasets}. The datasets are commonly used benchmarks of machine learning and subgroup discovery, which are publicly available from the UCI\footnote{\url{https://archive.ics.uci.edu/ml/}}, Keel\footnote{\url{http://www.keel.es/}}, and MULAN\footnote{\url{http://mulan.sourceforge.net/datasets.html}} repositories. The datasets were selected to be the most varied possible. In the case of the nominal target datasets in Table~\ref{table:data_nominal}, the number of targets ranging from $1$ to $374$, the classes from $2$ to $28$, the samples from $150$ to $45\:222$, and the variables from $3$ to $1\:186$. In the case of the numeric target datasets in Table~\ref{table:data_numeric}, the number of targets ranging from $1$ to $16$, the samples from $154$ to $22\:784$. Note that we used multi-label datasets instead of multi-nominal as the latter are not widely available.


\subsection{Nominal target results}\label{sec:empirical_nominal}

\begin{figure}[!b]
	\centering
	\begin{subfigure}[b]{\textwidth}
		\centering
		\includegraphics[width=\textwidth]{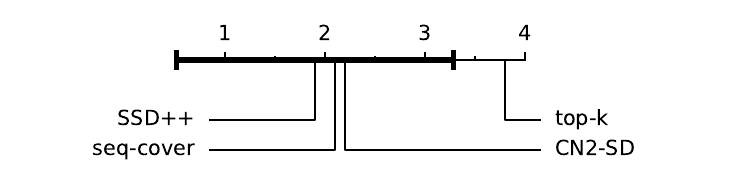}
		\caption{Single-binary targets}
		\label{fig:bonferroni-dunn_binary}
	\end{subfigure}
    \centering
	\begin{subfigure}[b]{\textwidth}
		\centering
		\includegraphics[width=\textwidth]{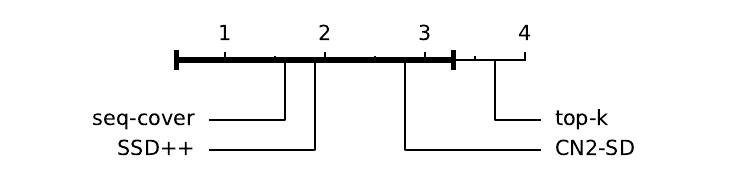}
		\caption{Single-nominal target.}
		\label{fig:bonferroni-dunn_nominal}
    \end{subfigure}
    \centering
	\begin{subfigure}[b]{\textwidth}
		\centering
		\includegraphics[width=\textwidth]{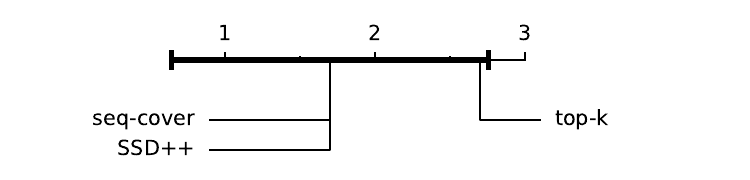}
		\caption{Multi-label targets}
		\label{fig:bonferroni-dunn_multi-label}
	\end{subfigure}
	\caption{Comparison of SSD++ against other algorithms for numeric targets datasets with the Bonferroni-Dunn test \citep{demvsar2006statistical} of the SWKL measure. The values shown represent the average ranking of the respective algorithms. Ranks outside the marked interval are significantly different---from a frequentist perspective---($p < 0.05$) from SSD++. Note that these graphs were added to help visualisation and the authors do not recommend inferring from the ``significance'' obtained. Moreover, the lack of significance was expected given the small number of datasets per target type}
	\label{fig:bonferroni-dunn_nominal_all}
\end{figure}

\begin{table*}[t!]\centering
\begin{threeparttable}[b]
\caption{Nominal target results. This includes single-binary, single-nominal, and multi-label, separated by horizontal lines in the table (top to bottom). The properties of the datasets can be seen in Table~\ref{table:data_nominal}, and are ordered in ascending number of: 1) target variables; 2) number of classes; and 3) number of samples. The evaluation measures are \{quality of the subgroup set swkl; number of subgroups $|S|$; and average number of conditions $|a|$\}. `avg. rank' stands for the average ranking for the respective target variable type, where $1$ represents the best rank. Note that CN2-SD does not work for multi-label case and thus the empty values $-$. }\label{table:results_nominal}																										
	\ra{1.1}
	\setlength{\tabcolsep}{5.1pt}
	 \begin{tabular}{@{}lrrrrrrrrrrrr@{}}\toprule												
		&	\multicolumn{3}{r}{top-$k$}					&	\multicolumn{3}{r}{seq-cover}					&	\multicolumn{3}{r}{CN2-SD}					&	\multicolumn{3}{r}{SSD++}					\\	\cmidrule(l){2-4} \cmidrule(l){5-7}\cmidrule(l){8-10}\cmidrule(l){11-13}
		datasets	&	swkl	&	$|S|$\tnote{a}	&	$|a|$	&\small	swkl	&	$|S|$	&	$|a|$	&\small	swkl	&	$|S|$	&	$|a|$	&\small	swkl	&	$|S|$	&	$|a|$	\\	\hline
		sonar 	&$	0.24	$&$	2	$&$	4	$&$\pmb{	0.96	}$&$	9	$&$	2	$&$	0.67	$&$	11	$&$	2	$&$	0.43	$&$	2	$&$	3	$\\	
		haberman 	&$	0.08	$&$	1	$&$	5	$&$\pmb{	0.39	}$&$	20	$&$	4	$&$	0.18	$&$	12	$&$	4	$&$	0.04	$&$	1	$&$	1	$\\	
		breastCancer 	&$	0.37	$&$	6	$&$	2	$&$	0.80	$&$	13	$&$	2	$&$	0.80	$&$	11	$&$	2	$&$\pmb{	0.82	}$&$	6	$&$	2	$\\	
		australian 	&$	0.26	$&$	5	$&$	3	$&$\pmb{	0.69	}$&$	13	$&$	3	$&$	0.54	$&$	24	$&$	3	$&$	0.55	$&$	5	$&$	2	$\\	
		tictactoe 	&$	0.50	$&$	16	$&$	3	$&$	0.73	$&$	18	$&$	3	$&$	0.21	$&$	21	$&$	3	$&$\pmb{	0.87	}$&$	16	$&$	2	$\\	
		german 	&$	0.08	$&$	4	$&$	5	$&$	0.30	$&$	22	$&$	4	$&$\pmb{	0.42	}$&$	48	$&$	4	$&$	0.14	$&$	4	$&$	3	$\\	
		chess 	&$	0.25	$&$	17	$&$	3	$&$	0.87	$&$	13	$&$	2	$&$	0.68	$&$	51	$&$	3	$&$\pmb{	0.97	}$&$	17	$&$	2	$\\	
		mushrooms 	&$	0.49	$&$	12	$&$	4	$&$	0.92	$&$	11	$&$	1	$&$\pmb{	1.00	}$&$	36	$&$	1	$&$\pmb{	1.00	}$&$	12	$&$	1	$\\	
		magic 	&$	0.16	$&$	69	$&$	5	$&$	0.38	$&$	35	$&$	4	$&$	0.42	$&$	616	$&$	3	$&$\pmb{	0.47	}$&$	69	$&$	4	$\\	
		adult	&$	0.11	$&$	103	$&$	5	$&$	0.27	$&$	79	$&$	4	$&$\pmb{	0.43}	$&$	1230	$&$	4	$&$	0.31	$&$	103	$&$	4	$\\[0.1cm]	
		avg. rank	&$	3.8	$&$	1.9	$&$	3.8	$&$	2.1	$&$	2.4	$&$	2.2	$&$	2.2	$&$	3.8	$&$	2.5	$&$\pmb{	1.9	}$&$	1.9	$&$	1.5	$\\ 	\midrule
		iris 	&$	0.53	$&$	4	$&$	2	$&$\pmb{	1.45	}$&$	5	$&$	2	$&$	0.96	$&$	4	$&$	2	$&$	1.44	$&$	4	$&$	1	$\\	
		balance	&$	0.21	$&$	9	$&$	3	$&$\pmb{	0.80	}$&$	19	$&$	3	$&$	0.18	$&$	3	$&$	3	$&$	0.69	$&$	9	$&$	3	$\\	
		CMC 	&$	0.07	$&$	7	$&$	3	$&$\pmb{	0.30	}$&$	38	$&$	4	$&$	0.27	$&$	42	$&$	3	$&$	0.25	$&$	7	$&$	2	$\\	
		page-blocks 	&$	0.19	$&$	21	$&$	5	$&$	0.45	$&$	26	$&$	2	$&$	0.44	$&$	12	$&$	4	$&$\pmb{	0.49	}$&$	21	$&$	3	$\\	
		nursery 	&$	0.92	$&$	81	$&$	2	$&$	1.36	$&$	22	$&$	3	$&$	0.87	$&$	8	$&$	4	$&$\pmb{	1.63	}$&$	81	$&$	3	$\\	
		automobile 	&$	0.38	$&$	5	$&$	4	$&$\pmb{	1.61	}$&$	11	$&$	3	$&$1.54$&$	7	$&$	4	$&$	1.25	$&$	5	$&$	2	$\\	
		glass 	&$	1.01	$&$	5	$&$	2	$&$	1.55	$&$	5	$&$	2	$&$\pmb{	2.14	}$&$	6	$&$	2	$&$	1.92	$&$	5	$&$	1	$\\	
		dermatology 	&$	0.54	$&$	9	$&$	2	$&$\pmb{	2.28	}$&$	9	$&$	2	$&$	2.12	$&$	7	$&$	3	$&$	2.11	$&$	9	$&$	2	$\\	
		kr-vs-k 	&$	0.45	$&$	351	$&$	5	$&$	0.75	$&$	43	$&$	4	$&$	0.20	$&$	61	$&$	5	$&$\pmb{	1.83	}$&$	351	$&$	3	$\\	
		abalone 	&$	0.26	$&$	16	$&$	5	$&$	0.62	$&$	29	$&$	4	$&$	0.60	$&$	49	$&$	3	$&$\pmb{	0.74	}$&$	16	$&$	2	$\\[0.1cm]	
		avg. rank	&$	3.7	$&$	2.4	$&$	3.0	$&$\pmb{	1.6	}$&$	3.0	$&$	2.2	$&$	2.8	$&$	2.3	$&$	3.4	$&$	1.9	$&$	2.4	$&$	1.4	$\\	\midrule
		emotions	&$	0.71	$&$	17	$&$	5	$&$	1.93	$&$	22	$&$	4	$&$	-	$&$	-	$&$	-	$&$\pmb{	2.68	}$&$	17	$&$	3	$\\	
		scene	&$	0.39	$&$	49	$&$	5	$&$	1.85	$&$	33	$&$	4	$&$	-	$&$	-	$&$	-	$&$\pmb{	3.05	}$&$	49	$&$	4	$\\	
		birds	&$	0.49	$&$	8	$&$	5	$&$\pmb{	2.02	}$&$	20	$&$	4	$&$	-	$&$	-	$&$	-	$&$	1.57	$&$	8	$&$	3	$\\	
		flags	&$	0.44	$&$	5	$&$	4	$&$\pmb{	2.40	}$&$	17	$&$	4	$&$	-	$&$	-	$&$	-	$&$	1.21	$&$	5	$&$	2	$\\	
		yeast	&$	0.49	$&$	35	$&$	5	$&$	1.83	$&$	55	$&$	5	$&$	-	$&$	-	$&$	-	$&$\pmb{	2.20	}$&$	35	$&$	5	$\\	
		genbase	&$	0.88	$&$	15	$&$	2	$&$	5.51	$&$	12	$&$	1	$&$	-	$&$	-	$&$	-	$&$\pmb{	5.82	}$&$	15	$&$	1	$\\	
		mediamill	&$	0.43	$&$	131	$&$	5	$&$	1.44	$&$	60	$&$	5	$&$	-	$&$	-	$&$	-	$&$\pmb{	2.96	}$&$	131	$&$	5	$\\	
		CAL500	&$	1.46	$&$	1	$&$	5	$&$\pmb{	16.91	}$&$	36	$&$	4	$&$	-	$&$	-	$&$	-	$&$	1.24	$&$	1	$&$	5	$\\	
		corel5k	&$	5.81	$&$	144\tnote{b}	$&$	3	$&$\pmb{	5.39	}$&$	144	$&$	4	$&$	-	$&$	-	$&$	-	$&$	0.00	$&$	0	$&$	0	$\\[0.1cm]	
		avg. rank	&$	2.7	$&$	1.9	$&$	2.7	$&$\pmb{	1.7	}$&$	2.3	$&$	1.9	$&$	-	$&$		$&$		$&$\pmb{	1.7	}$&$	1.8	$&$	1.4	$\\	
		\bottomrule																										
	\end{tabular}
	\begin{tablenotes}
\item [a] $k$ was selected as the number of subgroups found by SSD++.
\item [b] Seq-cover number of subgroups was used as a reference for this case.
\end{tablenotes}
	
	\end{threeparttable}
								
\end{table*}
The results obtained on binary, nominal, and multi-label datasets with sequential subgroup set miners can be seen in Table~\ref{table:results_nominal} and in a graphical representation for the SWKL measure in Figure~\ref{fig:bonferroni-dunn_nominal_all}, while the results for algorithms that output sets can be found in Table~\ref{table:results_nominal_nonseq} in Appendix~\ref{appendix:non-sequential}. Overall, we can see that our algorithm gets $14$ out of $29$ best results, compared with seq-cover in second place with $13$ best results. In terms of SWKL per type of data, SSD++ achieves the smallest ranking for binary, seq-cover for nominal, and both are tied for multi-nominal. This small difference in the results between SSD++ and seq-cover is important for two reasons. First, it validates SWKL, showing that seq-cover is already implicitly maximising it without knowing it. Second, it shows that SSD++ can obtain on par or slightly better results than other established approaches. Our non-diverse baseline, top-$k$, shows that covering different dataset regions is important to maximise SWKL. 

Regarding the number of found subgroups, we can see that in most cases, all algorithms are in the same order of magnitude, except when SSD++ obtains many more subgroups (for \emph{adult}, \emph{nursery}, \emph{kr-vs-k}, and \emph{mediamill}). These results can be explained by the use of normalised gain ($\beta=1$) by SSD++, together with the fact that these datasets have a large number of samples, few variables, or a large number of categories. First, let us recall that the normalised compression gain of Eq.~\eqref{eq:greedy_gain} is composed of a data covering part and a model penalisation part and that both are normalised by the number of instances covered, which gives an advantage to subgroups that cover less data but are well-covered (only one category, or few categories). When the datasets are larger and the number of variables is reasonably small, like \emph{adult} with $45\:222$ examples and $14$ variables, there is a larger chance of finding more statistically ``significant'' subgroups, as there can be more regions where subgroups only (or almost only) cover one class, and the penalisation of the model encoding is small as there are not many variables. On the other hand, subgroups covering more data can more easily have a larger entropy in the class label distribution. For example, \emph{kr-vs-k}, which is a reasonably large dataset with $28\,056$ and with $18$ class labels, a subgroup that only covers one class label, as opposed to covering many class labels, will have a higher chance of being chosen. The number of subgroups found can be large, but it was shown in a classification setting that they generalise well \citep{proencca2020interpretable}. It is interesting to note that in the case of \emph{corel-5k}, SSD++ does not find any ``significant'' subgroup to add.

Regarding the number of conditions per subgroup, the two best-performing algorithms in terms of SWKL, SSD++, and seq-cover, tend to have a similar and lower number of conditions than the other algorithms. As Top-$k$ only covers the same region, it tends to be close to the maximum depth of $5$.

\subsection{Numeric target results}\label{sec:empirical_numeric}

\begin{figure}[!b]
	\centering
	\begin{subfigure}[b]{\textwidth}
		\centering
		\includegraphics[width=\textwidth]{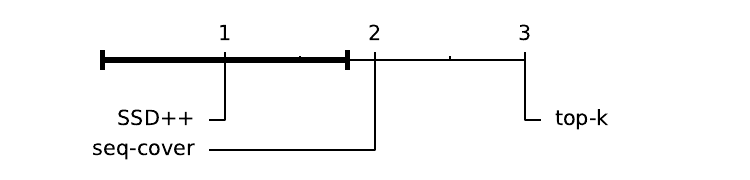}
		\caption{Single-numeric targets}
		\label{fig:bonferroni-dunn_numeric}
	\end{subfigure}
    \centering
	\begin{subfigure}[b]{\textwidth}
		\centering
		\includegraphics[width=\textwidth]{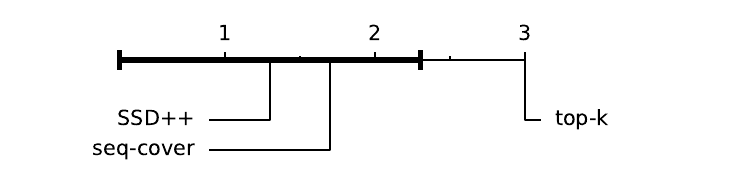}
		\caption{Multi-numeric targets}
		\label{fig:bonferroni-dunn_multi-numeric}
	\end{subfigure}
	\caption{Comparison of SSD++ against the rest with the Bonferroni-Dunn test \citep{demvsar2006statistical} for numeric targets for the SWKL measure. The values represent the average ranking of the respective algorithms and all algorithm with ranks outside the marked interval are significantly different---from a frequentist perspective---($p < 0.05$) from the SSD++. Note that this graphs were added to help visualisation and the authors do not recommend to infer from the ``significance'' obtained. Moreover, the lack of significance was expected given the small number of datasets per target type}
	\label{fig:bonferroni-dunn_numeric_all}
\end{figure}

The results for the single-target and multi-target numeric datasets can be seen in Table~\ref{table:results_numeric} and in a graphical representation for the SWKL measure in Figure~\ref{fig:bonferroni-dunn_numeric_all}. In general, SSD++ obtains the best results for $23$ out of $25$ datasets. This is expected as SWKL and SSD++ take into account the dispersion/deviation of the subgroup target while top-$k$ and seq-cover do not. Moreover,  the normalised standard deviation of the first subgroup found supports this claim, as SSD++ tends to find subgroups with smaller deviations for $10$ out of $15$ cases.

\begin{table*}[t!]\centering															\begin{threeparttable}[b]
											
	\caption{Numeric target results. This includes single-numeric and multi-numeric, separated by a horizontal line in the table (top to bottom). The properties of the datasets can be seen in Table~\ref{table:data_numeric}, and are ordered in ascending number of: 1) target variables; 2) number of classes; and 3) number of samples.  The evaluation measures are \{quality of the subgroup set swkl; number of subgroups $|S|$; normalised standard deviation of the first subgroup $\tilde{\sigma}_{t1}$; and average number of conditions $|a|$\}.  `avg. rank' stands for the average ranking for the respective target variable type, where $1$ represents the best ranking. Note that $\tilde{\sigma}_{t1}$ is not shown for the multi-numeric case as it is not easy to understand.}\label{table:results_numeric}																										
	\ra{1.1}
	\setlength{\tabcolsep}{5.3pt}
	\begin{tabular}{@{}lrrrrrrrrrrrr@{}}\toprule																										
		&	\multicolumn{4}{r}{top-$k$}							&	\multicolumn{4}{r}{seq-cover}							&	\multicolumn{4}{r}{SSD++}							\\	\cmidrule(l){2-5} \cmidrule(l){6-9}\cmidrule(l){10-13}
		datasets	&	swkl	&	$\tilde{\sigma}_{t1}$	&	$|S|$\tnote{a}	&	$|a|$	&	swkl	&	$\tilde{\sigma}_{t1}$	&	$|S|$	&	$|a|$	&	swkl	&	$\tilde{\sigma}_{t1}$	&	$|S|$	&	$|a|$	\\	\hline
		baseball 	&$	0.26	$&$	0.82	$&$	7	$&$	4	$&$	1.40	$&$	1.22	$&$	26	$&$	4	$&$\pmb{	1.86	}$&$	0.01	$&$	7	$&$	2	$\\	
		autoMPG8 	&$	0.43	$&$	0.54	$&$	8	$&$	4	$&$	1.45	$&$	1.85	$&$	22	$&$	4	$&$\pmb{	1.57	}$&$	0.18	$&$	8	$&$	2	$\\	
		dee 	&$	0.46	$&$	0.50	$&$	9	$&$	4	$&$	1.29	$&$	2.01	$&$	20	$&$	4	$&$\pmb{	1.35	}$&$	0.32	$&$	9	$&$	2	$\\	
		ele-1 	&$	0.29	$&$	1.06	$&$	8	$&$	4	$&$	1.14	$&$	0.94	$&$	22	$&$	4	$&$\pmb{	1.22	}$&$	1.24	$&$	8	$&$	2	$\\	
		forestFires 	&$	0.61	$&$	6.84	$&$	22	$&$	4	$&$	2.73	$&$	0.15	$&$	57	$&$	4	$&$\pmb{	3.91	}$&$	7.57	$&$	22	$&$	3	$\\	
		concrete 	&$	0.28	$&$	0.65	$&$	18	$&$	4	$&$	1.27	$&$	1.53	$&$	35	$&$	4	$&$\pmb{	1.31	}$&$	0.21	$&$	18	$&$	3	$\\	
		treasury 	&$	0.43	$&$	0.68	$&$	31	$&$	4	$&$	2.74	$&$	1.46	$&$	21	$&$	4	$&$\pmb{	3.85	}$&$	0.01	$&$	31	$&$	2	$\\	
		wizmir 	&$	0.70	$&$	0.31	$&$	22	$&$	4	$&$	2.15	$&$	3.22	$&$	26	$&$	4	$&$\pmb{	2.72	}$&$	0.15	$&$	22	$&$	2	$\\	
		abalone 	&$	0.23	$&$	0.59	$&$	26	$&$	4	$&$	0.47	$&$	1.68	$&$	126	$&$	5	$&$\pmb{	0.71	}$&$	1.32	$&$	26	$&$	3	$\\	
		puma32h 	&$	0.55	$&$	0.59	$&$	48	$&$	4	$&$	1.39	$&$	1.68	$&$	70	$&$	5	$&$\pmb{	1.44	}$&$	0.29	$&$	48	$&$	3	$\\	
		ailerons 	&$	0.24	$&$	1.23	$&$	98	$&$	4	$&$	1.04	$&$	0.82	$&$	105	$&$	4	$&$\pmb{	1.44	}$&$	0.98	$&$	98	$&$	4	$\\	
		elevators 	&$	0.23	$&$	1.44	$&$	158	$&$	4	$&$	0.83	$&$	0.69	$&$	150	$&$	5	$&$\pmb{	1.31	}$&$	1.40	$&$	158	$&$	4	$\\	
		bikesharing	&$	0.26	$&$	1.09	$&$	136	$&$	4	$&$	1.24	$&$	0.92	$&$	91	$&$	4	$&$\pmb{	1.70	}$&$	0.02	$&$	136	$&$	4	$\\	
		california 	&$	0.19	$&$	0.90	$&$	174	$&$	4	$&$	0.69	$&$	1.11	$&$	116	$&$	5	$&$\pmb{	1.14	}$&$	0.00	$&$	174	$&$	4	$\\	
		house 	&$	0.19	$&$	1.59	$&$	269	$&$	4	$&$	0.91	$&$	0.63	$&$	143	$&$	5	$&$\pmb{	2.02	}$&$	2.83	$&$	269	$&$	5	$\\[0.1cm]	
		avg. rank	&$	3.0	$&$	2.1	$&$	1.8	$&$	2.0	$&$	2.0	$&$	2.3	$&$	2.3	$&$	2.7	$&$\pmb{	1.0	}$&$	1.6	$&$	1.8	$&$	1.3	$\\	\midrule
		edm	&$	0.47	$&$	-	$&$	5	$&$	5	$&$	0.81	$&$	-	$&$	9	$&$	2	$&$\pmb{	1.88	}$&$	-	$&$	5	$&$	2	$\\	
		enb	&$	2.73	$&$	-	$&$	41	$&$	2	$&$	3.54	$&$	-	$&$	19	$&$	2	$&$\pmb{	8.71	}$&$	-	$&$	41	$&$	2	$\\	
		slump	&$	1.38	$&$	-	$&$	4	$&$	5	$&$\pmb{	2.74	}$&$	-	$&$	17	$&$	4	$&$	2.57	$&$	-	$&$	4	$&$	3	$\\	
		sf1	&$	0.16	$&$	-	$&$	3	$&$	5	$&$\pmb{	2.06	}$&$	-	$&$	47	$&$	4	$&$	1.24	$&$	-	$&$	3	$&$	3	$\\	
		sf2	&$	0.86	$&$	-	$&$	2	$&$	5	$&$	2.29	$&$	-	$&$	18	$&$	4	$&$\pmb{	0.91	}$&$	-	$&$	2	$&$	4	$\\	
		jura	&$	0.47	$&$	-	$&$	15	$&$	5	$&$	2.38	$&$	-	$&$	28	$&$	4	$&$\pmb{	3.52	}$&$	-	$&$	15	$&$	3	$\\	
		osales	&$	2.17	$&$	-	$&$	45	$&$	4	$&$	18.09	$&$	-	$&$	48	$&$	3	$&$\pmb{	26.44	}$&$	-	$&$	45	$&$	3	$\\	
		oes97	&$	6.55	$&$	-	$&$	16	$&$	3	$&$	30.79	$&$	-	$&$	19	$&$	4	$&$\pmb{	34.36	}$&$	-	$&$	16	$&$	4	$\\	
		oes10	&$	6.56	$&$	-	$&$	23	$&$	3	$&$	29.11	$&$	-	$&$	27	$&$	4	$&$\pmb{	40.65	}$&$	-	$&$	23	$&$	3	$\\	
		wq	&$	0.87	$&$	-	$&$	62	$&$	5	$&$	2.06	$&$	-	$&$	47	$&$	4	$&$\pmb{	11.14	}$&$	-	$&$	62	$&$	4	$\\[0.1cm]	
		avg. rank	&$	3.0	$&$	-	$&$	1.7	$&$	2.4	$&$	1.7	$&$	-	$&$	2.6	$&$	1.8	$&$\pmb{	1.3	}$&$	-	$&$	1.7	$&$	1.8	$\\	
		\bottomrule																										
	\end{tabular}																		
\begin{tablenotes}
\item [a] $k$ was selected as the number of subgroups found by SSD++.	
\end{tablenotes}
\end{threeparttable}
\end{table*}

Comparing SWKL results for top-$k$ with seq-cover and SSD++ shows that irrespective of dispersion-aware (SSD++) or not (seq-cover), covering different regions of the data increases the quality of the list in terms of SWKL, validating the use of our measure.
It should be noted that both top-$k$ and seq-cover could, in practice, support taking into account the deviation, but that would require several non-trivial modifications in their source code. 

Regarding the number of subgroups, seq-cover tends to have more rules than SSD++ for datasets with less than $5\:000$ examples, while SSD++ tends to have more for a larger number of examples. This makes sense as there is more evidence to identify possible significant subgroups. 

Regarding the number of antecedents, SSD++ tends to have, on average, one condition fewer than seq-cover for single-target and a similar number for the multi-target case.

\subsection{Statistical robustness and generalisation}\label{sec:statistica_robustness}

\begin{figure}[!htb]
	\centering
	\begin{subfigure}[b]{\textwidth}
		\centering
		\includegraphics[width=\textwidth]{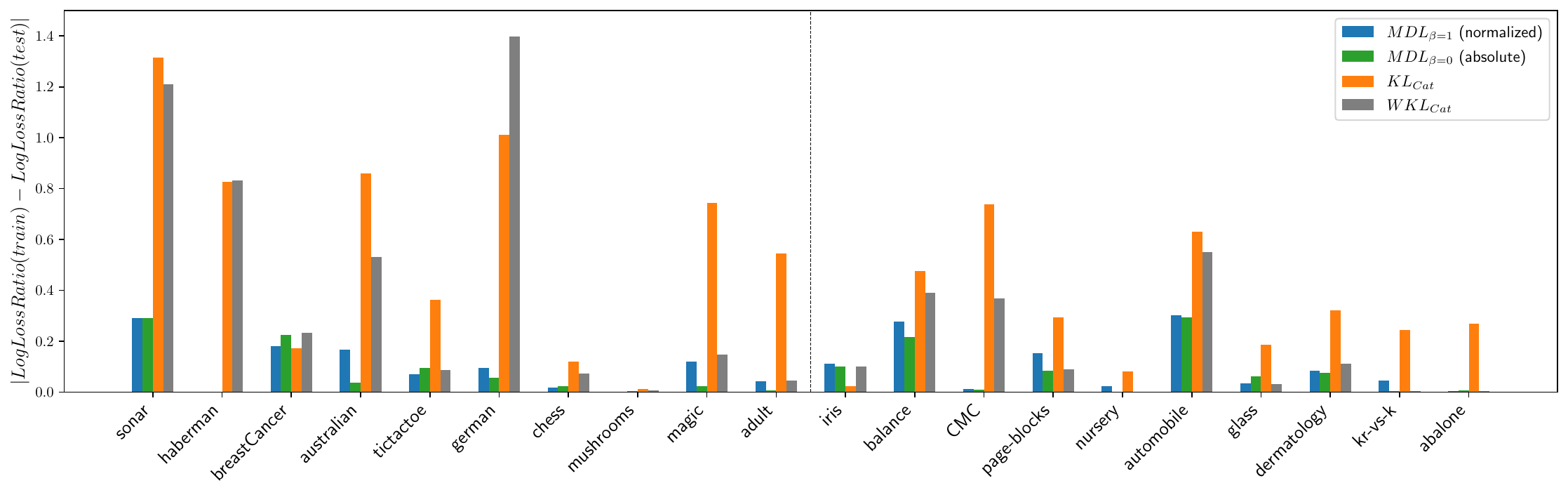}
		\caption{Single-binary and single-nominal targets}
		\label{fig:results_generalisation_nominal}
		\vspace{0.5cm}
	\end{subfigure}
    \centering
	\begin{subfigure}[b]{\textwidth}
		\centering
		\includegraphics[width=\textwidth]{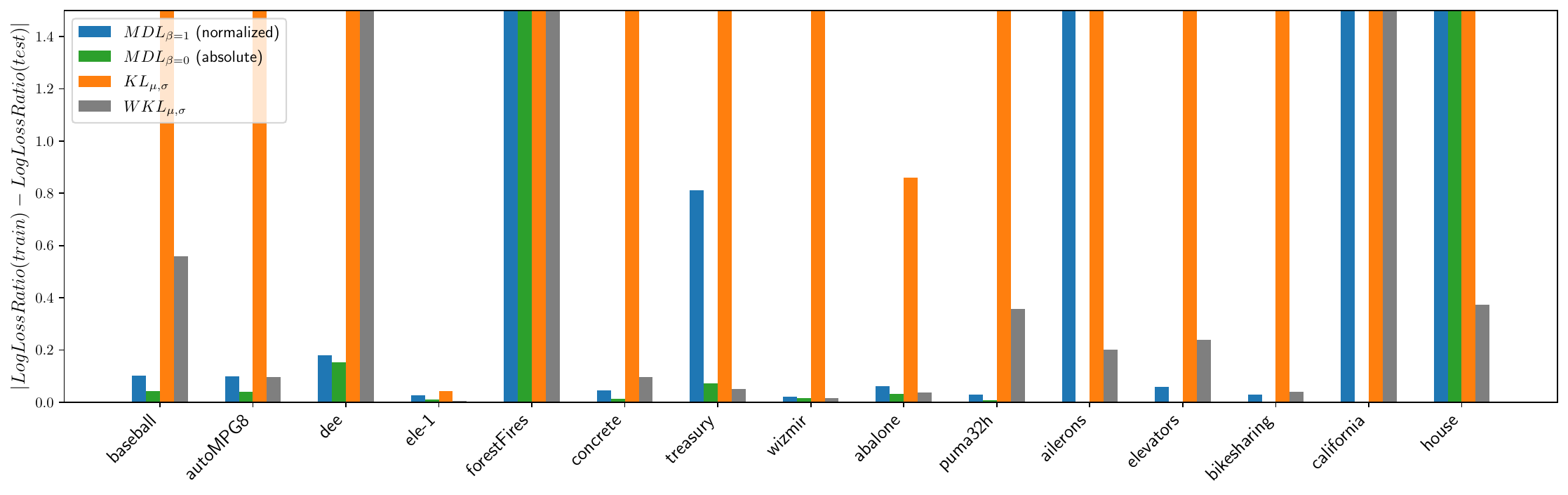}
		\caption{Single-numeric targets}
		\label{fig:results_generalisation_numeric}
	\end{subfigure}
	\caption{Statistical robustness analysis for nominal and numeric target datasets. The figures show how subgroup lists obtained with the proposed approach---SSD++ with normalised and absolute gain, i.e., $MDL_{\beta= 1}$ and $MDL_{\beta= 0}$---generalise to unseen data, tested with a $50\%$--$50\%$ train--test split. As baselines, we used SSD++ with KL and WKL divergence as quality measures, i.e., the same as our approach but without accounting for multiple hypothesis testing and distribution complexity (that is, without $L(M)$ and $\texttt{COMP}(n_a) $ in Eq.~\eqref{eq:gain_kld}). The properties of the datasets can be seen in Table~\ref{table:data_nominal}, and are ordered in ascending number of: 1) number of classes (if nominal); and 2) number of samples. The evaluation measure is $| LogLossRatio(train) -LogLossRatio(test)|$, where LogLossRatio(*) is the ratio between the log loss measure of the obtained subgroup list and the log loss of the dataset marginal distribution (dataset rule). The lower the value of the measure, the better, as this indicates that the model generalises well and does not overfit on the training set. The complete results can be seen in Tables~\ref{table:results_generalisation_nominal} and~\ref{table:results_generalisation_numeric} in Appendix~\ref{appendix:statistical_robustness}}
	\label{fig:results_generalisation}
\end{figure}

The main results of the statistical robustness analysis for single-binary, -nominal, and -numeric targets are shown in Figures~\ref{fig:results_generalisation_nominal} and \ref{fig:results_generalisation_numeric}, while the complete results can be seen in Tables~\ref{table:results_generalisation_nominal} and \ref{table:results_generalisation_numeric} of Appendix~\ref{appendix:statistical_robustness}. Only single-target datasets are used for an easier interpretation of the results. Our proposed formulations---$\mathrm{MDL}_{\beta= 1}$ (normalised gain) and $\mathrm{MDL}_{\beta= 0}$ (absolute gain)---are compared against KL and WKL divergence, i.e., their counterparts, that do not take into account distribution complexity and multiple-hypothesis testing in Eq.~\eqref{eq:gain_kld}. These counterpart `non-testing' versions of SSD++ are similar, in the essential parts, to the state-of-the-art seq-cover algorithm. To compare the different quality measures used, we use the difference between log loss in train and test sets\footnote{For the formal definition of the difference between log loss in train and test, please refer to Eq.~\eqref{eq:logloss} and \eqref{eq:loglossratio} in Appendix~\ref{appendix:statistical_robustness}.} with a $50\%$--$50\%$ train--test split. The log loss measures how well the estimated probabilities model the distribution present in the data (a lower value is better), so that the difference between log losses measures how well the model generalises to unseen data. 

At first glance, the two MDL-based approaches achieve the best generalisation (difference of log losses), lower log loss on the test set, and the smallest number of subgroups in $86\%$, $77\%$, and $100\%$ of the cases, respectively, when compared to their `non-testing' counterparts. 

It is interesting to observe that, on the one hand, $\mathrm{MDL}_{\beta= 0}$ obtains the lowest difference of log losses in $69\%$ of the cases. On the other hand, however, $\mathrm{MDL}_{\beta= 1}$ has the best test set log loss overall, obtaining the best value $54\%$ of the cases. This difference is expected,  as $\mathrm{MDL}_{\beta= 0}$ produces more conservative subgroup lists in terms of log loss in the train set, which is reflected by a lower generalisation error. Also, while  $\mathrm{MDL}_{\beta= 1}$ has the best test set performance, its counterpart ($KL$) has the worst, showing that our MDL formulation adds statistical robustness to well-known existing measures.

Further, we observe test log losses with large or infinite values in the numeric target case. This happens when one subgroup with a small variance sees a point far from its mean in the test set. For the case of $\mathrm{MDL}_{\beta= 1}$ and $WKL_{\mu,\sigma}$, the infinite values appear in few (one to two) subgroups in their lists, not making it a problematic behaviour for description; however, for $KL_{\mu,\sigma}$ this happens for most subgroups found.

\subsection{Runtime comparison}\label{sec:empirical_time}

\begin{figure}[!htb]
	\centering
	\begin{subfigure}[b]{0.48\textwidth}
		\centering
		\includegraphics[width=\textwidth]{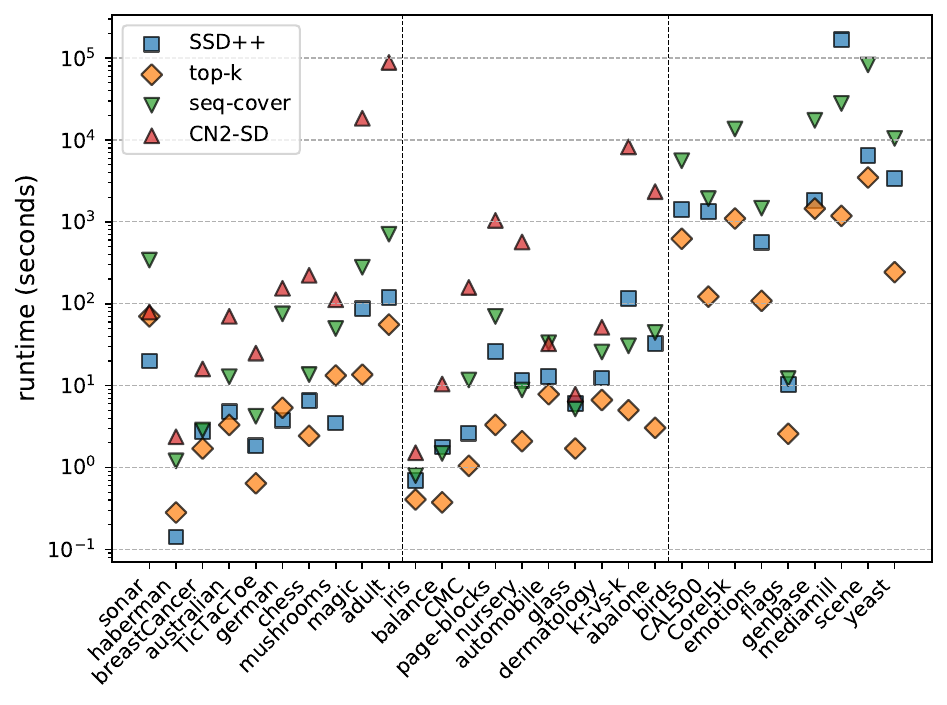}
		\caption{Nominal targets}
		\label{fig:runtime_nominal}
	\end{subfigure}
	\hfill
	\begin{subfigure}[b]{0.48\textwidth}
		\centering
		\includegraphics[width=\textwidth]{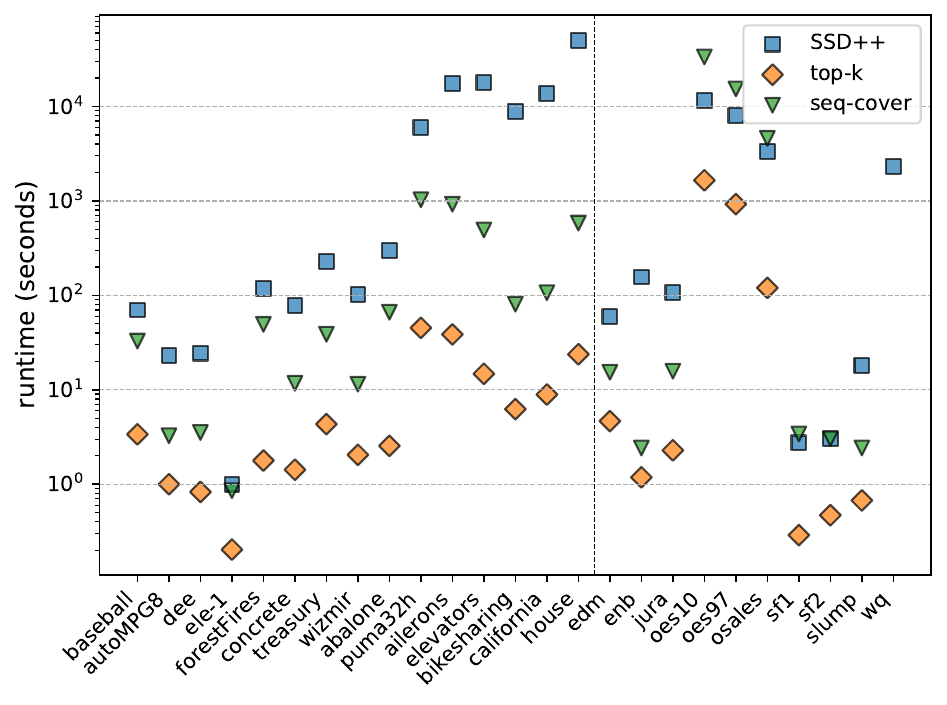}
		\caption{Numeric targets}
		\label{fig:runtime_numeric}
	\end{subfigure}
	\caption{Runtime in seconds for all algorithms for each dataset. The black vertical line divides the type of datasets, i.e., from left to right: univariate binary, nominal, and multi-label for nominal targets, and univariate and multivariate for numeric}
	\label{fig:runtime}
\end{figure}

Runtimes of all algorithms compared, i.e., top-$k$, seq-cover, CN2-SD, and SSD++ are shown in Figures~\ref{fig:runtime_nominal} and \ref{fig:runtime_numeric}. In general, the runtime increases with the number of samples in the dataset for a fixed data type. For the nominal datasets, there is an increase in runtime with the number of target variables, which does not seem to happen for numeric targets. This is because the number of subgroups found for multivariate numeric targets was, in general, smaller.

Comparing the algorithms against each other, as expected, top-$k$ was the fastest algorithm, as it only needs to search for the subgroups once, while the others need multiple iterations. 

For nominal targets, CN2-SD was the slowest algorithm, which stems from entropy as a quality measure---experiments with WRAcc proved to be much faster. On the other hand, SSD++ seems to perform on par with seq-cover and is often even faster.

For numeric targets, SSD++ was one order of magnitude slower than seq-cover. One possible reason is the extra time to compute the variance, although this does not explain the difference between both algorithms. A further study of the numeric implementation could make for an interesting research direction.  

\clearpage

%
%

\section{Case study: associations between socioeconomic background and university grades of Colombia engineering students}
\label{sec:case_study}

In this section, we apply SSD++ to a real use case to assess its usefulness and limitations. To this end, we aim to understanding how socioeconomic factors affect the grades of engineering university students in Colombia on their national exams. The dataset used to study this is fully described by \cite{delahoz2020dataset}. It contains socioeconomic variables and grades in national exams done at the high school and university level for engineering students in Colombia. For our specific case study, we have selected two of their exam grades at the university for two reasons. First, the relationship between socioeconomic variables and university grades is weaker (than for high school grades), thus more interesting to see if we can find relations, and second, only having two exam grades improves the visualisation of the results.

\paragraph{Dataset.} The dataset used is composed of $12\:412$ samples, $22$ explanatory variables, and $2$ numeric target variables. The explanatory variables refer to the socioeconomic background of the students at the time of high school, and they are made of variables such as parent' level of education, the household income, which type of high school they attended, the utilities available at home (e.g., internet and television), and their neighbourhood stratum\footnote{Stratum is a classification system unique to Colombia, where districts are ranked based on their affluence level from $1$ to $6$, where $1$ is the lowest level \url{https://www.dane.gov.co/index.php/servicios-al-ciudadano/servicios-informacion/estratificacion-socioeconomica} (Accessed on 29 June. 2022).}. The numeric targets represent their grades, from $0 \%$ to $100\%$, in two national university-level exams, namely quantitative reasoning and English. 

An additional reason for selecting this dataset is that it violates two of our model assumptions: $1)$ the target variables values are truncated between $0$ and $100$, thus violating the use of a continuous normal distribution to describe them; and $2)$ the target variables are not independent, as suggested by a correlation of $53\%$. If our approach is shown to work despite these violations, we may consider this is a good result.

\subsection{Analysis of the subgroups obtained with SSD++}

\begin{figure}[!htb]
	\centering
	\begin{subfigure}[b]{0.48\textwidth}
		\centering
		\includegraphics[width=\textwidth]{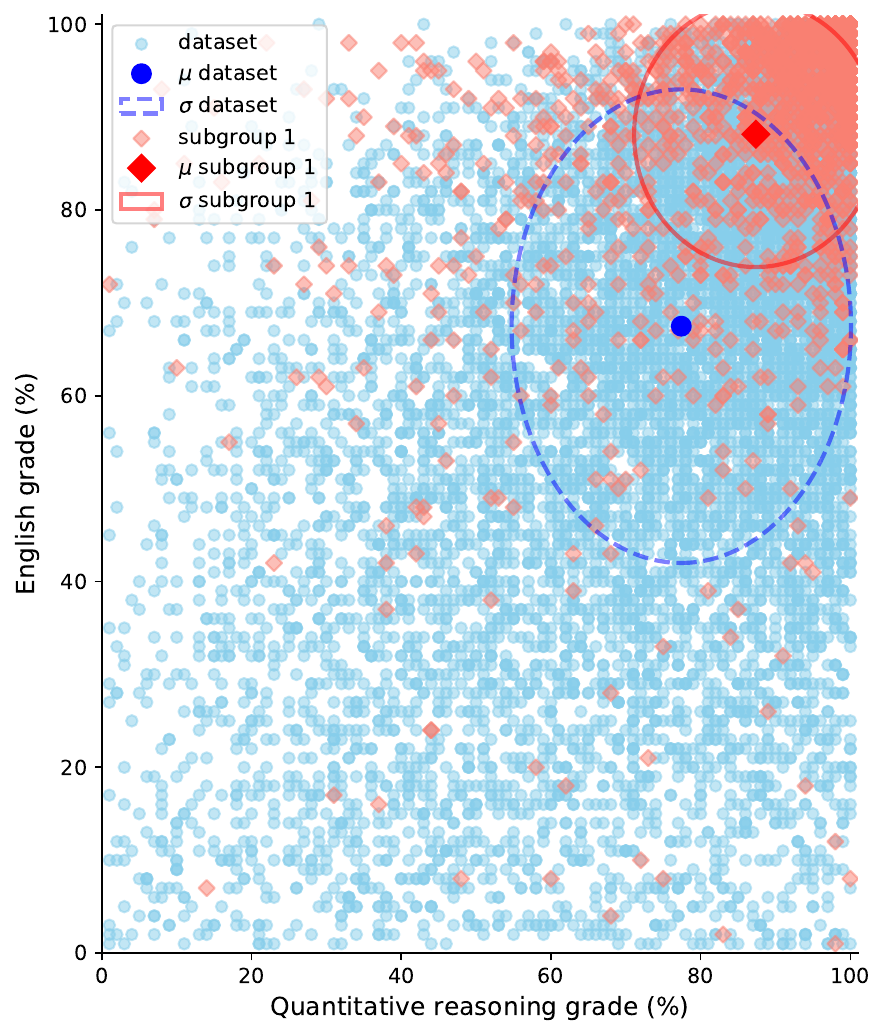}
		\caption{$1^{st}$ subgroup with \emph{absolute} gain}
		\label{fig:1st_subgroup_absolute}
	\end{subfigure}
	\hfill
	\begin{subfigure}[b]{0.48\textwidth}
	\centering
	\includegraphics[width=\textwidth]{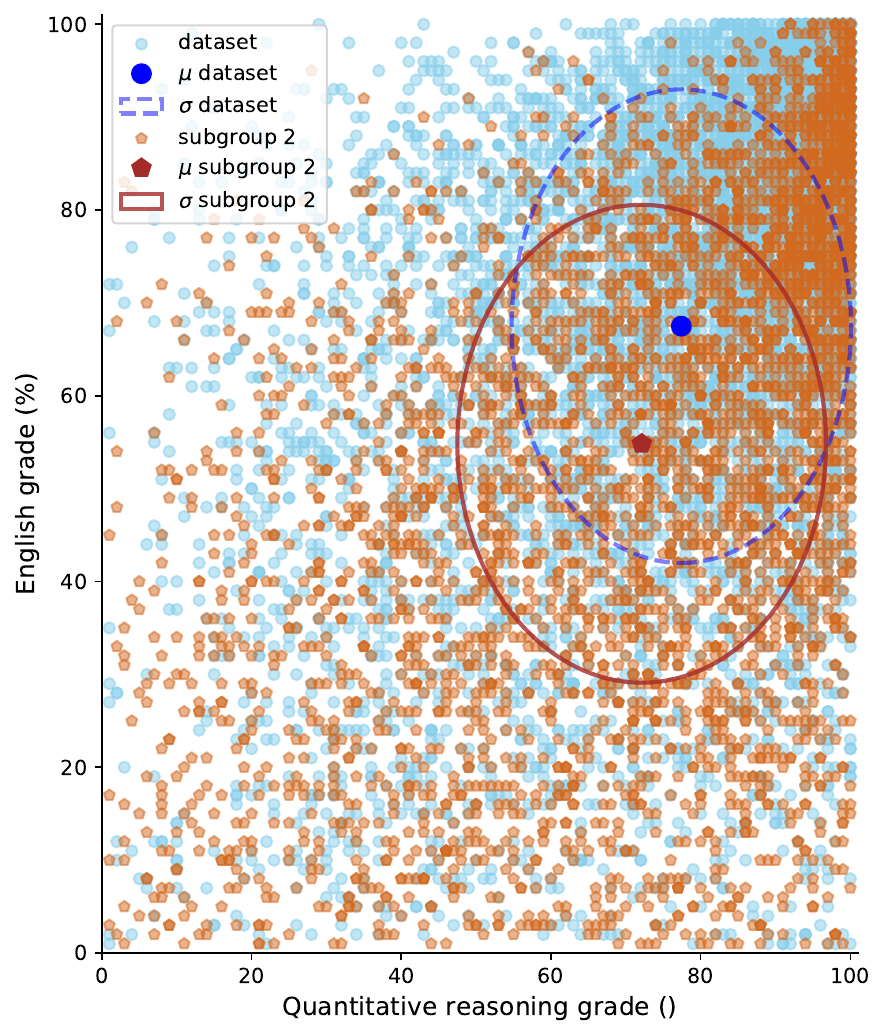}
	\caption{$2^{nd}$ subgroup with \emph{absolute} gain}
	\label{fig:2nd_subgroup_absolute}
	\end{subfigure}
	\\
	\begin{subfigure}[b]{0.48\textwidth}
	\centering
	\includegraphics[width=\textwidth]{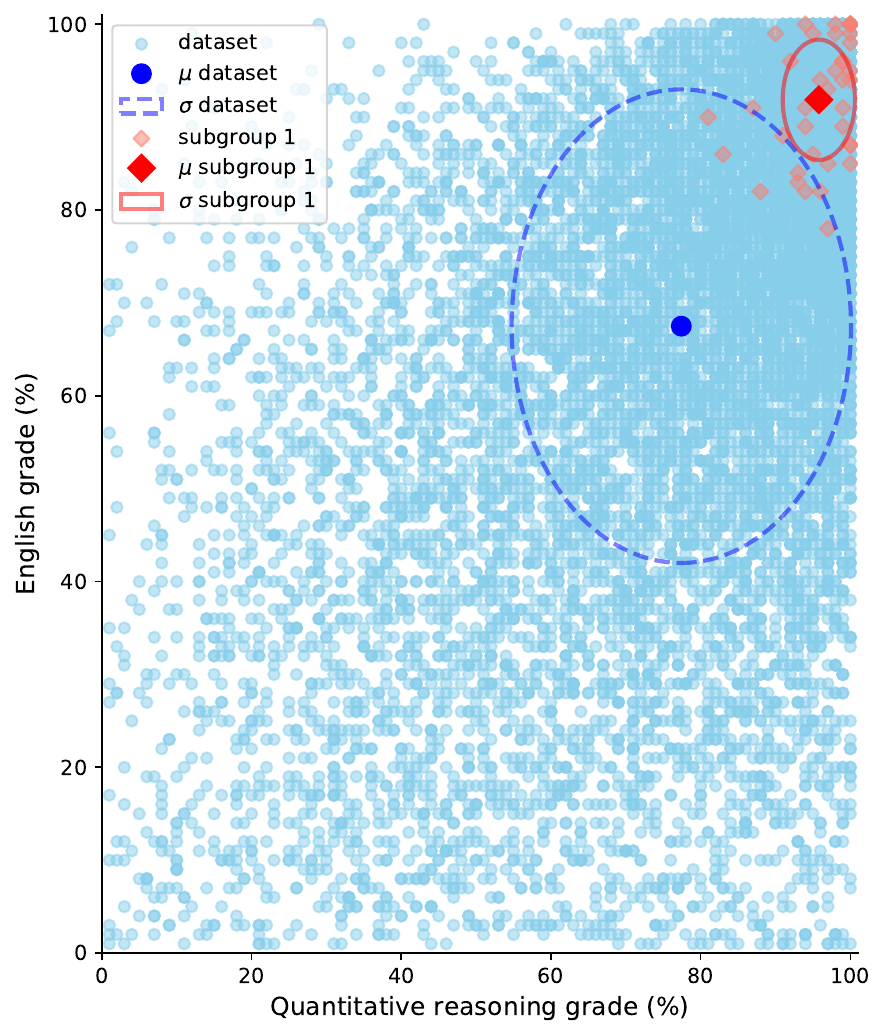}
	\caption{$1^{st}$ subgroup with \emph{normalised} gain}
	\label{fig:1st_subgroup_normalized}
	\end{subfigure}
	\hfill
	\begin{subfigure}[b]{0.48\textwidth}
	\centering
	\includegraphics[width=\textwidth]{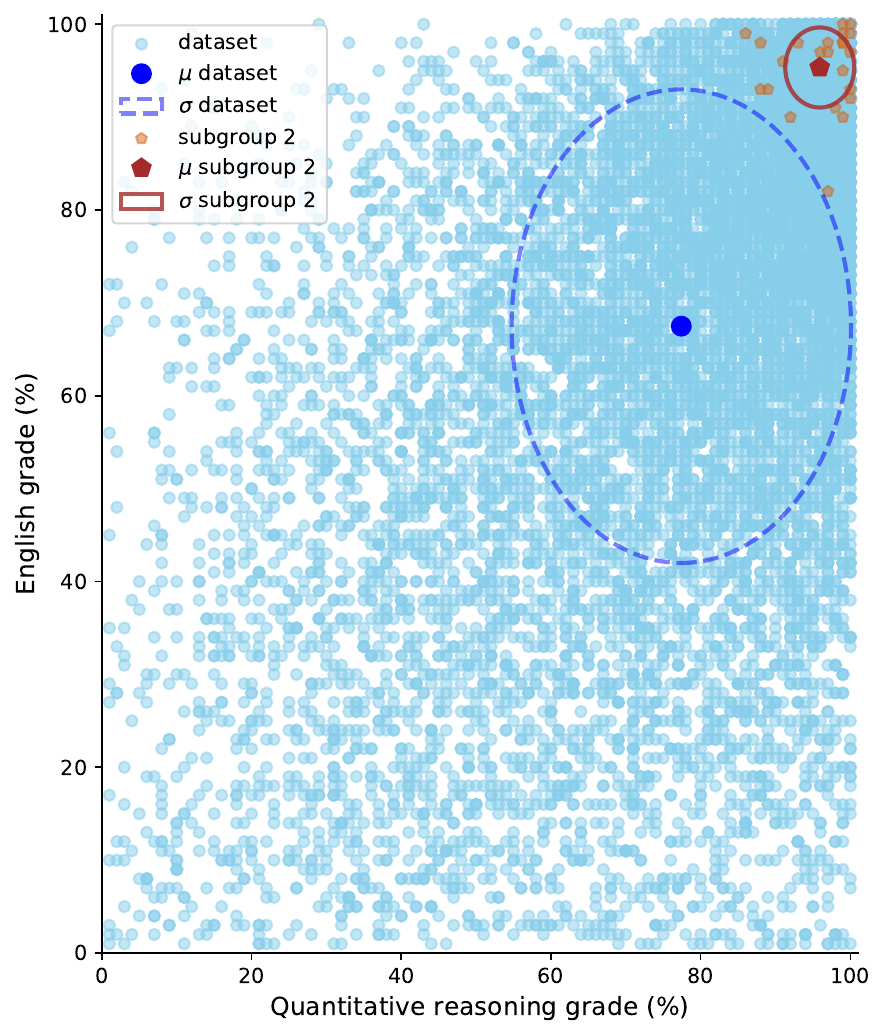}
	\caption{$2^{nd}$ subgroup with \emph{normalised} gain}
	\label{fig:2nd_subgroup_normalized}
	\end{subfigure}
	\caption{Scatter plot of the grades of students for Quantitative Reasoning and English exam, together with the grades associated with the descriptions of the $1^{st}$ and $2^{nd}$ with \emph{absolute} and \emph{normalised} gain}
	\label{fig:students_subgroup}
\end{figure}

The first four subgroups with absolute ($\beta=0$) and normalised ($\beta=1$) gain can be seen in Figures~\ref{fig:student_absolute} and \ref{fig:student_normalized}, respectively. The distributions of the first two subgroups for both gains can be seen in Figures~\ref{fig:1st_subgroup_absolute}, \ref{fig:2nd_subgroup_absolute}, \ref{fig:1st_subgroup_normalized}, and \ref{fig:2nd_subgroup_normalized}. The two extreme gains were used to show the interest (from a user perspective) of using different gains depending on the goal of the data exploration, i.e., coarse versus fine-grained perspective.

\paragraph{Comparison of absolute and normalised gain.} Overall, with absolute and normalised gain, our method finds $7$ and $34$ subgroups that cover a total of $84\%$ and $92\%$ of the data, respectively. Looking at Figures ~\ref{fig:1st_subgroup_absolute}, \ref{fig:2nd_subgroup_absolute}, \ref{fig:1st_subgroup_normalized}, and \ref{fig:2nd_subgroup_normalized}, it can be seen that normalised gain favours smaller and compact subgroups that deviate more from the dataset distribution, while absolute gain favours larger subgroups that deviate less from the dataset distribution. These conclusions can be verified by noting that normalised gain subgroups tend to have a smaller standard deviation, between $5\%$ and $9\%$, while absolute gain has values in the same order of magnitude of the dataset distribution, i.e., around $23\%$.

\paragraph{Interpretation of the results.} Both normalised and absolute gain results show that having a `better' socioeconomic background is associated with higher average grades in both exams, and the contrary is associated with lower grades. This is clearer in the absolute gain case, as each subgroup covers more data. It is noticeable in Figure \ref{fig:2nd_subgroup_absolute} that a subgroup with a standard deviation similar to the dataset leads to subgroups that are spread throughout the whole range of values. Nonetheless, that subgroup covers more regions with lower grades than the dataset, making it a relevant result to understand the dataset better.

In general, it can be seen that some conditions often appear in the subgroups, such as $household \_ income$ above and below $5$ minimum wages and education of one of the parents equal or above high school. It seems that the presence or absence of these variables is highly associated with above or below-average performance, respectively. 

Looking at specific subgroups, it is interesting to see that in the $4^{th}$ subgroup of the absolute gain, the Quantitative reasoning grade is equal to the average behaviour of the dataset ($77\%$), while the English grade is $8\%$ above average. Looking at the subgroups with normalised gain, we see that there are only slight variations of their descriptions and that they belong to a similar socioeconomic macro group but with slight differences in their descriptions, which corresponds to small differences in their grades distribution. 

\paragraph{Violation of the model assumptions.} Here, we can observe how our method behaves when some modelling assumptions are violated. Regarding the truncated values, it seems that the normalised gain is affected by grades around $100$ (as seen in Figures \ref{fig:1st_subgroup_normalized} and \ref{fig:2nd_subgroup_normalized}) as most of its subgroups capture these students, which increases the average and lowers the standard deviation, making them rank higher. 
Our method was not developed for highly stratified target values, but the results seem to show that it does not seem prohibitive to the use of SSD++ in these cases as long as the stratification is mild and the user takes into account this fact.

Regarding the independence assumption, it seems that the subgroups found are still relevant, although both grades are almost always taken into account together, i.e., as the values are positively correlated, it is more likely to find subgroups with mean values that are high or low for both exams, but not high for one and low for the other. This is expected as the encoding of independent normal distributions does not take into account the covariance between target variables, and thus that case is not deemed a deviation by the current model formulation.

\begin{figure}[!hbt]\centering
	\begin{subfigure}[b]{\textwidth}													
		\ra{1.0}
		\setlength{\tabcolsep}{6pt}
		\begin{tabular}{@{}llrrr@{}}												
			$s$	&	\textbf{description} of a student socioeconomic background	&	$n_s$	&	Quant.	&	English		\\ 	\midrule
			1	&	household\_income $\geq 5$ min. wage  \& public\_school = no   	&$	1676	$&$	87 \pm 16	$ &$	88 \pm 14	$ 	\\ 	
			&	  \& edu\_mother $>$ high\_school   \& Microwave = yes	&		&		&			\\ 	\cmidrule(l){2-5}
			2	&	household\_income $< 5$ min. wage  	&$	4031	$&$	72 \pm 25	$&$	54 \pm 26	$	\\ 	
			&	  \& stratum $<5$ \& public\_school = yes	&		&		&			\\ 	\cmidrule(l){2-5}
			3	&	gender = M \& edu\_father $\geq$ high\_school	&$	1478	$&$	85 \pm 17	$&$	78 \pm 20	$	\\ 	
			&	\& social\_support = None  \& stratum $>3$	&		&		&			\\ 
			&	\& public\_school = no								&		&		&			\\ \cmidrule(l){2-5}
			4	&	social\_support = None  \& edu\_father  $>$ high\_school   	&$	997	$&$	77 \pm 22	$&$	76 \pm19	$	\\ 	
			&	 \& public\_school = no \& internet = yes	&		&		&			\\ 	
			&	 \& mobile $=$ yes	&		&		&			\\ 	
			&	$\vdots$	&		&		&			\\ 	\midrule
			\multicolumn{2}{@{}l}{dataset distribution}			&$	1945^*	$&$	77 \pm 23	$&$	68 \pm 26	$	\\ 	\bottomrule
		\end{tabular}												
		\caption{Subgroup list with absolute gain ($\beta = 0$). First $4$ subgroups of a total of $7$ and swkl $=0.41$}\label{fig:student_absolute}
		\bigskip											
	\end{subfigure}													
	
	\begin{subfigure}[b]{\textwidth}													
		\ra{1.0}
		\setlength{\tabcolsep}{4.9pt}
		\begin{tabular}{@{}llrrr@{}}												
			$s$	&	\textbf{description} of a student socioeconomic background	&	$n_s$	&	Quant.	&	English		\\ 	\midrule
			1	&	household\_income $\geq 5$ min. wage \&  gender = M  \& 	&$	39	$&$	96 \pm 5	$ &$	92 \pm 6	$ 	\\ 	
			&	household\_size $< 3$ \& edu\_father $>$ high-school 	&		&		&			\\ 	
			&	\& mobile = yes	&		&		&			\\ 	\cmidrule(l){2-5}
			2	&	household\_income $\geq 5$ min. wage 	&$	23	$&$	96 \pm 5	$&$	95 \pm 4	$	\\ 	
			&	\& school\_type = academic \& occ.\_mother = retired	&		&		&			\\ 	
			&	\& edu\_father $\geq$ Undergrad	&		&		&			\\ 	\cmidrule(l){2-5}
			3	&	household\_income $\geq 5$ min. wage  	&$	30	$&$	96 \pm 5	$&$	93 \pm 6	$	\\ 	
			&	\& job\_mother = independent  \& stratum $\geq 4$ \& gender = M	&		&		&			\\ 	
			&	\& job\_father = independent &		&		&			\\ 	\cmidrule(l){2-5}
			4	&	job\_mother = executive \& stratum $\geq 4$  \& mobile = yes	&$	32	$&$	93 \pm 9	$&$	94 \pm 6	$	\\ 	
			&	 \& job\_father = independent \& public\_school = no	&		&		&			\\ 	
			&	$\vdots$	&		&		&			\\ 	\midrule
			\multicolumn{2}{@{}l}{dataset distribution}			&$	942^*	$&$	77 \pm 23	$&$	68 \pm 26	$	\\ 	\bottomrule
		\end{tabular}												
		\caption{Subgroup list with normalized gain ($\beta = 1$). First $4$ subgroups of a total of $34$ and swkl $=0.52$}\label{fig:student_normalized}
		\smallskip										
	\end{subfigure}													
	\caption{\emph{Colombia engineering students} performance in Quantitative Reasoning and English  exams. The results of Fig.~\ref{fig:student_absolute} and \ref{fig:student_normalized} were obtained by SSD++ with \emph{absolute gain} ($\beta = 0$) and \emph{normalised gain} ($\beta=1$). The dataset contains two numeric target variable \emph{Quantitative Reasoning} and \emph{English} exams in a 0-100\% scale. The dataset represents $12\:412$ engineering students in Colombia, their grades in university national exams and their social-economic background. \emph{Description} contains information regarding students socio-economic background, $n_s$ the number of instances covered, Quant. and English the average grade and standard deviation in the respective exams. $^*$ The $n$ of the dataset is the total number of instances in the dataset}\label{fig:student_example}
\end{figure}

\clearpage

%
%

\section{Related work}
\label{sec:related_work}

In this section we cover work related to our proposed MDL subgroup lists, in three categories: \emph{subgroup discovery}; \emph{rule learning}; and \emph{MDL for pattern mining}. The relevance of each topic is as follows: subgroup discovery directly relates to the task at hand; rule learning are generalisations of subgroup discovery; and MDL for pattern mining shares the same theory for formalising the problem.

\subsection{Subgroup discovery}\label{sec:work_sd}

In its traditional form, subgroup discovery is also known as top-$k$ subgroup mining \citep{atzmueller2015subgroup}, entails the mining of the $k$ top-ranking subgroups according to a quality measure and a number $k$ selected by the user. This formulation suffers from three main issues that make it impractical for most applications: $1)$ lack of \emph{efficient search algorithms} for more relevant quality measures \citep{van2012diverse,bosc2018anytime}; $2)$ \emph{redundancy of subgroup sets} mined, i.e., the fact that subsets with the highest deviation according to a certain quality measure tend to cover the same region of the dataset with slight variations in their description of the subset \citep{van2012diverse}; $3)$ lack of \emph{statistical guarantees} and generalisation of mined subgroups \citep{van2016expect}. We will now go over the contributions of previous works on all these issues, with a specific focus on the last two---redundancy and statistical guarantees---which our work proposes to solve. 

\subsubsection{Efficient search algorithms} 

Algorithms for subgroup discovery can be broadly divided into three categories: 1) exhaustive search; 2) sampling-based methods; and 3) heuristics. In our work, we use a heuristic approach based on beam search to generate the candidate subgroups to add at each iteration. We will now present the developments in each of these three topics and why they are not feasible for finding good subgroup lists.

\paragraph{Exhaustive search} methods have the advantage of guaranteeing to find the best solution. Most of these approaches in SD are either based on branch-and-bound \citep{webb1995opus} or on extending frequent pattern mining algorithms \citep{aggarwal2014frequent} to SD, such as Apriori-SD \citep{kavvsek2003apriori,kavvsek2006apriori} based on Apriori, and SD-Map \citep{atzmueller2006sd} and SD-Map*\citep{atzmueller2009fast} based on FP-growth. However, for the implementation to be efficient in terms of time complexity, e.g., SD-Map and SD-Map*, they combine pruning of the search space with efficient traversal and data structures. And even though these approaches can handle multi-target problems, their efficiency is constrained to specific quality measures that allow for efficient search and/or pruning of the search space. To address some of the limitations regarding quality measures, \cite{boley2017identifying} proposed an efficient exhaustive search (for numeric targets) to quality measures that consider the dispersion of the target values. Also, note that the previous methods could only find the optimal given already discretised explanatory variables, thus \cite{belfodil2018anytime} proposing to mine subgroups over numeric explanatory variables with guarantees. 
The main limitation of these approaches is that they need to be tailored for specific quality measures, mostly focus on binary targets, require special handling of numeric explanatory variables, do not take the redundancy of the found subgroups into account, and are less efficient for a task where they need to be run multiple times---such as in SSD.

\paragraph{Sampling} approaches can be seen as an attractive time-efficient alternative to exhaustive search, especially when interacting with user's preferences is required \cite{boley2011direct,moens2014instant}. Nonetheless, they require a probability distribution over the pattern space, which needs to be tailored to specific quality measures and attribute types. 

\paragraph{Heuristic} approaches are used when an exhaustive search is not feasible, such as in the case of non-trivial targets, e.g., Bayesian networks in exceptional model mining \citep{duivesteijn2010subgroup}, when the optimal subgroup definition changes throughout the problem such as in SSD, or when faster solutions are deemed necessary. The most common heuristic is the \emph{beam search} \citep{lavravc2004subgroup,meeng2011flexible,van2012diverse,meeng2020forreal}. It allows for a non-exhaustive but efficient procedure that can easily generalise any quality measure or type of target variables. This makes for an obvious choice for our problem as our quality measure---equivalent to WKL plus some additional terms--cannot be easily pruned. Thus, beam-search has the efficiency and flexibility that we need. Other heuristics include Genetic Algorithms (GAs) \citep{carmona2010nmeef,carmona2014overview}, which, to be efficient, need to appropriately select a suitable formulation of the problem and tweak the hyperparameters for each dataset.


\subsubsection{Redundancy of subgroup sets and subgroup set discovery}\label{sec:work_ssd}
To address redundancy among the found subgroups, most previously proposed approaches encompass supervised pattern set mining \citep{bringmann2007chosen}, and methods based on relevance \citep{grosskreutz2012enhanced}, and diversity \citep{van2011non,van2012diverse}. Unlike diversity-based methods, the supervised pattern set mining objective is to find a fixed number of patterns, which must be chosen in advance. At the same time, relevance is limited to non-numeric targets. It is the last group, the diversity-based methods, that share the most similarities to our work, i.e., the area of \emph{Subgroup Set Discovery}. \\

As introduced in Section~\ref{sec:back_ssd}, \emph{Subgroup Set Discovery} is an instantiation of the LeGo framework, that passes from local descriptions of the data (SD) to a global model (SSD) \citep{knobbe2008local}. The main approaches in SSD are CN2-SD~\citep{lavravc2004subgroup}, Diverse Subgroup Set Discovery (DSSD) \citep{van2012diverse}, \emph{Skylines} of subgroup sets \citep{van2013discovering}, Monte Carlo Tree Search for Data Mining (MCTS4DM)~\citep{bosc2018anytime}, Subjectively Interesting Subgroup Mining (SISD)~\citep{lijffijt2018subjectively}, and FSSD~\citep{belfodil2019fssd}.  
Table~\ref{table:literature_SSD} summarises the differences between Subgroup Set Discovery methods, with SSD++ representing our approach and where all methods are compared in terms of: if they use a list or a set; the target variables they support; if they have statistical guarantees; if they have automatic stopping criteria (not defined by the user); and if they have a global definition of a subgroup set or list. Note that top-$k$ mining algorithms could be directly applied to SSD if one iteratively adds subgroups and re-weights the instances \citep{lavravc2004subgroup}. However, these methods would also miss a \emph{global} definition of the problem and would need to be fine-tuned for the dataset at hand.

Considering the methods in more detail, CN2-SD~\citep{lavravc2004subgroup} is a direct adaptation of CN2---a classical rule learner for classification---was one of the first methods to deal with redundancy and can be applied to nominal target variables. Algorithmically, it uses a sequential approach, wherein each iteration adds one subgroup to the set and then removes the data covered by that subgroup until no more data can be covered in this way. More specifically, the method can also be used for unordered search---where only the data of the class of interest covered by each subgroup is iteratively removed---or use a weighted covering scheme that, instead of iteratively removing the covered instances, weighs them based on how many times they were covered before. 
DSSD~\citep{van2012diverse} developed a technique based on a novel measure of overlap between subgroups to iteratively find a set of subgroups. It can be applied to single-and-multi-target nominal and numeric variables with different types of quality measures. Skylines of subgroup sets \citep{van2013discovering} are proposed to directly account for quality-diversity trade-off and find the Pareto optimal subgroup sets of size $k$. MCTS4DM~\citep{bosc2018anytime} uses Monte Carlo tree search to improve the quality of the subgroups found. However, it can only be applied to binary target variables and explanatory variables of the same type (all numeric or all nominal). Subjectively Interesting Subgroup Discovery~\citep{lijffijt2018subjectively} finds the subjectively most interesting subgroup for numeric target variables with regard to the user's prior knowledge, based on an information-theoretic framework for formalising subjective interestingness. By successively updating the prior knowledge based on the found subgroups, it iteratively mines a diverse set of subgroups that are also dispersion-aware. FSSD~\citep{belfodil2019fssd} is a more recent approach that considers the `union' of all subgroups as a single pattern by forming a disjunction of subgroups and evaluating its quality and can only be applied to binary target variables. This approach is similar to a sequential approach for mining subgroups. However, the individual contributions of each subgroup are dissolved in the `new' subgroup formed by the disjunction of all subgroups.

\begin{table}[!h]\centering																		
	\caption{Comparison of Subgroup Set Discovery methods in terms of their key properties. From left to right: model class (list or set); types of supported target variables: binary, nominal, numeric and multi-target; \emph{statistical} guarantees of the subgroups mined; automatic \emph{stopping} criterion (not defined by the user); \emph{global} formulation of a subgroup set/list. }\label{table:literature_SSD}																		
	\ra{1.0}
	\begin{tabular}{@{}lllllllll@{}}\toprule																		
		&		&	\multicolumn{4}{c@{}}{Target variables}							&		&		&		\\ 	\cmidrule(l){3-6}
		Method	&	Model	&	binary	&	nom.	&	num.	&	multi	&	Statistical	&	Stopping	&	Global	\\ \midrule	
		SSD++	&	list	&	\cmark	&	\cmark	&	\cmark	&	\cmark	&	\cmark	&	\cmark	&	\cmark	\\	
		\citetalias{lavravc2004subgroup}	&	list	&	\cmark	&	\cmark	&	-	&	-	&	-	&	-	&	-	\\	
		\citetalias{van2012diverse}	&	set	&	\cmark	&	\cmark	&	\cmark	&	\cmark	&	-	&	-	&	-	\\	
		\citetalias{van2013discovering}	&	set	&	\cmark	&	\cmark	&	-	&	-	&	-	&	-	&	\cmark	\\	
		\citetalias{bosc2018anytime}	&	set	&	\cmark	&	-	&	-	&	-	&	-	&	-	&	-	\\	
		\citetalias{lijffijt2018subjectively}	&	set	&	-	&	-	&	\cmark	&	\cmark	&	\cmark	&	-	&	-	\\	
		\citetalias{belfodil2019fssd}	&	list	&	\cmark	&	-	&	-	&	-	&	-	&	\cmark	&	\cmark	\\	
		\bottomrule																		
	\end{tabular}																	
\end{table}

\subsubsection{Subgroup discovery with statistical guarantees} 
In terms of statistical guarantees to subgroup discovery, most approaches consider first mining the top-$k$ subgroups and then post-processing them in terms of a test to find statistically significant subgroups \citep{duivesteijn2011exploiting,van2016expect}. 

\cite{duivesteijn2011exploiting} proposed to use random permutations of the target variable with respect to a quality measure to evaluate how the discovered subgroups compare against the null hypothesis generated by those permutations. Later, \cite{van2016expect} discussed the concept of significance for subgroup discovery and concluded that p-values should be used with caution as not all false discoveries can be removed in this way, as there will always be random subsets with large effect sizes.

Two approaches that automatically find statistically robust subgroups are \cite{lijffijt2018subjectively} and \cite{song2016subgroup,song2017model}. The first approach (already mentioned in the last section), uses the maximum entropy principle to iteratively find subjectively interesting subgroups against a user's prior knowledge. The second proposes a quality measure that directly considers the subgroup distribution and if this is statistically different from the background/dataset distribution.\\

Our approach strongly deviates from the first two, as our method tests for statistical guarantees during the mining process, and it is parametric---as we use categorical and normal distributions to model the targets. Also, our notion of statistical robustness takes into account the concept of the subgroup list model class. Regarding \cite{lijffijt2018subjectively}, even though they also mine subgroups iteratively, they lack a definition of an optimal subgroup set. Their goal is to model the user's subjective knowledge and find regions in the data the user has no prior knowledge. Finally, our quality measure is similar to that of \cite{song2016subgroup,song2017model} in their top-$k$ nominal target case; however, we also take into account multiple-hypothesis testing and focus on subgroup lists.

\subsection{Rule learning}
Pattern mining and association rule mining \citep{agrawal1993mining} are concerned with mining items that co-occur together, i.e., itemsets or patterns, and relationships between itemsets and a target item, e.g., a class, respectively. A known problem of their direct approach is the infamous \emph{pattern explosion}, i.e., they tend to return enormous amounts of patterns/rules. To solve this problem, many approaches were proposed, but two stand out concerning our work, namely, rule-based classifier and statistical rule mining.

To see the relationship with rule mining, first, note that subgroup discovery can be seen as a specialisation of association rule mining. Second, subgroup lists could be regarded as rule lists with a \emph{fixed} default rule, i.e., the last rule that gets activated when no other rule applies is fixed to `predict' the global distribution of the complete dataset. Rule lists and rule sets have long been a common and successful way to compactly apply rules for classification \citep{rivest1987learning}. \\

\subsubsection{Rule-based classifiers} 

Earlier approaches to finding good rule-based models can be broadly divided into two categories based on their model construction: greedy top-down or bottom-up approaches. From a top-down perspective, methods such as CBA \citep{ma1998integrating} and CMAR \citep{li2001cmar} start by mining all association rules from the data and then adding them to the model one by one. From a bottom-up perspective, methods such as \citep{cohen1995fast} mine one rule at a time until a final model is obtained. \cite{cheng2008direct} effectively removes the two-step approach by using a branch-in-bound search on the FP-growth process and iteratively reduces the search space until a set of discriminant sets for classification is found. Nonetheless, the main limitation of these approaches is that they are based on a heuristic definition of a rule-based model, i.e., they add rules without a \emph{global} optimal criteria.\\

Over the past years, rule learning methods that go beyond greedy approaches have been developed, i.e., Monte-Carlo search for Bayesian rule lists \citep{letham2015interpretable,yang2017scalable}, and branch-and-bound with tight bounds for decision lists \citep{angelino2017learning} and rule sets \citep{boley2021better}. However, the main limitation of these methods is that they can only be applied to small or mid-size datasets and are mostly limited to binary targets.\\

Even though all algorithms mentioned in this section resemble our approach, their main goal is to make the best predictions---not to find the largest deviations in the data. Even though the two problems are related, we emphasise the theoretical difference between subgroup discovery and prediction in Appendix~\ref{appendix:proof_sd_vs_prediction}, where the former focuses on \emph{local} deviations and the latter on a \emph{globally} homogeneous partition of the data.

\subsubsection{Statistical rule mining} 

The idea of mining rules with statistical guarantees is appealing as it increases the users' trust in the patterns found while at the same time reducing the number of rules returned by a miner \citep{hamalainen2019tutorial}. The concept of statistical rule mining progressed by incrementally adding more statistical guarantees. \cite{webb07sigpatts} proposed for the first time the mining of statistically significant patterns, then \cite{hamalainen2012kingfisher} proposed KingFisher, an efficient algorithm to mine dependent rules, i.e., rules that show a dependency with respect to a target in terms of a dependency test like Fisher's exact test. After that, \cite{hamalainen2017specious} added extra procedures to remove spurious relations from the miner findings. Lastly, the criteria under which causal rules can be mined were defined and an efficient algorithm to mine them was proposed \citep{budhathoki2020discovering}. All these methods focus on mining all the possible \emph{individual} statistically significant (or causal) rules and not on finding a non-redundant set, as is the case of Subgroup Set Discovery. In this paper, we aim to accomplish both at the same time, finding the best \emph{global} subgroup list while assuring \emph{local} statistically robust subgroups.


\subsection{MDL in pattern mining}

In data mining, Krimp~\citep{vreeken2011krimp} was the first method to apply the MDL principle holistically, i.e., for the whole model selection process. This seminal work used a version of crude MDL, i.e., a not completely optimal `two-part' encoding of the data, to find the pattern list that compressed a transaction dataset best to address the \emph{pattern explosion} issue in pattern mining. Recent works have aimed at improving the encoding through refined MDL for encoding the data, i.e., an encoding that enjoys optimal properties at least in expectation \citep{grunwald2007minimum}. The first of such approaches was DiffNorm \citep{budhathoki2015difference}, which used a prequential plug-in code to improve the encoding of transaction data, and recently MINT was proposed to mine real-valued pattern sets with a similar encoding \citep{makhalova2020mint}. Although Krimp, DiffNorm, and MINT are used to describe data, they aim to find regularities---not deviations---and do not consider a target variable. For an in-depth survey of MDL in pattern mining, please refer to the survey by \cite{galbrun2020minimum}.

\paragraph{MDL for rule learning} MDL has been used to find optimal sets of association rules for two-view data \citep{van2015association} and tabular data \citep{fischer2019sets}. The latter is the  most related to our work, as it aims to find rule sets that describe the data well. Like Krimp, it aims to find all associations in the data, though not at identifying deviations as we do, and no specific target variable(s) are defined.

As discussed in the contributions of our work in Section~\ref{sec:intro}, this work builds on top of MDL-based rule lists for classification \citep{proencca2020interpretable}. Compared to our work, \cite{proencca2020interpretable} focuses solely on classification, can only handle discretised explanatory variables while using a less optimal model and data encoding.

\section{Conclusions}
\label{sec:conclusion}

We showed that finding good subgroup lists (ordered sets) that are both non-redundant and statistically robust, i.e., \emph{robust subgroup discovery}, is computationally feasible. To achieve this, we formally define the subgroup list model class and the problem of \emph{robust subgroup discovery}. Then, based on these two, we propose an optimal formulation of subgroup lists based on the MDL principle---that includes top-$1$ subgroup discovery in its definition. 
As optimally solving the problem is not possible, we proposed a heuristic algorithm dubbed SSD++ that approximates this objective using a greedy search that adds the subgroup that locally minimises the MDL criterion to the list in each consecutive iteration. Moreover, this approximation was shown to be equivalent to Bayesian testing between subgroup and dataset marginal target distributions plus a penalty for multiple hypothesis testing, which guarantees that each subgroup added to the list is statistically sound. \\

These assertions are supported by empirical evidence obtained on a varied set of $54$ datasets. In the case of nominal targets, our method performed on par in terms of subgroup list quality while obtaining smaller lists with fewer conditions. In the case of numeric targets and through the use of a deviation-aware measure, our method dominated in $92\%$ of the cases. 
Finally, we evaluated statistical robustness by testing the generalisation on unseen data. Our MDL-based formulations obtained the lowest generalisation error $86\%$ of the time when compared to KL and WKL quality measures. \\

Through a case study relating the socioeconomic background and national exam grades of Colombia engineering university students, we showed that SSD++ could be flexibly adapted to different goals of the user. In particular, it can change from a \emph{fine-grained} perspective of the data that finds many subgroups covering small parts of the data well, to a \emph{coarse} perspective that finds few subgroups covering large parts of the data. Also, it was shown that our method is robust to mild violations of our model assumptions.\\

In short, our approach can find interpretable, non-redundant, and statistically robust ordered lists of subsets' descriptions that largely deviate from `normality' for (selected) target variables---deviation from `normality' is measured as a divergence between the subset and the dataset/background distributions for those variables---based on the user-specified interests on coverage---descriptions that cover a small or large portion of the data.

\paragraph{Limitations.} Even though the SSD++ algorithm has some appealing local statistical properties, we do not know how far the found models are from the optimal subgroup lists as defined by the global MDL criteria we proposed. Also, it does not scale very well for numeric targets, which was to be expected from the time complexity analysis. At the moment, multiple target variables are assumed to be independent, which can produce erroneous results when this assumption is violated. Preliminary experiments show that for moderately correlated variables (e.g., with a correlation of $0.5$) this does not seem to be an issue, but there is no quantification of its implications. Similarly, for numeric targets, we use a normal distribution, and several datasets violate this assumption, either by behaving like a multi-modal or truncated distribution.  

\paragraph{Future work.} The main lines of research for future work can be divided into three categories: $1)$ extending subgroup lists to other target variables and/or distributions; $2)$ algorithmic developments; and $3)$ generalise this framework to other model classes. In the first category, an obvious extension would be to distributions that take into account multiple dependent target variables, such as multivariate-normal distributions for numeric targets and over itemsets for the nominal case. Another interesting and straightforward development would be the extension of our work to mixed targets, combining both nominal and numeric variables. In the second category, algorithmic developments could go from mere upper-and-lower bounds to improvements in search methods and to study the feasibility of global search such as Markov Chain Monte Carlo methods used by \cite{yang2017scalable} or branch-and-bound algorithms used by \cite{boley2021better}. In the third category, our approach could be formalised for subgroup sets, allowing for overlap between the subgroups.

\begin{acknowledgements}
	This work is part of the research programme Indo-Dutch Joint Research Programme for ICT 2014 with project number 629.002.201, SAPPAO, which is (partly) financed by the Netherlands Organisation for Scientific Research (NWO).
\end{acknowledgements}

%
%
\clearpage
\begin{appendix}
\noindent{\huge\bfseries Appendices\par}

\section{Normalised maximum likelihood independence for non-overlapping multinomials}\label{appendix:NML_derivation}	
	
For this section, let us assume that we have a dataset $D =\{\Xmat,Y\}$ and model $M$ that forms a partition over the whole data. The model $M$ divides the data $D$ in $\omega$ parts, of the form $\{(\Xmat^1,Y^1), \cdots,(\Xmat^\omega,Y^\omega)  \}$. Each part has an associated categorical distribution with estimated parameters $\T^i$ over the target part $Y^i$ (as defined in Section~\ref{sec:sd}).

Our goal in this section is to show that the NML encoding of a partition equals to the sum of the NML encoding of its parts:
\begin{equation}\label{eq:NML_separation}
L_{\mathrm{NML}}(Y \given \Xmat, M) = \sum_{i=1}^{\omega} L_{\mathrm{NML}}(\Yi).
\end{equation}	
Note that in the case of a subgroup list, as the default rule does not require NML encoding, the $M$ used in this section represents the subgroups $S$, and $D$ means the data covered by these. In the case of a tree or rule list, $M$ represents the model that partitions the data at the level of leaves and rules (including default rule), respectively, and $D$ the whole dataset. This is done without any loss of generality as the separation property allows us to separate the encoding of the default rule for a subgroup list.
	
First, let us recall the definition of the NML probability distribution \citep{shtar1987universal}:
\begin{equation*}
L_{\mathrm{NML}}(Y \given \Xmat, M) =  - \log\left(
\frac{\Pr (Y \given \Xmat;\hat{M}(Y \given \Xmat))  }{\sum_{Z \in \mathcal{Y}^{n}}   \Pr (Z  \given \Xmat;\hat{M}(Z \given \Xmat)) } \right), 
\end{equation*}
where $\mathcal{Y}^{n}$ is the set of all possible sequences of $n$ points with $k=|\mathcal{Y}|$ categories, $\hat{M}(Y \given \Xmat)$ and $\hat{M}(Z \given \Xmat)$ are the models with parameters estimated according to the maximum likelihood over the data $Y$ and $Z$, respectively.
Taking into account that our data is independent and identically distributed (\emph{i.i.d.}), and that our model $M$ partitions the data into $\omega$ parts, we can further develop the previous formula to:

\begin{equation}\label{eq:NML_development}
\begin{split}
L_{\mathrm{NML}}(Y \given \Xmat, M) &\myeq -\log \left(	\frac{ \prod_{i=1}^{n} \Pr (y^i \given \x^i;\hat{M}(Y \given \Xmat))   }{\sum_{Z \in \mathcal{Y}^{n}}  \prod_{i=1}^{n}  \Pr (z^i  \given \x^i;\hat{M}(Z \given X)) } \right) \\
&= -\log \left(
\frac{ \prod_{i'=1}^{\omega} \Pr (Y^{i'};\T(Y^{i'}))   }{\sum_{Z \in \mathcal{Y}^{n}} \prod_{i'=1}^{\omega} \Pr (Z^{i'} ;\T(Z^{i'})) } \right) \\
&= -\log \left(
\frac{ \prod_{i'=1}^{\omega} l(\T^{i'} \given Y^{i'}) }{g(Y,X,M) } \right) \\
&= -\log \left(
\sum_{i'=1}^{\omega} l(\T^{i'} \given Y^{i'}) \right) + \log g(Y,X,M)  	,
\end{split}
\end{equation}
where $l(\T^{i'} \given Y^{i'})$ is the likelihood function for each of the $\omega$ parts and $g(Y,X,M)$ is a complexity function that depends on these $3$ variables.
	
The first term is already independent for each part; however, the second is not. 

Let us now look at $g(Y,X,M)$ in the case where we only have one part in the dataset, i.e., $D^1$. We will call this term the NML complexity of a multinomial distribution and denote it by $\COMP(n_1, k)$ of one part $D^1 = \{Y^1, X^1\}$, with $n_1 = |D^1|$ and $k = \Y$
\begin{equation}
\begin{split}
\COMP(n_{1}, k)&=\log \left(\sum_{Z \in \mathcal{Y}^{n_{1}}}  \Pr (Z^{1} ;\T(Z^{1})) \right) \\
&= \log \left(\sum_{Z \in \mathcal{Y}^{n_{1}}}  \prod_{i=1}^{n_1}  \Pr (z^i ;\T(Z^{1})) \right) \\
&=\log \left( \sum_{n_{11}+n_{12}+...+n_{1k} = n_1} \frac{n_1!}{n_{11}!n_{12}!...n_{1k}!} \prod_{c \in \Y}  \left( \frac{n_{1c}}{n_1} \right)^{n_{1c}} \right)\\ 
\end{split}
\end{equation}
where $n_{1c}$ is the number of points of category $c$ in $Y^1$, and the passage from the second equality to the last is a property of multinomial distributions commonly used to make the computation of $\COMP(n_a, k)$ simpler \citep{grunwald2007minimum}. It is interesting to note that  $\COMP(n_a, k)$ only depends on the number of points in $Y^1$ and its cardinality, not on the actual values. This term, i.e., the complexity of a multinomial distribution over $n_1$ points with $k$ possible values, measures the likelihood of each possible sequence. 

\begin{table}[!h]\centering													\caption{All possible sequences of a partition of fixed length of the data in three parts. Fixed length means that all possible parts always have the same amount of points, as e.g. $|A_1| = |A_2| = \cdots = |A_a|= n_{A}$.} \label{table:paritionsdataset}				 \begin{tabular}{@{}ccc@{}}\toprule														
		Part $1$	&	Part $2$	& Part $3$	\\  \midrule		
		$A_1$	&	$B_1$	&  $C_1$	\\ 		
		$A_1$	&	$B_1$	&  $C_2$   	\\ 	
		\vdots	&	\vdots	&  \vdots  	\\ 	
		$A_1$	&	$B_2$	&  $C_1$   	\\
		\vdots 	&	\vdots	&  \vdots  	\\ 	
		$A_a$	&	$B_b$	&  $C_c$   	\\ 		
		\bottomrule
	\end{tabular}																	
\end{table}	
	
Now we must generalise from a part to the whole partition of the dataset. To illustrate how to do this, let us first look at Table~\ref{table:paritionsdataset}, which shows an example of all the possible sequences in a fixed-length three-part partition of the data. Then, taking into account those three parts, let us look at how the probabilities of all those sequences could be computed:
\begin{equation*}\label{eq:multiplicationdist}	
\begin{split}
\sum_{\forall a,b,c} \Pr(A_a)  \Pr(B_b)    \Pr(C_c) &= \left( \sum_{\forall a} \Pr(A_a) \right) \cdot\left( \sum_{\forall b,c} \Pr(B_b)    \Pr(C_c) \right) \\
&= 	\left( \sum_{\forall a} \Pr(A_a) \right) \cdot\left( \sum_{\forall b} \Pr(B_b)  \right)\cdot \left( \sum_{\forall c} \Pr(C_c)  \right),
\end{split}
\end{equation*}
where this follows naturally from the distributive property of the multiplication. It is easy to see that this generalises to partitions of any number of parts. Thus, going back to the complexity term $g(Y,X,M)$, we can see that

\begin{equation}
\begin{split}
\log g(Y,X,M) &= \log \sum_{Z \in \mathcal{Y}^{n}} \prod_{i'=1}^{\omega} \Pr (Z^{i'} ;\T(Z^{i'}))  \\
&= \log  \prod_{i'=1}^{\omega} \sum_{Z^{i'} \in \mathcal{Y}^{n_{i'}}} \Pr (Z^{i'} ;\T(Z^{i'}))\\
&=   \sum_{i'=1}^{\omega} \log\sum_{Z^{i'} \in \mathcal{Y}^{n_{i'}}} \Pr (Z^{i'} ;\T(Z^{i'}))\\
&=    \sum_{i'=1}^{\omega} \log \COMP(n_{i'}, k)\\
\end{split}
\end{equation}

Substituting this back into Eq.~\eqref{eq:NML_development}, we obtain what we wanted:
\begin{equation}
\begin{split}
L_{\mathrm{NML}}(Y \given \Xmat, M) &= -\log \left(
\sum_{i=1}^{\omega} l(\T^{i} \given Y^{i}) \right) + \sum_{i=1}^{\omega}\log \COMP(n_i, k) \\
&= \sum_{i=1}^{\omega}  l(\T^{i} \given Y^{i})  +  \COMP(n_i, k)\\
&= \sum_{i=1}^{\omega}  L_{\mathrm{NML}}(Y^i)\\
\end{split}
\end{equation}

\section{Bayesian encoding of a normal distribution with mean and standard deviation unknown}\label{appendix:Bayesian_derivation}
For encoding a sequence of numeric valued i.i.d. observations such as $Y = \{y_1,....,y_n\}$, the Bayesian encoding takes the following form:
\begin{equation}
	P_{\mathrm{Bayes}} (Y) = \int_{\Theta} f (Y \given \Theta) w(\Theta) \dif \Theta,
\end{equation}
where $f$ is the probability density function (pdf), $\Theta$ is the set of parameters of the distribution, and $w(\Theta)$ the prior over the parameters. 
In the case of a normal distribution $\Theta = \{\mu, \sigma \}$, with $\mu$ and $\sigma$ being its mean and standard deviation, respectively, the pdf $f(Y \given \Theta)$ over a sequence $Y$ is the multiplication of the individual pdfs, thus:
\begin{equation}
	f(Y \given \mu, \sigma) = \frac{1}{(2 \pi)^{n/2} \sigma^n } \exp \left[ -\frac{1}{2 \sigma^2} \sum_i^n (y^i-\mu)^2   \right],
\end{equation}

In order not to bias the encoding for specific values of the parameters, we choose to use the constant Jeffrey's prior of $1/\sigma^2$ for the unknown parameters $\mu$ and $\sigma$, and add an extra. Thus, our prior is given by:
\begin{equation}
	w(\mu,\sigma) = \frac{1}{\sqrt{2 \pi} \sigma^2 },
\end{equation}
where $ 1/\sqrt{2 \pi} $ was added for normalisation reasons.

Putting everything together, one obtains:
\begin{equation}\label{eq:BayesStep2}
	\begin{split}
		&P_{\mathrm{\mathrm{Bayes}}} (Y)= \\
		&= (2 \pi)^{-\frac{n+1}{2}} \int_{-\infty}^{+\infty}\int_{0}^{+\infty} \frac{1}{\sigma^{n+2} } \exp \left[ -\frac{1}{2 \sigma^2} \left(  \sum_i^n (y^i-\mu)^2  \right) \right] \dif \sigma \dif \mu.
	\end{split}
\end{equation}

The integrals over the whole space of the parameters $\mu$ and $\sigma$ allow us to penalise the fact that we do not know the statistics \emph{a priori}, thus penalising the fact that distribution over $n$ points could, by chance, have the same statistics like the one found in the data. 

Note that using an improper prior requires that we somehow make it proper, i.e., we need to find a way to make the integration over the prior finite $ \int \int w(\mu, \sigma) = K,$ where $K$ is a constant value. The usual way to make an improper prior finite is to condition on the $k$ minimum number observations $Y^{|k} \in Y$ needed to make the integral proper \citep{grunwald2007minimum}, which in the case of two unknowns ($\mu$ and $\sigma$) is $k =2$. Thus, instead of using $w(\mu,\sigma)$ we will in practice be using $w(\mu,\sigma \given \Ytwon)$, and using the chain rule and the Bayesian formula returns a total encoding of $Y$ equal to
\begin{equation}\label{eq:conditionalbayes}
	P(Y) = P_{\mathrm{Bayes}} (Y \given \Ytwon)P (\Ytwon) = \frac{P_{\mathrm{Bayes}} (Y)}{P_{\mathrm{Bayes}} (\Ytwon)} P (\Ytwon)
\end{equation}
where $P (\Ytwon)$ is a non-optimal probability used to define $\Ytwon = \{ y^1,y^2 \}$ that we will define later and $y^1,y^2$ chosen in a way that maximises $P(Y)$. 
Now that we have all the ingredients to define $P(Y)$ we will start by defining $P_{\mathrm{Bayes}} (Y)$ and then choose the appropriate probability for $P(\Ytwon)$.

To solve the first integral of $P_{\mathrm{Bayes}} (Y)$ in Eq.~\eqref{eq:BayesStep2}, we integrate in $\sigma$ and note that the formula is an instance of the gamma function,
\begin{equation}
	\Gamma (k) = \int_{0}^{+\infty} z^{k-1} e^{-z} \dif z,
\end{equation}
with the corresponding variable transformation:
\begin{equation}
	z= \frac{A}{2 \sigma^2} ;\; \frac{1}{\sigma} = \frac{2^{1/2}z^{1/2}}{A^{1/2}} ;\;  \dif \sigma = - \frac{\sigma}{2z} \dif z ;\;  A = \left[ \sum_i^n (y^i-\mu)^2 \right],
\end{equation}
Performing the variable transformation and noting that the minus sign of $\dif z$ cancels with the reversing of the integral limits, we get:
\begin{equation}\label{eq:BayesStep3}
	\begin{split}
		&P_{\mathrm{Bayes}} (Y)= \\
		&= \Gamma \left( \frac{n+1}{2}\right) 2^{\frac{n+1}{2}-1} (2 \pi)^{-\frac{n+1}{2}} \int_{-\infty}^{+\infty} \left[ \sum_i^{n} (y^i-\mu)^2  \right]^{-\frac{n+1}{2}} \dif \mu .
	\end{split}
\end{equation}

To solve the integral in $\mu$ we need to introduce the statistics $\hat{\mu}$ and $\hat{\sigma}$ as the values estimated from the data. We define these quantities as:
\begin{equation}
	\hat{\mu}= \frac{1}{n} \sum_i^n y^i ; \;  \hat{\sigma}^2 = \frac{1}{n} \sum_i^n (y^i-\hat{\mu})^2\;,
\end{equation}
where $\hat{\mu}$ is the mean estimator over $n$ data points and $\hat{\sigma}^2$ is the estimator of the variance. Note that for the variance the biased version with $n$ was used instead of with $n-1$ as it allows to compute the Residual Sum of Squares (RSS) directly by $RSS = n \hat{\sigma}$.

Focusing now on the interior part of the integral of Eq.~\ref{eq:BayesStep3} and rewriting it in order to resemble the t-student distribution, we obtain: 
\begin{equation}
	\begin{split}
		&\left[ \sum_i^{n} (y^i-\mu)^2 \right]^{-(n+1)/2} = \\
		& \left[\sum_i^{n} (y^i)^2- n\hat{\mu}^2 + n\hat{\mu}^2 -2 n\hat{\mu}\mu + n\mu^2 \right]^{-(n+1)/2}=\\
		& \left[\sum_i^{n} (y^i)^2- n\hat{\mu}^2 +n(\hat{\mu}-\mu)^2 \right]^{-(n+1)/2}=\\
		&\left[ n \hat{\sigma}^2 +n(\hat{\mu}-\mu)^2 \right]^{-(n+1)/2} =\\
		&\left[n\hat{\sigma}^2\right]^{-(n+1)/2}  \left[1  + \frac{(\hat{\mu}-\mu)^2}{\hat{\sigma}^2}  \right]^{-(n+1)/2} \\
		&\left[n\hat{\sigma}^2\right]^{-(n+1)/2}  \left[1  + \frac{1}{n} \left( \frac{\hat{\mu}-\mu}{s_s^2} \right)^2  \right]^{-(n+1)/2}, \\
	\end{split}
\end{equation}
where $s_s^2 = \hat{\sigma}^2/n$ is the ``sampling" variance. Now, taking into account the fact that the integral of the t-student distribution over the whole space is equal to one, and reshuffling around its terms we get
\begin{equation}
	\int_{-\infty}^{+\infty}  \left[1 + \frac{1}{n} \left( \frac{\hat{\mu}-\mu}{s_s}  \right)^2   \right]^{-\frac{n+1}{2}} \dif \mu = \frac{\Gamma\left(\frac{n}{2}\right)\sqrt{\pi n}s_s}{\Gamma\left(\frac{n+1}{2}\right)}.
\end{equation}
Inserting this back in Eq.~\ref{eq:BayesStep2} we obtain:

\begin{equation}
	\begin{split}
		&P_{\mathrm{Bayes}} (Y)= \\
		&= \Gamma \left(\frac{n+1}{2} \right) 2^{\frac{n+1}{2}-1} (2 \pi)^{-\frac{n+1}{2}} \frac{\Gamma(\frac{n}{2})\sqrt{\pi n}s_s}{\Gamma(\frac{n+1}{2})} \left[n\hat{\sigma}^2\right]^{-(n+1)/2}  \\
		&=  2^{-1}\pi^{-\frac{n}{2}} \Gamma\left(\frac{n}{2}\right)\frac{1}{\sqrt{n}}   \left[n\hat{\sigma}^2\right]^{-\frac{n}{2}},  \\
	\end{split}
\end{equation}

Returning to the the conditional probability of Eq.~\eqref{eq:conditionalbayes}, we see that we still need to define $P(\Ytwon)$, the non-optimal probability of the first two-points. As in the case of our model class we assume that the dataset overall statistics are known, i.e., $\Theta = \{\hat{\mu}_d, \hat{\sigma}_d \}$, we will use this distribution to find the probability of the points $\Ytwon = \{y^1,y^2\}$ as :
\begin{equation}
	P(\Ytwon) = \log 2 \pi + \log \hat{\sigma}_d +\left[ \frac{1}{2 \hat{\sigma}_d^2} \sum_i^2 (y^i-\hat{\mu}_d)^2   \right] \log e. \\
\end{equation}

Finally, applying the minus logarithm base $2$ to all the terms in Eq~\eqref{eq:conditionalbayes} to obtain the total code length in bits,
\begin{equation}\label{eq:BayesStepFinal}
	\begin{split}
		&L_{Bayes2.0}(Y) = -\log P_{\mathrm{Bayes}} (Y) +\log P_{\mathrm{Bayes}} (\Ytwon) - \log P(\Ytwon)  \\
		&=1+ \frac{n}{2} \log \pi- \log \Gamma \left( \frac{n}{2} \right) + \frac{1}{2} \log n +\frac{n}{2} \log \left (n\hat{\sigma}_n^2  \right)\\
		&-1- \frac{2}{2}\log \pi+ 0 - \frac{1}{2} -\log \left (\sum_i^2 (y^i-\hat{\mu}_2)^2   \right) \\
		&+ \frac{2}{2}\log \pi+ \log \hat{\sigma}_d +\left[ \frac{1}{2 \hat{\sigma}_d^2} \sum_i^2 (y^i-\hat{\mu}_d)^2   \right] \log e  \\
		& = \frac{n}{2} \log \pi- \log \Gamma \left( \frac{n}{2} \right) + \frac{1}{2} \log n +\frac{n}{2} \log \left (n\hat{\sigma}_n^2  \right) + L_{cost}(\Ytwon),
	\end{split}
\end{equation}

where $\hat{\mu}_2$ is the estimated mean of $y^1,y^2$ and $L_{cost}(\Ytwon)$ is the extra cost incurred of not being able to use a refined encoding for $\Ytwon$. Now that the encoding length is defined, we need to choose the two points. i.e., $y^1,y^2$. Because we want to minimise this length, we notice that there are only two terms that contribute to it in $L_{cost}(\Ytwon)$, and thus by choosing the two observations close to $\hat{\mu}_d$ minimises both the encoding of $P(\Ytwon)$ and maximise $P_{\mathrm{Bayes}} (\Ytwon)$ for most cases. There are exceptions to this, depending on the respective values of $\mu_d$ and $y^1,y^2$, but these are not significant to change the values too much and require less computational search to find the points.

\subsection{Convergence to BIC for large $n$}\label{appendix:BIC}
In this section, it is shown that for a large number of instances $n$, the Bayesian encoding of a normal distribution with unknown mean and standard deviation (Eq.~\eqref{eq:BayesStepFinal}) converges to the encoding of a normal distribution with mean and standard deviation known plus $\log n$, i.e., proportional to the definition of the Bayes Information Criterion (BIC).
	First, the encoding of a normal distribution with mean and standard deviation known over $n$ \emph{i.i.d.} points is equal to the sum of the individual encodings: 
	\begin{equation}
	L(Y \given \hat{\Theta}) = \frac{n}{2} \log 2\pi + \frac{n}{2} \log \hat{\sigma}^2 +  \left[ \frac{1}{2 \hat{\sigma}^2} \sum_i^n (y^i-\hat{\mu})^2   \right] \log e. \\
	\end{equation}
	Second, we need to use the Stirling approximation of the Gamma function for large $n$:
	\begin{equation}
	\begin{split}
	&- \log \Gamma \left( \frac{n}{2} \right)  \sim\\
	&\sim -\frac{1}{2}\log \pi  -\frac{1}{2}\log \left (n-2 \right) - \left (\frac{n}{2}-1 \right) \log \left (\frac{n}{2}-1 \right) + \left (\frac{n}{2}-1 \right) \log e, \\
	\end{split}
	\end{equation}
	and finally we insert it into Eq.~\eqref{eq:BayesStepFinal} and assume $\tau = 1$ to obtain:
	\begin{equation}
	\begin{split}
	&L(Y)  \sim \\
	&\sim 1 +\frac{n-1}{2} \log \pi + \frac{1}{2} \log \left (\frac{n}{n-2} \right) +\frac{n}{2} \log \left ( \frac{n\hat{\sigma}^2}{n/2-1} \right)+ \left (\frac{n}{2}-1\right) \log e  \\ 
	&+\log \left (\frac{n}{2}-1 \right) + L_{cost}(\Ytwon)  \\
	& \sim \frac{n}{2} \log \pi + \frac{n}{2} \log 2\hat{\sigma}^2 + \left[ \frac{1}{2 \hat{\sigma}^2} \sum_i^n (y^i-\mu)^2   \right] \log e  + \log n -\log e + L_{cost}(\Ytwon)\\
	&= L(Y \given \hat{\Theta}) + \log \frac{n}{e}  + L_{cost}(\Ytwon)\\
	& \sim \frac{1}{2} \left( 2L(Y \given \hat{\Theta}) + 2\log n -2\log e \right) \\
	&= \frac{1}{2}BIC,
	\end{split}
	\end{equation}
	where from the second to the third line, we assumed large $n$, making some of the terms disappear, while the definition $n \hat{\sigma}^2 = \sum_i^n (y^i-\mu)^2 $ is used for making the third term of the third expression appear. From the fourth to the fifth expressions, it was assumed that $L_{cost}(\Ytwon)$ is negligible, as it is the cost of not being able to encode the first two points optimally. For the Bayes information criterion, we used its standard definition,
	\begin{equation}
	BIC = -2 \ln \ell(\Theta \given Y) + k\ln n,
	\end{equation}
	where $\ell(\Theta \given Y)$ is the likelihood as estimated from the data, and $k$ is the number of parameters, which in our case is $2$.\\

\section{Derivation of MDL-based optimal subgroup lists equivalence to WKL-based SD}\label{appendix:wkl_derivation}	

In this appendix we derive the formula that relates the MDL-based subgroup lists with WKL-based subgroup discovery for categorical and normal distributions. This arises as the solution of the maximisation problem (equivalent to the standard MDL minimisation) of:
\begin{equation*}
s^* = \argmax_{s \in \M} \left[ L(\Yd  \given \boldsymbol{\Theta}^d ) - L(Y \given \Xmat,M)  - L(M) \right].
\end{equation*}

\paragraph{Categorical distribution derivation:} 
\begin{equation}\label{eq:app_KLproof_nominal} 
\begin{split}
L(Y \given  \Td)&-L(Y \given \Xmat, M)- L(M) = \\ 
& = L(\Ya \given  \Td) + \cancel{L(Y^{\neg a} \given  \Td)} -L_{\mathrm{NML}}(\Ya)-\cancel{L(Y^{\neg a} \given  \Td)} - L(M)   \\
&=\sum_{y \in Y^s} \log \frac{\pay}{\pdy} - \COMP(n_a,k) - L(M) \\
&=n_a \sum_{c \in \Y}  \pac \log \left( \frac{\pac}{\pdc} \right) - \COMP(n_a,k)- L(M) \\
&=n_a KL(\Ta; \Td) -  \COMP(n_a,k)- L(M),
\end{split}
\end{equation}
where $n_a KL(\Ta; \Td)$ is the Weighted Kullback-Leibler divergence from $\Ta$ to $\Td$. 

\paragraph{Normal distribution derivation:} Using the Stirling approximation of the gamma function: $\Gamma(n+1) \sim  \sqrt{2\pi n } \left( \frac{n}{e} \right)^n$; in Appendix~\ref{appendix:BIC}, the derivation is as follows:
\begin{equation}\label{eq:app_KLproof_numeric}
\begin{split}
&L(Y \given  \Td)-L(Y \given X,M)  \\
& =  L(\Ya \given  \Td)-L_{Bayes2.0}(\Ya \given \Xmata) - L(M) \\
& \sim \frac{n_a}{2} \log \frac{\hsigd^2}{\hsiga^2} +\left[ \frac{1}{2 \hsigd^2} \sum_{y^i \in \Ya} (y^i-\hmud)^2   \right] \log e - \frac{n_a}{2} \log e - \log n_a - L(M)\\
& = \frac{n_a}{2} \log \frac{\hsigd^2}{\hsiga^2} +\left[ \frac{ \sum_{y^i \in \Ya} (y^i)^2 - n \hmua^2 + n \hmua^2 -2n \hmua \hmud-\hmud)^2}{2 \hsigd^2}  \right] \log e \\
&\phantom{=,}- \frac{n_a}{2} \log e - \log n_a - L(M) \\
&= n_a \left [ \log\frac{\hsigd}{\hsiga}+ \frac{\hsiga^2+(\mu_a-\mu_d)^2}{2 \hsigd^2}\loge -\frac{\loge}{2} \right ] - \log(n_a) - L(M)  \\
& = n_a KL(\Ta; \Td) - \log n_a - L(M),
\end{split}
\end{equation}

where $n_a KL(\Ta; \Td)$ is the usage-weighted Kullback-Leibler divergence between the normal distributions specified by the respective parameter vectors.


\section{Difference between subgroup discovery and rule-based predictive models}\label{appendix:proof_sd_vs_prediction}
	
	This appendix shows the difference between the objective being maximised for subgroup discovery and for predictive rules. We do this through the comparison of the equivalent maximisation MDL scores for subgroup lists and classification rule lists \citep{proencca2020interpretable} with only one rule/subgroup---without loss of generality for greater sizes or for regression tasks. To differentiate both model classes, $SL$ and $RL$ will be used for subgroup lists and classification rule lists, respectively.
	
	First, lets recall the form of a subgroup list $SL$ as given in Figure~\ref{fig:subgroup_list_nominal}:
	\begin{equation*}\small
	\begin{split}
	\text{subgroup 1}: & \textsc{ if }   a \sqsubseteq  \x   \textsc{ then }  y \sim Cat(\Ta)\\ 
	\text{dataset}: &\textsc{ else }     y \sim Cat(\Td)
	\end{split}
	\end{equation*}
	where, $\Ta$ are the estimated parameters of subgroup $1$ and $\Td$ are the estimated parameters of the marginal distribution of the dataset and are thus constant for each dataset. Second, the model form of a classification rule list $RL$ takes the following form:
	\begin{equation*}\label{eq:modelclass2} \small
	\begin{split}
	\text{rule 1}: & \textsc{ if }   a \sqsubseteq  \x   \textsc{ then }  Cat(\Ta)\\ 
	\text{default}: &\textsc{ else }     y \sim Cat(\hat{\Theta}^{\neg a})
	\end{split}
	\end{equation*}
	where $\hat{\Theta}^{\neg a}$ was used to emphasise that the default rule of a rule list is not fixed, and is equivalent to the  `not rule $1$'.
	This is the key difference between these two types of models, the default rule is fixed to the marginal distribution of the dataset for subgroup lists, and the default rule has the distribution of the negative set of the rules in the list for rule lists.
	It should be noted that there are many definitions of rule lists that use a fixed rule; however, having a variable default rule that maximises the prediction quality is the best representative of rule lists and of the objective of finding the best machine learning model, i.e., returning the best partition of the data with the smallest error possible. Note that a decision tree is also part of this family of models, as any path starting at the tree's root to one of its leaves also forms a rule. Thus, a decision tree is equivalent to a set of disjoint rules, i.e., none of the rules described in this way overlap on a dataset. For the type of classification rule lists defined above, the encoding of the first rule and default rule is given by Eq.~\ref{eq:NML} as for both rules; the parameters are unknown.
	
	Thus the MDL score of a rule list is given by:
	\begin{equation}
	\begin{split}
	L(D,RL) & = L(\Ya \given \Xmata) + L(Y^{\neg a} \given  \Xnega) + L(RL),
	\end{split}
	\end{equation}
	and note that the model encoding $L(RL)=L(SL)$, has both lists can be described in the same manner.
	
	Following the same steps as in Section~\ref{sec:sd_proof} by turning the MDL score objective from a minimisation to maximisation by multiplying by minus one and adding the constant $L(\Yd  \given \boldsymbol{\Theta}^d )$, we obtain the same objective as in Eq.~\ref{eq:maximize}:
	\begin{equation*}
	r^* = \argmax_{s \in \M} \left[ L(\Yd  \given \boldsymbol{\Theta}^d ) - L(Y \given \Xmat,RL)  - L(RL) \right],
	\end{equation*}
	where $r$ is the rule that maximises the objective. Working out this equation, maximisation objective of a \emph{classification rule list} for a target variable of $k$ class labels is given by:
	\begin{equation}\label{eq:MDLregression}
	\begin{split}
	&L(Y \given  \Td)-L(Y \given \Xmat,M) - L(RL)  \\
	& =  L(\Ya \given  \Td)+L(\Ynega \given  \Td)-L(\Ya \given X_a) -L(\Ynega \given \Xnega ) -L(RL) \\
	& = n_a KL(\Ta; \Td) - \COMP(n_{a},k) +  n_{\neg a} KL(\hat{\Theta}^{\neg a}; \Td) - \COMP(n_{\neg a},k)- L(RL),\\
	\end{split}
	\end{equation}
	
	This should be contrasted with the maximisation objective of \emph{subgroup list} of Eq.~\ref{eq:KLproof}, which is given by:
	\begin{equation*}
	\begin{split}
	&L(Y \given  \Td)-L(Y \given \Xmat,M)- L(SL) =\\
	&n_a KL(\Ta; \Td) - \COMP(n_a,k)- L(SL).
	\end{split}
	\end{equation*}
	
	Comparing both of the last equations, we can notice the crucial distinction between subgroup discovery and classification: the \emph{local} nature of subgroup discovery and the \emph{global} nature of the classification task. In other words, subgroup discovery aims at finding subgroups that locally maximise their quality, independently of the rest of the dataset, and even though rules for classification try to maximise their \emph{local} quality also, they have to take into account the quality of their negative set, i.e., a classification rule cannot be considered by its quality alone, it has to be considered in terms of its \emph{global} impact in the dataset. 
	On the other hand, this result also shows the similarity between both tasks and where the confusion sometimes arises, i.e., in some cases, the best subgroup can also be the best rule. An example of this would be a very large dataset (relatively to the number of observations covered by the rule). Here, the best rule/subgroup would cover a small number of observations compared to the rule formed by the negative set of that rule, i.e., $D^{\neg a}$, as a similar distribution to $\Td$, making $\hat{\Theta}^{\neg a} \sim \Td$. Nonetheless, this similarity decreases in the case of larger lists, as the default rule will always represent what is left. In contrast, in a subgroup list, it remains constant and represents what we consider uninteresting. The same result can be obtained for regression rule lists.
	
\clearpage


\section{Datasets for empirical experiments}\label{appendix:datasets}
The datasets selected are commonly used in machine learning and were retrieved from UCI~\citep{Dua2019uci}, Keel~\citep{alcala2011keel}, MULAN~\citep{mulan} repositories. The datasets used for nominal and numeric targets experiments can be seen in Table~\ref{table:data_nominal} and \ref{table:data_numeric}, respectively. 
	
	\begin{table}[!htb]\centering												
		\caption{Nominal targets datasets: single-binary, single-nominal and multi-label. Dataset properties: number of \{target variables $T$; target labels $|\Y|$; samples $|D|$; type of variables (nominal/numeric)\}.}\label{table:data_nominal}												
		\ra{1.0} \begin{tabular}{@{}lrrrrr@{}}\toprule												
			Dataset	&	$T$	&	$|\Y|$	&	$|D|$	&	$V(nom./num.)$				\\ \midrule
			sonar 	&$	1	$&$	2	$&$	208	$&$(	0	/	60	)$	\\
			haberman 	&$	1	$&$	2	$&$	306	$&$(	0	/	3	)$	\\
			breastCancer	&$	1	$&$	2	$&$	683	$&$(	0	/	9	)$	\\
			australian 	&$	1	$&$	2	$&$	690	$&$(	0	/	14	)$	\\
			TicTacToe 	&$	1	$&$	2	$&$	958	$&$(	9	/	0	)$	\\
			german 	&$	1	$&$	2	$&$	1\,000	$&$(	13	/	7	)$	\\
			chess 	&$	1	$&$	2	$&$	3\,196	$&$(	36	/	0	)$	\\
			mushrooms 	&$	1	$&$	2	$&$	8\,124	$&$(	22	/	0	)$	\\
			magic 	&$	1	$&$	2	$&$	19 \, 020	$&$(	0	/	10	)$	\\
			adult 	&$	1	$&$	2	$&$	45\,222	$&$(	8	/	6	)$	\\ \midrule
			iris 	&$	1	$&$	3	$&$	150	$&$(	0	/	4	)$	\\
			balance 	&$	1	$&$	3	$&$	625	$&$(	0	/	4	)$	\\
			CMC 	&$	1	$&$	3	$&$	1\,473	$&$(	0	/	9	)$	\\
			page-blocks 	&$	1	$&$	5	$&$	5\,472	$&$(	0	/	10	)$	\\
			nursery 	&$	1	$&$	5	$&$	12\,960	$&$(	7	/	1	)$	\\
			automobile 	&$	1	$&$	6	$&$	159	$&$(	10	/	15	)$	\\
			glass 	&$	1	$&$	6	$&$	214	$&$(	0	/	10	)$	\\
			dermatology 	&$	1	$&$	6	$&$	358	$&$(	0	/	34	)$	\\
			kr-vs-k 	&$	1	$&$	18	$&$	28\,056	$&$(	6	/	0	)$	\\
			abalone 	&$	1	$&$	28	$&$	4\,174	$&$(	1	/	7	)$	\\ \midrule
			emotions	&$	6	$&$	2	$&$	593	$&$(	0	/	72	)$	\\
			scene	&$	6	$&$	2	$&$	2407	$&$(	0	/	294	)$	\\
			flags	&$	7	$&$	2	$&$	194	$&$(	9	/	10	)$	\\
			yeast	&$	14	$&$	2	$&$	2417	$&$(	0	/	103	)$	\\
			birds	&$	19	$&$	2	$&$	645	$&$(		/	258	)$	\\
			genbase	&$	27	$&$	2	$&$	662	$&$(	1186	/	0	)$	\\
			mediamill	&$	101	$&$	2	$&$	43\,907	$&$(	0	/	120	)$	\\
			CAL500	&$	174	$&$	2	$&$	502	$&$(	0	/	68	)$	\\
			Corel5k	&$	374	$&$	2	$&$	5000	$&$(	499	/	0	)$	\\
			\bottomrule												
		\end{tabular}												
	\end{table}

	\begin{table}[!htb]\centering														
		\caption{Numeric targets datasets: single-numeric and multi-numeric. Dataset properties:  \{number of target variables $T$; minimum and maximum target values $[min.,max.]$; number of samples $|D|$; number of type of variables (nominal/numeric)\}.}\label{table:data_numeric}														
		\ra{1.0} \begin{tabular}{@{}lrrrrr@{}}\toprule														
			Dataset	&	T	&			$[min.;max.]$	&	$|D|$	&	$V(nom./num.)$				\\  \midrule
			baseball 	&$	1	$&$[	109	;	6100	]$&$	337	$&$(	4	/	12	)$	\\ 
			autoMPG8 	&$	1	$&$[	9	;	46.6	]$&$	392	$&$(	0	/	6	)$	\\ 
			dee 	&$	1	$&$[	0.8	;	5.1	]$&$	365	$&$(	0	/	6	)$	\\ 
			ele-1 	&$	1	$&$[	80	;	7675	]$&$	495	$&$(	0	/	2	)$	\\ 
			forestFires 	&$	1	$&$[	0	;	1091	]$&$	517	$&$(	0	/	12	)$	\\ 
			concrete 	&$	1	$&$[	3	;	21	]$&$	1030	$&$(	0	/	8	)$	\\ 
			treasury 	&$	1	$&$[	29	;	90	]$&$	1049	$&$(	0	/	15	)$	\\ 
			wizmir 	&$	1	$&$[	29	;	90	]$&$	1461	$&$(	0	/	9	)$	\\  
			abalone 	&$	1	$&$[	1	;	29	]$&$	4177	$&$(	0	/	8	)$	\\ 
			puma32h 	&$	1	$&$[	-0.0867	;	0.0898	]$&$	8192	$&$(	0	/	32	)$	\\ 
			ailerons 	&$	1	$&$[	-0.0036	;	0	]$&$	13750	$&$(	0	/	40	)$	\\ 
			elevators 	&$	1	$&$[	0.012	;	0.078	]$&$	16599	$&$(	0	/	18	)$	\\ 
			bikesharing	&$	1	$&$[	1	;	977	]$&$	17379	$&$(	2	/	10	)$	\\ 
			california 	&$	1	$&$[	14999	;	500001	]$&$	20640	$&$(	0	/	8	)$	\\ 
			house 	&$	1	$&$[	0	;	500001	]$&$	22784	$&$(	0	/	16	)$	\\  \midrule
			edm	&$	2	$&$[	-1	;	1	]$&$	154	$&$(	0	/	16	)$	\\
			enb	&$	2	$&$[	6.01	;	48.03	]$&$	768	$&$(	0	/	8	)$	\\
			slump	&$	3	$&$[	0	;	78	]$&$	103	$&$(	0	/	7	)$	\\
			sf1	&$	3	$&$[	0	;	4	]$&$	323	$&$(	0	/	10	)$	\\
			sf2	&$	3	$&$[	0	;	8	]$&$	1066	$&$(	0	/	10	)$	\\
			jura	&$	3	$&$[	0.135	;	166.4	]$&$	359	$&$(	0	/	15	)$	\\
			osales	&$	12	$&$[	500	;	795000	]$&$	639	$&$(	0	/	413	)$	\\
			wq	&$	14	$&$[	0	;	5	]$&$	1060	$&$(	0	/	16	)$	\\
			oes97	&$	16	$&$[	30	;	48890	]$&$	334	$&$(	0	/	263	)$	\\
			oes10	&$	16	$&$[	30	;	64560	]$&$	403	$&$(	0	/	298	)$	\\
			\bottomrule														
		\end{tabular}														
	\end{table}
	
\clearpage

\section{Empirical results of non-sequential subgroup discovery algorithms}\label{appendix:non-sequential}

The comparison of SSD++ with subgroup set discovery algorithms that return sets (and not lists) can be seen in Table~\ref{table:results_nominal_nonseq}.

\begin{table*}[htb!]\centering															\begin{threeparttable}[b]
											
	\caption{Single nominal target results for non-sequential methods plus SSD++. This includes single-binary and single-nominal, respectively, separated by a horizontal line. The properties of the datasets can be seen in Table~\ref{table:data_nominal}, and are ordered by the number of target variables, number of classes, and number of samples, in this order. The evaluation measures are \{quality of the subgroup set swkl; the number of subgroups $|S|$; and the average number of conditions $|a|$\}. Note that FSSD does not work for the single-nominal case, and MCTS4DM only works for datasets with the same type of explanatory variables, thus the empty values $-$. }\label{table:results_nominal_nonseq}																								
	\ra{1.1}
    \setlength{\tabcolsep}{5.5pt}
	\begin{tabular}{@{}lrrrrrrrrrrrr@{}}\toprule																										
		&	\multicolumn{3}{r}{DSSD}					&	\multicolumn{3}{r}{MCTS4DM}					&	\multicolumn{3}{r}{FSSD}					&	\multicolumn{3}{r}{SSD++}					\\	\cmidrule(l){2-4} \cmidrule(l){5-7}\cmidrule(l){8-10}\cmidrule(l){11-13}
		datasets	&	swkl	&	$|S|\tnote{a}$	&	$|a|$	&\small	swkl	&	$|S|$	&	$|a|$	&	swkl	&	$|S|$	&	$|a|$	&	swkl	&	$|S|$	&	$|a|$	\\	\hline
		sonar 	&$	0.33	$&$	2	$&$	5	$&$	-	$&$	-	$&$	-	$&$	0.05	$&$	1	$&$	43	$&$\pmb{	0.43	}$&$	2	$&$	3	$\\	
		haberman 	&$\pmb{	0.08	}$&$	1	$&$	4	$&$\pmb{	0.08	}$&$	1	$&$	3	$&$	0.04	$&$	11	$&$	3	$&$	0.04	$&$	1	$&$	1	$\\	
		breastCancer	&$	0.79	$&$	6	$&$	3	$&$	0.81	$&$	6	$&$	4	$&$	0.35	$&$	6	$&$	9	$&$\pmb{	0.82	}$&$	6	$&$	2	$\\	
		australian 	&$	0.50	$&$	3	$&$	3	$&$	0.54	$&$	7	$&$	6	$&$	0.33	$&$	15	$&$	12	$&$\pmb{	0.55	}$&$	5	$&$	2	$\\	
		tictactoe 	&$	0.50	$&$	4	$&$	3	$&$	-	$&$	-	$&$	-	$&$	0.20	$&$	5	$&$	3	$&$\pmb{	0.87	}$&$	16	$&$	2	$\\	
		german 	&$\pmb{	0.15	}$&$	4	$&$	5	$&$	-	$&$	-	$&$	-	$&$	0.10	$&$	6	$&$	11	$&$	0.14	$&$	4	$&$	3	$\\	
		chess 	&$	0.76	$&$	11	$&$	4	$&$	-	$&$	-	$&$	-	$&$	0.34	$&$	4	$&$	15	$&$\pmb{	0.97	}$&$	17	$&$	2	$\\	
		mushrooms 	&$	0.97	$&$	3	$&$	4	$&$	-	$&$	-	$&$	-	$&$	0.40	$&$	5	$&$	20	$&$\pmb{	1.00	}$&$	12	$&$	1	$\\	
		magic 	&$	0.30	$&$	40	$&$	3	$&$	-	$&$	-	$&$	-	$&$	0.06	$&$	3	$&$	10	$&$\pmb{	0.47	}$&$	69	$&$	4	$\\	
		adult	&$	0.24	$&$	31	$&$	5	$&$	-	$&$	-	$&$	-	$&$	0.00	$&$	1	$&$	10	$&$\pmb{	0.31	}$&$	103	$&$	4	$\\[0.1cm]	
		avg. rank	&$	1.8	$&$	1.7	$&$	2.0	$&$	-	$&$	-	$&$	-	$&$	3.0	$&$	1.9	$&$	2.9	$&$\pmb{	1.2	}$&$	2.5	$&$	1.1	$\\ 	\midrule
		iris 	&$	1.44	$&$	3	$&$	2	$&$\pmb{	1.45	}$&$	4	$&$	3	$&$	-	$&$	-	$&$	-	$&$	1.44	$&$	4	$&$	1	$\\	
		balance	&$	0.63	$&$	9	$&$	3	$&$	-	$&$	-	$&$	-	$&$	-	$&$	-	$&$	-	$&$\pmb{	0.69	}$&$	9	$&$	3	$\\	
		CMC 	&$	0.18	$&$	7	$&$	3	$&$	0.16	$&$	20	$&$	4	$&$	-	$&$	-	$&$	-	$&$\pmb{	0.25	}$&$	7	$&$	2	$\\	
		page-blocks 	&$	0.36	$&$	19	$&$	3	$&$	-	$&$	-	$&$	-	$&$	-	$&$	-	$&$	-	$&$\pmb{	0.49	}$&$	21	$&$	3	$\\	
		nursery 	&$	0.92	$&$	2	$&$	3	$&$	-	$&$	-	$&$	-	$&$	-	$&$	-	$&$	-	$&$\pmb{	1.63	}$&$	81	$&$	3	$\\	
		automobile 	&$	0.85	$&$	5	$&$	5	$&$	-	$&$	-	$&$	-	$&$	-	$&$	-	$&$	-	$&$\pmb{	1.25	}$&$	5	$&$	2	$\\	
		glass 	&$	1.55	$&$	3	$&$	1	$&$	1.12	$&$	5	$&$	6	$&$	-	$&$	-	$&$	-	$&$\pmb{	1.92	}$&$	5	$&$	1	$\\	
		dermatology 	&$	1.85	$&$	6	$&$	3	$&$	1.02	$&$	9	$&$	6	$&$	-	$&$	-	$&$	-	$&$\pmb{	2.11	}$&$	9	$&$	2	$\\	
		kr-vs-k 	&$	0.62	$&$	13	$&$	3	$&$	-	$&$	-	$&$	-	$&$	-	$&$	-	$&$	-	$&$\pmb{	1.83	}$&$	351	$&$	3	$\\	
		abalone 	&$	0.53	$&$	14	$&$	3	$&$	-	$&$	-	$&$	-	$&$	-	$&$	-	$&$	-	$&$\pmb{	0.74	}$&$	16	$&$	2	$\\[0.1cm]	
		avg. rank	&$	1.9	$&$	1.2	$&$	1.7	$&$		$&$	-	$&$	-	$&$	-	$&$	-	$&$	-	$&$\pmb{	1.1	}$&$	1.9	$&$	1.3	$\\ 	
		\bottomrule																										
	\end{tabular}
	\begin{tablenotes}
    \item [a] As DSSD does have a stopping criterion, the maximum number of subgroups was selected as the number of subgroups found by SSD++, and total overlapping subgroups were posteriorly removed.
    \end{tablenotes}
	\end{threeparttable}
\end{table*}																										

\pagebreak									

\section{Statistical robustness results of applying SSD++ and our MDL approach on unseen data.}\label{appendix:statistical_robustness}
In Table~\ref{table:results_generalisation_nominal} we show the statistical robustness analysis of the SSD++ algorithm and our MDL approach by seeing how its training performance translates to unseen data performance. The measure used to evaluate the quality and generalisation of the found subgroups is $| LogLossRatio(train) -LogLossRatio(test)|$, i.e., the absolute difference between the log loss ratio in the train and test sets. First, the log loss (LogLoss) is defined as follows:
\begin{equation}\label{eq:logloss}
    LogLoss(\Ymat| \Xmat, SL(D_{\mathrm{train}})) = \sum_{(y,\x) \in \{\Ymat,\Xmat\}} -\log \Pr (y \given \x, SL(D_{\mathrm{train}})),
\end{equation}
where $SL(D_{\mathrm{train}})$ is a subgroup list selected from the train dataset $D_{\mathrm{train}}$, $\Ymat$ and $\Xmat$ correspond to the target and exploratory data for which we want to know the log loss (it can be the same as $D_{\mathrm{train}}$ for the training set, or different in case of the test set), and $\Pr (y_i \given SL)$ denotes the probabilities based on a categorical or a normal distribution for nominal or numeric targets, respectively. For the numeric case we use the probability density function instead.
Now, the LogLossRatio is the ratio between the LogLoss of subgroup list SL and the LogLoss of the marginal distribution (equal to the default rule of SL):
\begin{equation}\label{eq:loglossratio}
  LogLossRatio(*) = \frac{LogLoss(\Ymat| \Xmat, SL(D_{\mathrm{train}}))}{LogLoss( \Ymat| \Td)},
\end{equation}
where $\Td$ is the dataset's marginal distribution. To avoid having probabilities equal to zero and infinite log losses, for nominal targets, we added a pseudo-count of $0.5$ to every subgroup distribution, i.e., the Jeffrey's prior for the multinomial distribution \citep{grunwald2007minimum}.

\begin{table*}[!ht]\centering																										
\caption{Statistical robustness analysis for nominal target datasets. This table shows how our approach---SSD++ with normalised and absolute gain (Eq.~\eqref{eq:gain_kld}), i.e., $MDL_{\beta= 1}$ and $MDL_{\beta= 0}$---generalises to unseen data, tested in a $50\%$--$50\%$ train--test split. As baselines, we run SSD++ with KL and WKL divergence as quality measures, i.e., the same as our approach but without accounting for multiple hypothesis testing or distribution complexity (without $L(M)$ and $\texttt{COMP}(n_a) $ in Eq.~\eqref{eq:gain_kld}). The properties of the datasets can be seen in Table~\ref{table:data_nominal}, and are ordered in ascending number of: 1) number of classes; and 2) number of samples. The evaluation measures are \{Log Loss ratio between subgroup list and dataset marginal distribution (equal to dataset rule) for train ($LL_{\mathrm{train}}$) and test  ($LL_{\mathrm{test}}$) sets; and number of subgroups $|S|$\}. }\label{table:results_generalisation_nominal}																										
\setlength{\tabcolsep}{4.0pt} \ra{1.1} \begin{tabular}{@{}lrrrrrrrrrrrr@{}}\toprule																										
	&	\multicolumn{3}{r}{$\mathrm{MDL}_{\beta= 1}$ }					&	\multicolumn{3}{r}{$\mathrm{MDL}_{\beta= 0}$}					&	\multicolumn{3}{r}{$\mathrm{KL}_{\mathrm{Cat}}$}					&	\multicolumn{3}{r}{$\mathrm{WKL}_{\mathrm{Cat}}$}					\\	\cmidrule(l){2-4} \cmidrule(l){5-7}\cmidrule(l){8-10}\cmidrule(l){11-13}
datasets	&	\small $LL_{\mathrm{tr}}$	&	\small $LL_{\mathrm{tt}}$	&	$|S|$	&	\small $LL_{\mathrm{tr}}$	&	\small $LL_{\mathrm{tt}}$	&	$|S|$	&	\small $LL_{\mathrm{tr}}$	&	\small $LL_{\mathrm{tt}}$	&	$|S|$	&	\small $LL_{\mathrm{tr}}$	&	\small $LL_{\mathrm{tt}}$	&	$|S|$	\\	\hline
sonar	&$	0.57	$&$\pmb{	0.87	}$&$	2	$&$	0.57	$&$\pmb{	0.87	}$&$	2	$&$	0.12	$&$	1.43	$&$	18	$&$	0.04	$&$	1.25	$&$	5	$\\	
haberman	&$	1.00	$&$\pmb{	1.00	}$&$	0	$&$	1.00	$&$\pmb{	1.00	}$&$	0	$&$	0.51	$&$	1.34	$&$	27	$&$	0.55	$&$	1.38	$&$	15	$\\	
breastCancer	&$	0.13	$&$	0.31	$&$	4	$&$	0.17	$&$	0.40	$&$	3	$&$	0.07	$&$\pmb{	0.24	}$&$	20	$&$	0.15	$&$	0.38	$&$	4	$\\	
australian	&$	0.44	$&$	0.61	$&$	4	$&$	0.50	$&$\pmb{	0.54	}$&$	2	$&$	0.11	$&$	0.97	$&$	50	$&$	0.24	$&$	0.77	$&$	14	$\\	
tictactoe	&$	0.09	$&$\pmb{	0.16	}$&$	15	$&$	0.45	$&$	0.55	$&$	9	$&$	0.08	$&$	0.44	$&$	41	$&$	0.28	$&$	0.37	$&$	12	$\\	
german	&$	0.85	$&$	0.94	$&$	2	$&$	0.86	$&$\pmb{	0.92	}$&$	2	$&$	0.26	$&$	1.27	$&$	74	$&$	0.26	$&$	1.65	$&$	27	$\\	
chess	&$	0.08	$&$\pmb{	0.10	}$&$	11	$&$	0.13	$&$	0.16	$&$	9	$&$	0.04	$&$	0.16	$&$	83	$&$	0.10	$&$	0.17	$&$	16	$\\	
mushrooms	&$	0.00	$&$\pmb{	0.00	}$&$	11	$&$	0.17	$&$	0.17	$&$	6	$&$	0.01	$&$	0.02	$&$	36	$&$	0.16	$&$	0.17	$&$	7	$\\	
magic	&$	0.51	$&$\pmb{	0.63	}$&$	39	$&$	0.66	$&$	0.69	$&$	9	$&$	0.13	$&$	0.88	$&$	1078	$&$	0.59	$&$	0.74	$&$	77	$\\	
adult	&$	0.62	$&$\pmb{	0.67	}$&$	65	$&$	0.72	$&$	0.72	$&$	8	$&$	0.34	$&$	0.88	$&$	2570	$&$	0.68	$&$	0.72	$&$	96	$\\	\midrule
iris	&$	0.18	$&$	0.07	$&$	3	$&$	0.16	$&$\pmb{	0.06	}$&$	3	$&$	0.16	$&$	0.18	$&$	9	$&$	0.16	$&$\pmb{	0.06	}$&$	3	$\\	
balance	&$	0.48	$&$	0.76	$&$	6	$&$	0.58	$&$	0.80	$&$	3	$&$	0.27	$&$	0.74	$&$	51	$&$	0.35	$&$\pmb{	0.74	}$&$	14	$\\	
CMC	&$	0.90	$&$	0.91	$&$	3	$&$	0.89	$&$\pmb{	0.88	}$&$	3	$&$	0.41	$&$	1.15	$&$	148	$&$	0.64	$&$	1.01	$&$	42	$\\	
page-blocks	&$	0.27	$&$\pmb{	0.42	}$&$	12	$&$	0.38	$&$	0.46	$&$	6	$&$	0.21	$&$	0.51	$&$	94	$&$	0.36	$&$	0.45	$&$	12	$\\	
nursery	&$	0.08	$&$\pmb{	0.11	}$&$	59	$&$	0.44	$&$	0.44	$&$	3	$&$	0.09	$&$	0.17	$&$	310	$&$	0.44	$&$	0.44	$&$	3	$\\	
automobile	&$	0.58	$&$	0.88	$&$	3	$&$	0.56	$&$\pmb{	0.86	}$&$	3	$&$	0.28	$&$	0.91	$&$	14	$&$	0.31	$&$	\pmb{0.86}	$&$	6	$\\	
glass	&$	0.26	$&$\pmb{	0.23	}$&$	4	$&$	0.38	$&$	0.31	$&$	3	$&$	0.20	$&$	0.39	$&$	14	$&$	0.36	$&$	0.33	$&$	3	$\\	
dermatology	&$	0.28	$&$\pmb{	0.37	}$&$	6	$&$	0.43	$&$	0.51	$&$	3	$&$	0.18	$&$	0.51	$&$	21	$&$	0.43	$&$	0.54	$&$	4	$\\	
kr-vs-k	&$	0.57	$&$\pmb{	0.62	}$&$	186	$&$	0.88	$&$	0.88	$&$	5	$&$	0.53	$&$	0.77	$&$	2864	$&$	0.88	$&$	0.88	$&$	5	$\\	
abalone	&$	0.85	$&$\pmb{	0.85	}$&$	7	$&$	0.88	$&$	0.87	$&$	3	$&$	0.78	$&$	1.05	$&$	331	$&$	0.87	$&$	0.87	$&$	7	$\\	
\bottomrule																										
\end{tabular}																										
\end{table*}

\begin{table*}[h!]\centering															\begin{threeparttable}[b]
											
\caption{Statistical robustness analysis for numeric target datasets. This table shows how our approach---SSD++ with normalised and absolute gain (Eq.~\eqref{eq:gain_kld}), i.e., $MDL_{\beta= 1}$ and $MDL_{\beta= 0}$---generalises to unseen data, tested in a $50\%$--$50\%$ train--test split. As baselines, we run SSD++ with KL and WKL divergence as quality measures, i.e., the same as our approach but without accounting for multiple hypothesis testing or distribution complexity (without $L(M)$ and $\texttt{COMP}(n_a) $ in Eq.~\eqref{eq:gain_kld}).  The properties of the datasets can be seen in Table~\ref{table:data_numeric}, and are ordered in ascending number of number of samples. The evaluation measures are \{Log Loss ratio between subgroup list and dataset marginal distribution (equal to dataset rule) for train ($LL_{\mathrm{train}}$) and test  ($LL_{\mathrm{test}}$) sets; and number of subgroups $|S|$\}. Note that $\infty$ values happen when at least one unseen point is not covered by the standard deviation of one subgroup.}\label{table:results_generalisation_numeric}																										
\setlength{\tabcolsep}{3.5pt} \ra{1.1} \begin{tabular}{@{}lrrrrrrrrrrrr@{}}\toprule																										
	&	\multicolumn{3}{r}{$\mathrm{MDL}_{\beta= 1}$ }					&	\multicolumn{3}{r}{$\mathrm{MDL}_{\beta= 0}$}					&	\multicolumn{3}{r}{$\mathrm{KL}_{\mathrm{Cat}}$}					&	\multicolumn{3}{r}{$\mathrm{WKL}_{\mathrm{Cat}}$}					\\	\cmidrule(l){2-4} \cmidrule(l){5-7}\cmidrule(l){8-10}\cmidrule(l){11-13}
datasets	&	\small $LL_{\mathrm{tr}}$	&	\small $LL_{\mathrm{tt}}$	&	$|S|$	&	\small $LL_{\mathrm{tr}}$	&	\small $LL_{\mathrm{tt}}$	&	$|S|$	&	\small $LL_{\mathrm{tr}}$	&	\small $LL_{\mathrm{tt}}$	&	$|S|$	&	\small $LL_{\mathrm{tr}}$	&	\small $LL_{\mathrm{tt}}$	&	$|S|$	\\	\hline
baseball	&$	0.86	$&$	0.97	$&$	4	$&$	0.88	$&$\pmb{	0.92	}$&$	3	$&$	0.66	$&$	163.57	$&$	42	$&$	0.84	$&$	1.40	$&$	7	$\\	
autoMPG8	&$	0.69	$&$\pmb{	0.79	}$&$	7	$&$	0.78	$&$	0.81	$&$	3	$&$	0.41	$&$	3.19	$&$	45	$&$	0.74	$&$	0.84	$&$	7	$\\	
dee	&$	0.32	$&$\pmb{	0.50	}$&$	6	$&$	0.52	$&$	0.67	$&$	2	$&$	-0.59\tnote{a}	$&$	13.39	$&$	44	$&$	0.21	$&$	7.59	$&$	10	$\\	
ele-1	&$	0.90	$&$	0.93	$&$	6	$&$	0.92	$&$	0.93	$&$	4	$&$	0.87	$&$\pmb{	0.92	}$&$	34	$&$	0.92	$&$	0.93	$&$	5	$\\	
forestFires	&$	0.45	$&$	196.32	$&$	13	$&$	0.60	$&$\pmb{	10.03	}$&$	6	$&$	0.25	$&$	157.70	$&$	52	$&$	0.55	$&$	26.29	$&$	11	$\\	
concrete	&$	0.82	$&$\pmb{	0.86	}$&$	10	$&$	0.87	$&$	0.88	$&$	6	$&$	0.51	$&$	6.59	$&$	123	$&$	0.84	$&$	0.94	$&$	11	$\\	
treasury	&$	0.05	$&$	0.86	$&$	18	$&$	0.41	$&$	0.48	$&$	6	$&$	-0.30\tnote{a}	$&$	10.90	$&$	99	$&$	0.36	$&$\pmb{	0.41	}$&$	6	$\\	
wizmir	&$	0.56	$&$\pmb{	0.58	}$&$	15	$&$	0.72	$&$	0.73	$&$	4	$&$	0.30	$&$	3.67	$&$	176	$&$	0.72	$&$	0.73	$&$	4	$\\	
abalone	&$	0.80	$&$\pmb{	0.87	}$&$	18	$&$	0.87	$&$	0.90	$&$	7	$&$	0.64	$&$	1.50	$&$	304	$&$	0.86	$&$	0.90	$&$	12	$\\	
puma32h	&$	0.68	$&$	0.71	$&$	28	$&$	0.76	$&$	0.76	$&$	8	$&$	2.49	$&$	\infty	$&$	1023	$&$	0.76	$&$\pmb{	0.40	}$&$	8	$\\	
ailerons	&$	0.87	$&$	\infty	$&$	59	$&$	0.92	$&$	0.92	$&$	5	$&$	1.39	$&$	\infty	$&$	1523	$&$	0.92	$&$\pmb{	0.72	}$&$	7	$\\	
elevators	&$	0.80	$&$	0.86	$&$	93	$&$	0.87	$&$	0.86	$&$	11	$&$	1.62	$&$	\infty	$&$	1951	$&$	0.86	$&$\pmb{	0.62	}$&$	38	$\\	
bikesharing	&$	0.82	$&$\pmb{	0.85	}$&$	84	$&$	0.90	$&$	0.90	$&$	6	$&$	0.68	$&$	3.33	$&$	1997	$&$	0.90	$&$	0.94	$&$	21	$\\	
california	&$	0.94	$&$	4.14	$&$	99	$&$	0.97	$&$\pmb{	0.97	}$&$	9	$&$	0.88	$&$	2.44	$&$	2124	$&$	0.97	$&$	\infty	$&$	37	$\\	
house	&$	0.88	$&$	\infty	$&$	151	$&$	0.94	$&$	2.57	$&$	18	$&$	0.76	$&$	\infty	$&$	2744	$&$	0.93	$&$\pmb{	1.31	}$&$	36	$\\	
\bottomrule																										
\end{tabular}
\begin{tablenotes}
\item [a]  Negative values are possible when the standard deviation is very small. 
\end{tablenotes}
\end{threeparttable}
							
\end{table*}
\clearpage

\section{Empirical analysis of compression gain}\label{appendix:empiricalabsvsnorm}

\begin{figure}[!htb]
	\centering
	\begin{subfigure}[b]{0.48\textwidth}
		\centering
		\includegraphics[width=\textwidth]{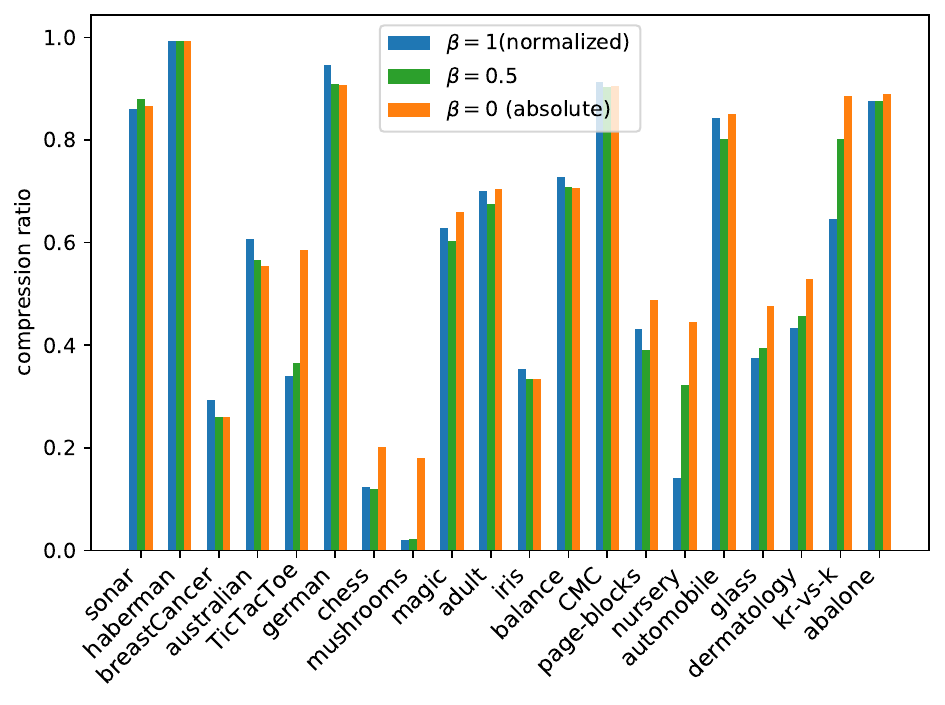}
		\caption{Univariate nominal target}
		\label{fig:beta_compression_nominal}
	\end{subfigure}
	\hfill
	\begin{subfigure}[b]{0.48\textwidth}
		\centering
		\includegraphics[width=\textwidth]{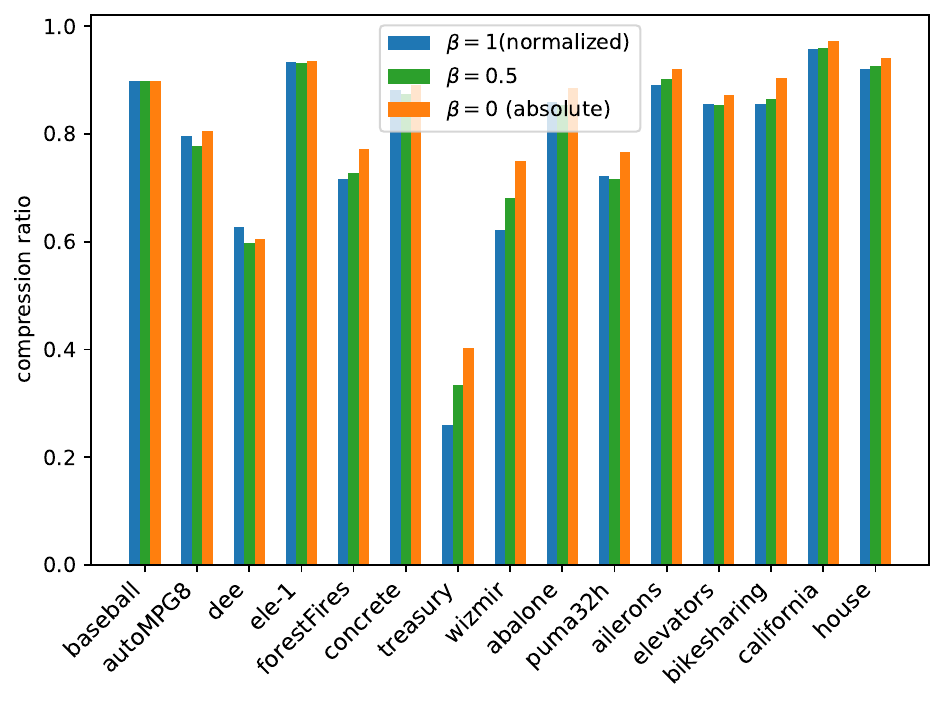}
		\caption{Univariate numeric target}
		\label{fig:beta_compression_numeric}
	\end{subfigure}
	\caption{Compression ratio obtained with $\beta = 0$ (absolute gain), $\beta = 0.5$, and $\beta = 1$ (normalised gain)}
	\label{fig:beta_compression}
\end{figure}

\begin{figure}[!htb]
	\centering
	\begin{subfigure}[b]{0.48\textwidth}
		\centering
		\includegraphics[width=\textwidth]{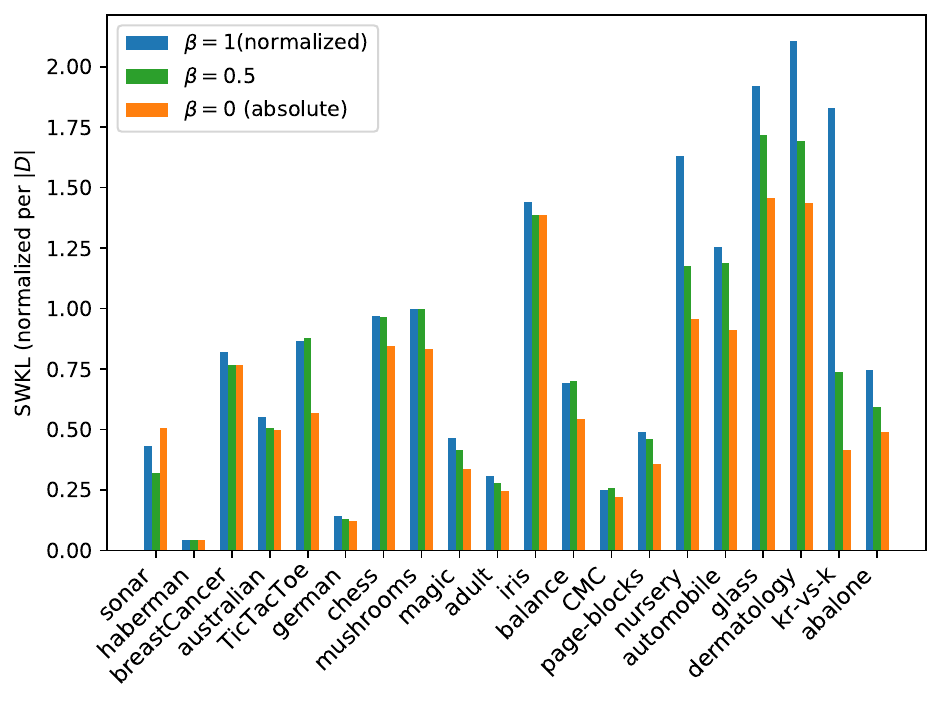}
		\caption{Univariate nominal target}
		\label{fig:beta_SWKL_nominal}
	\end{subfigure}
	\hfill
	\begin{subfigure}[b]{0.48\textwidth}
		\centering
		\includegraphics[width=\textwidth]{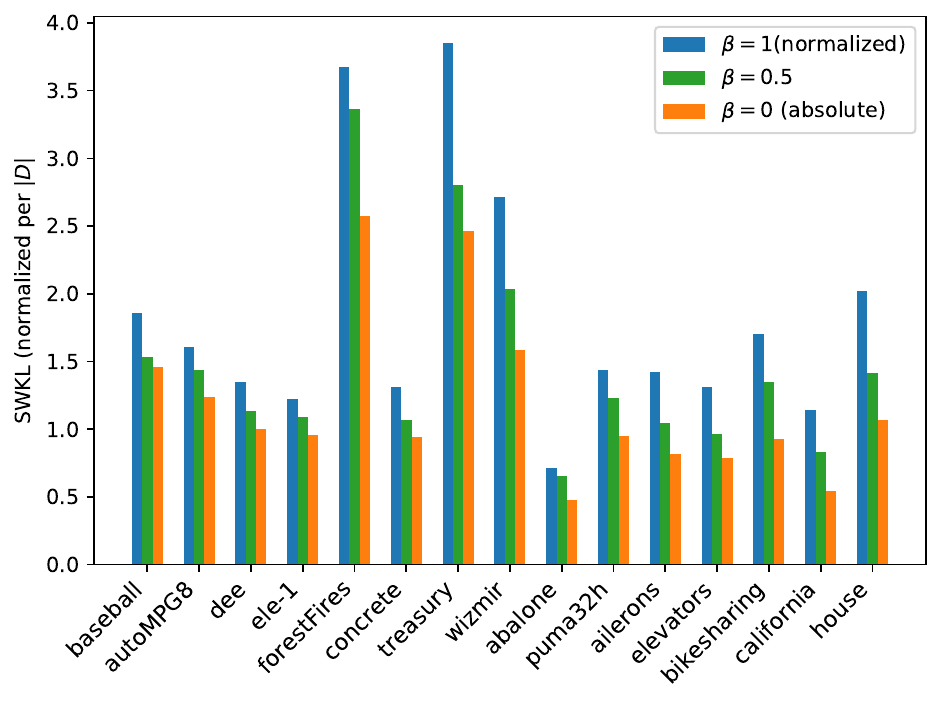}
		\caption{Univariate numeric target}
		\label{fig:beta_SWKL_numeric}
	\end{subfigure}
	\caption{Normalised SWKL obtained with $\beta = 0$ (absolute gain), $\beta = 0.5$, and $\beta = 1$ (normalised gain)}
	\label{fig:beta_SWKL}
\end{figure}

\begin{figure}[!htb]
	\centering
	\begin{subfigure}[b]{0.48\textwidth}
		\centering
		\includegraphics[width=\textwidth]{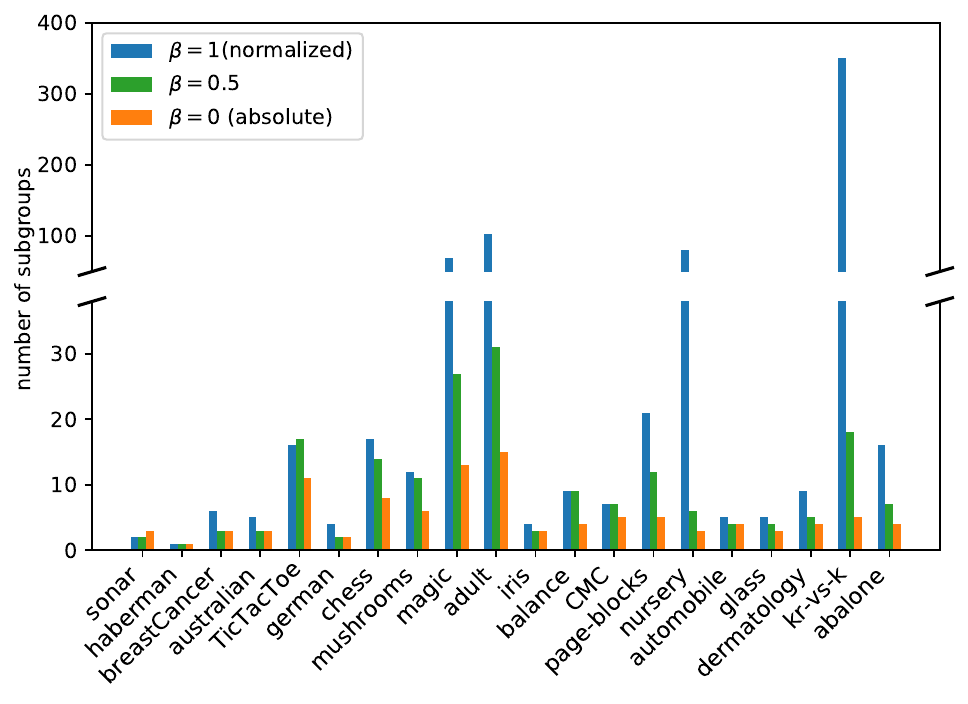}
		\caption{Univariate nominal target}
		\label{fig:beta_rules_nominal}
	\end{subfigure}
	\hfill
	\begin{subfigure}[b]{0.48\textwidth}
		\centering
		\includegraphics[width=\textwidth]{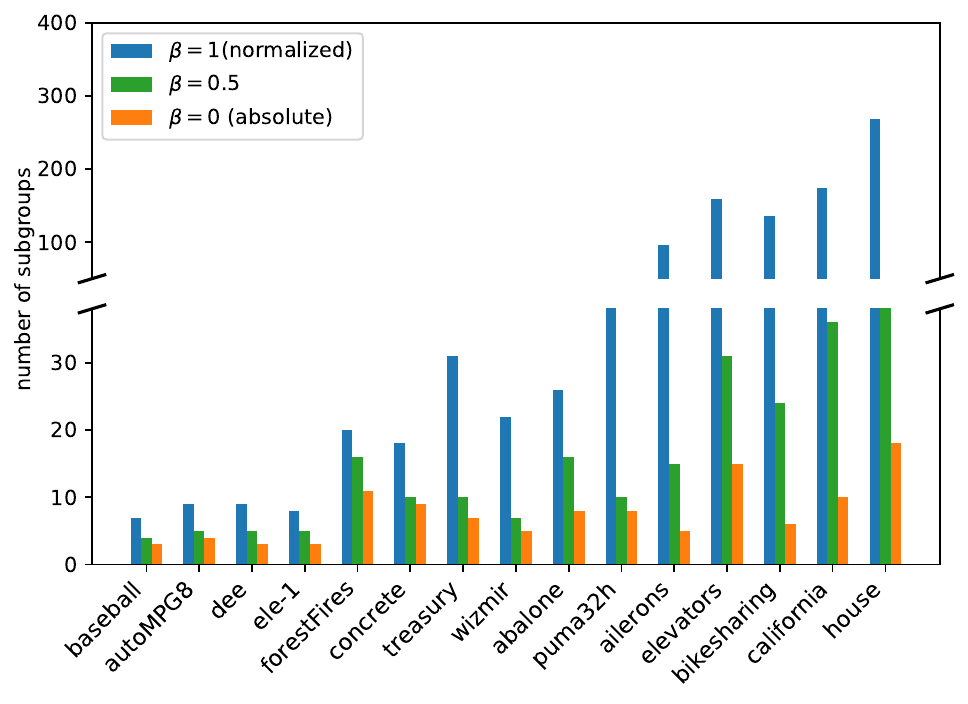}
		\caption{Univariate numeric target}
		\label{fig:beta_rules_numeric}
	\end{subfigure}
	\caption{Number subgroups obtained with $\beta = 0$ (absolute gain), $\beta = 0.5$, and $\beta = 1$ (normalised gain)}
	\label{fig:beta_rules}
\end{figure}

In this section we present a thorough comparison of the normalisation terms $\beta$ of SSD++, where  $\beta= 1$ is the \emph{normalised} gain and $\beta= 0$ the \emph{absolute} gain. SSD++ is executed with the same parameters (beam width, number of cut points for numerical variables, and maximum depth of search) as in the experiments section, i.e., $w_b =100$, $n_{cut} =5$, $d_{max} = 5$. The different types of gain are compared for all the benchmark datasets described in the paper in terms of their compression ratio (defined later) in Figure~\ref{fig:beta_compression}, Sum of Weighted Kullback-Leibler divergency (SWKL) in Figure~\ref{fig:beta_SWKL}, and number of rules in Figure~\ref{fig:beta_rules}. The compression ratio is the length of the found model $L(D, M)$ divided by the length of encoding the data with the dataset distribution (a model without subgroups) $L(D \given \Td)$, and formally it has the following form:
\begin{equation}
L \% = \frac{L(D, M)}{L(D \given \Td)}
\end{equation}

\pagebreak
				
\section{Empirical analysis of the influence of the beam search hyperparameters}\label{appendix:empiricalbeamsearch}
This section presents a thorough comparison of the influence of the hyperparameters of the beam search of SSD++ on its results. As a complete search over the whole combination of parameters is unfeasible, we present here an exploration of the hyperparameters used for the experimental comparison in the paper ($w_b =100$, $n_{cut} =5$, $d_{max} = 5$), i.e., we fix two of the parameters on the values above and then proceed to change the selected parameter of interest. We do this for all the $3$ parameters. The line between the dots of the same colour does not represent an interpolation and is merely used to aid visualisation and suggest trends. 

\emph{Note on relative compression.} It may seem that the values of the relative compression remain constant, but that is an illusion due to the scale of the $y$ axis. Moreover, as the compression ratio is given by dividing large values (usually above the thousands), its value with two decimal digits can be misleading. Nonetheless, in general, when zooming over the figures, one can discern a slight improvement (smaller values) for larger values of the hyperparameters.
	
\begin{figure}[!htb]
	\centering
	\begin{subfigure}[b]{0.48\textwidth}
		\centering
		\includegraphics[width=\textwidth]{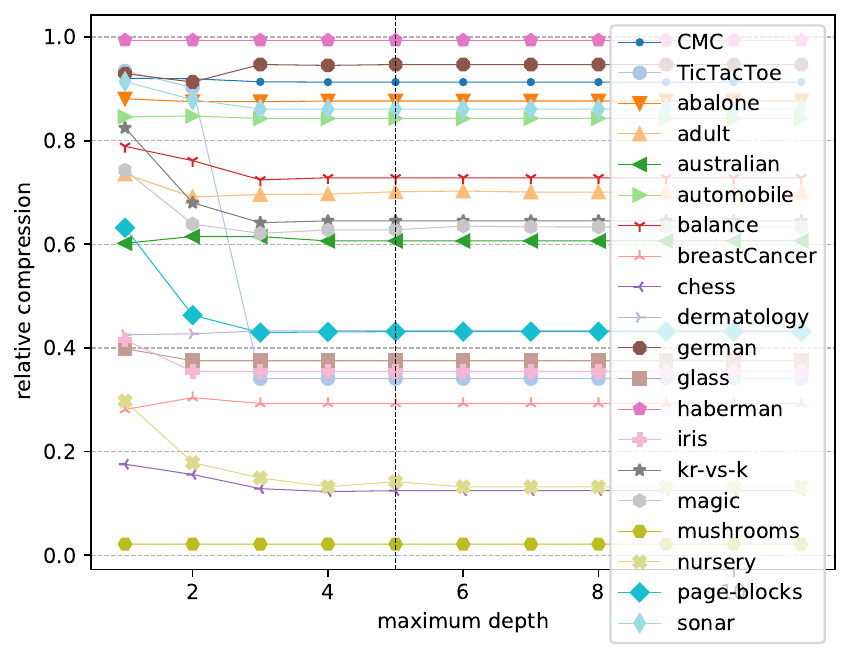}
		\caption{Univariate nominal target}
		\label{fig:maxdepth_vs_compression_nominal}
	\end{subfigure}
	\hfill
	\begin{subfigure}[b]{0.48\textwidth}
		\centering
		\includegraphics[width=\textwidth]{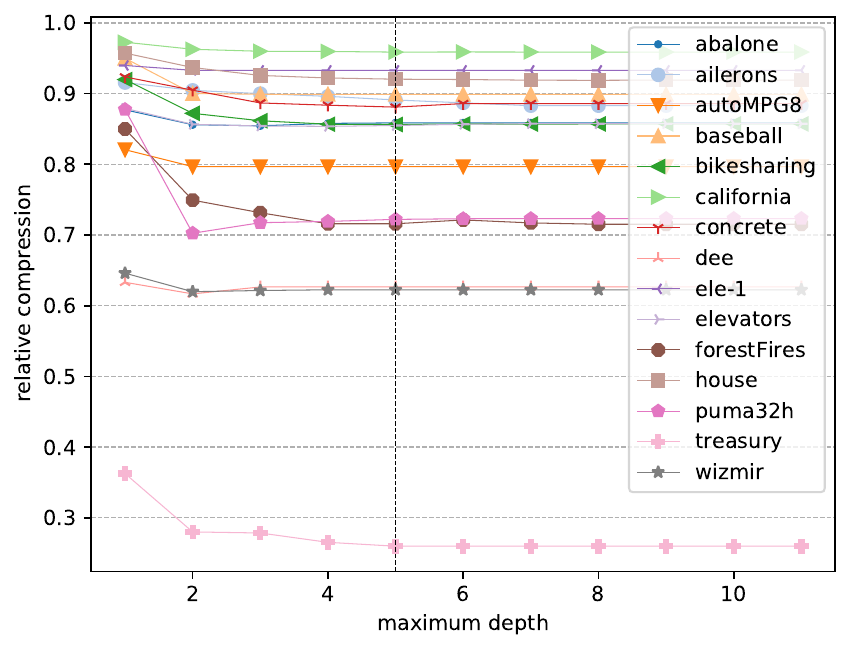}
		\caption{Univariate numeric target}
		\label{fig:maxdepth_vs_compression_numeric}
	\end{subfigure}
	\caption{Compression ratio obtained by varying the maximum search depth fixing  $w_b =100$, $n_{cut} =5$ and $\beta =1$ (normalised gain). The black vertical line represents the value used in Experiments section of the paper}
	\label{fig:maxdepth_vs_compression}
\end{figure}

\begin{figure}[!htb]
	\centering
	\begin{subfigure}[b]{0.48\textwidth}
		\centering
		\includegraphics[width=\textwidth]{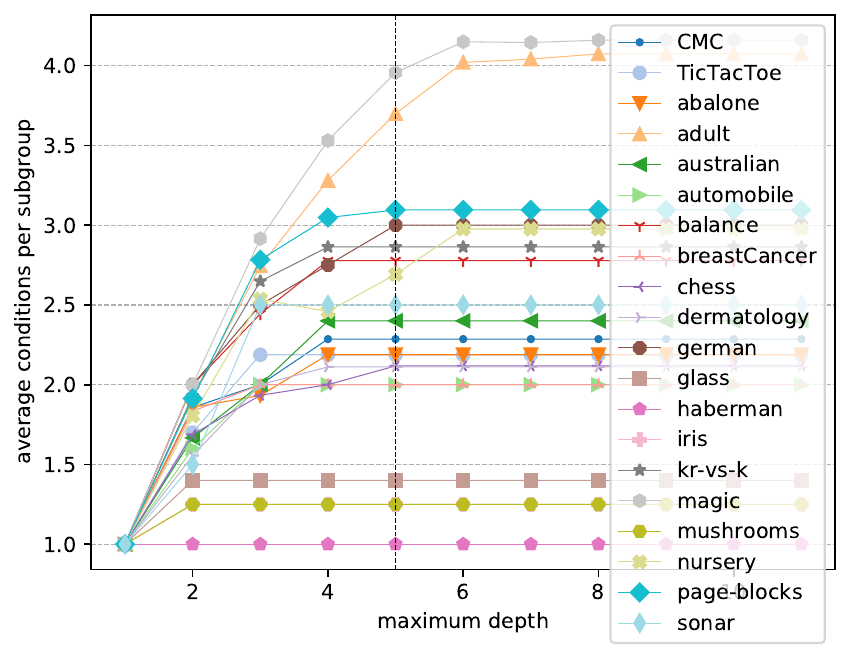}
		\caption{Univariate nominal target}
		\label{fig:maxdepth_vs_conditions_nominal}
	\end{subfigure}
	\hfill
	\begin{subfigure}[b]{0.48\textwidth}
		\centering
		\includegraphics[width=\textwidth]{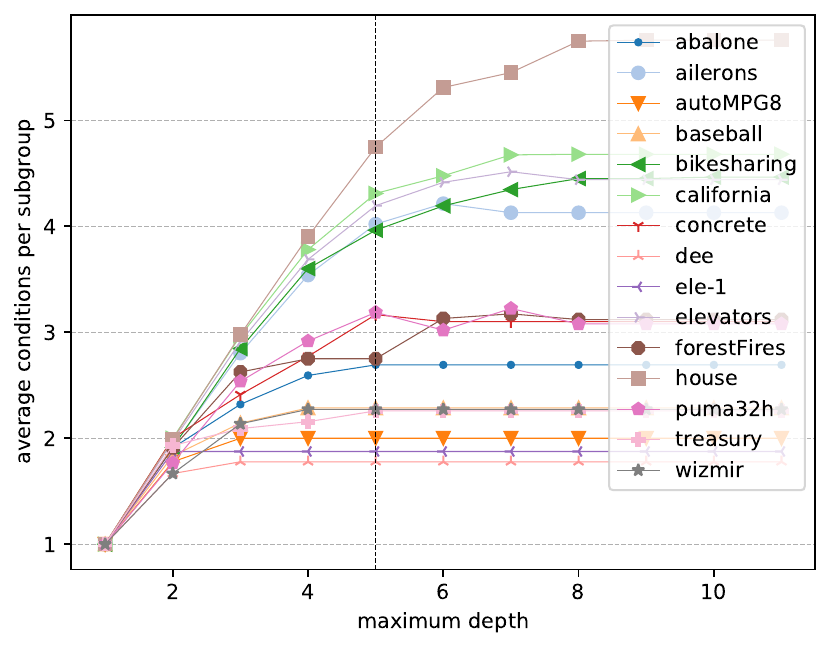}
		\caption{Univariate numeric target}
		\label{fig:maxdepth_vs_conditions_numeric}
	\end{subfigure}
	\caption{Average number of conditions per subgroup obtained by varying the maximum search depth fixing $w_b =100$, $n_{cut} =5$ and $\beta =1$ (normalised gain). The black vertical line represents the value used in Experiments section of the paper}
	\label{fig:maxdepth_vs_conditions}
\end{figure}

\begin{figure}[!htb]
	\centering
	\begin{subfigure}[b]{0.48\textwidth}
		\centering
		\includegraphics[width=\textwidth]{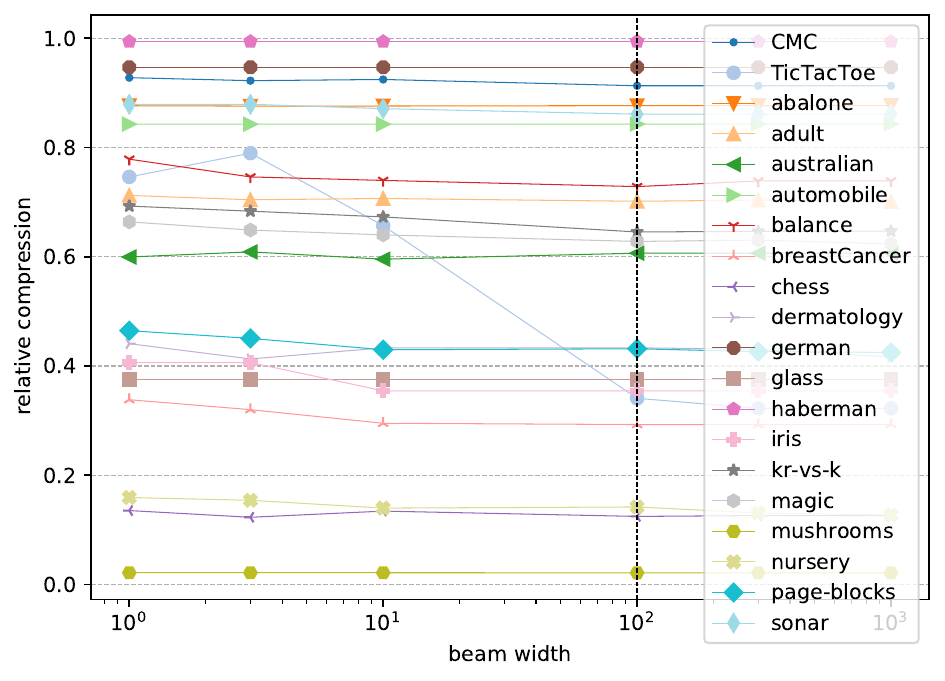}
		\caption{Univariate nominal target}
		\label{fig:beam_compression_nominal}
	\end{subfigure}
	\hfill
	\begin{subfigure}[b]{0.48\textwidth}
	\centering
	\includegraphics[width=\textwidth]{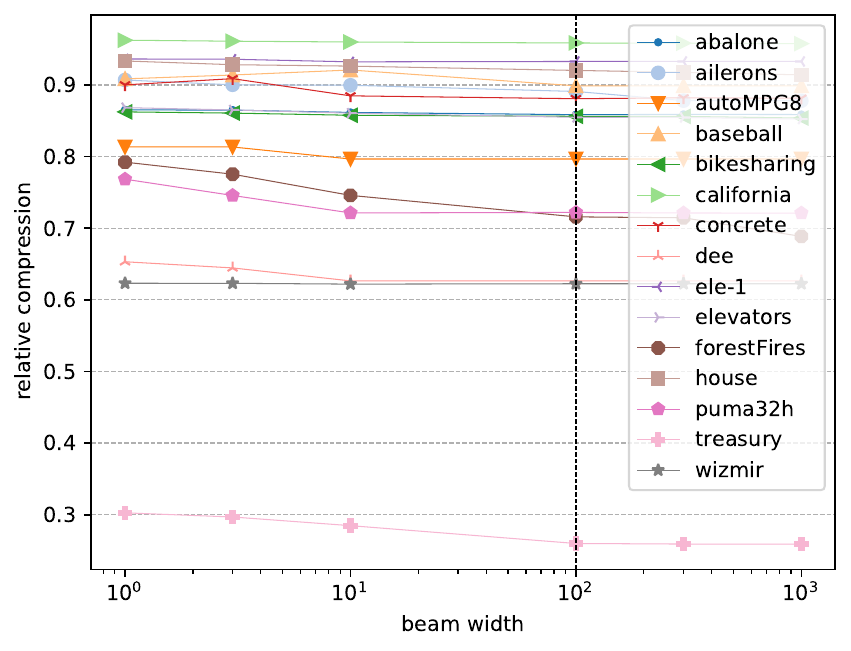}
	\caption{Univariate numeric target}
	\label{fig:beam_compression_numeric}
\end{subfigure}
\caption{Compression ratio obtained by varying the beam width and fixing  $d_{max} =5$, $n_{cut} =5$ and $\beta =1$ (normalised gain). The black vertical line represents the value used in Experiments section of the paper}
\label{fig:beam_compression}
\end{figure}

\begin{figure}[!htb]
	\centering
	\begin{subfigure}[b]{0.48\textwidth}
		\centering
		\includegraphics[width=\textwidth]{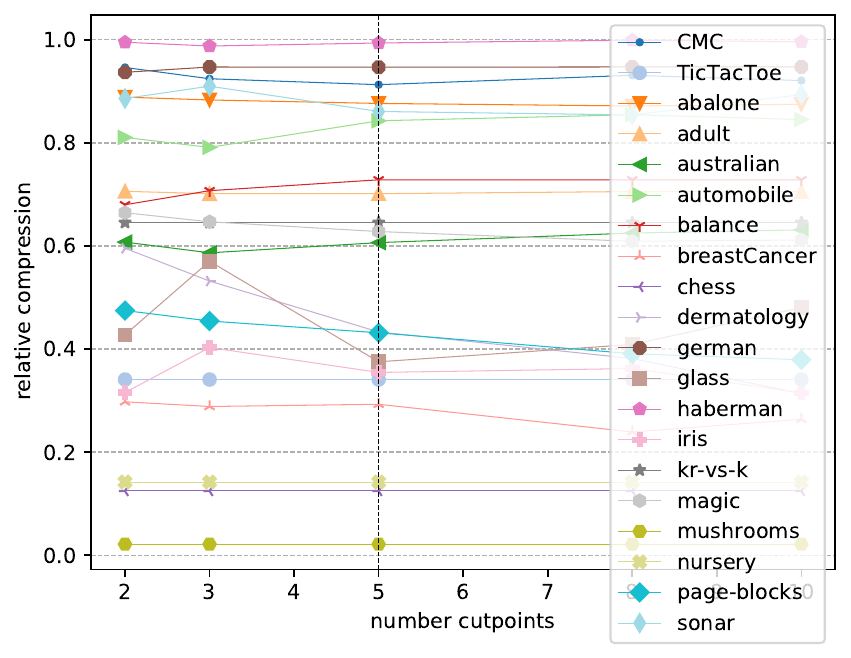}
		\caption{Univariate nominal target}
		\label{fig:ncutpoints_compression_nominal}
	\end{subfigure}
	\hfill
	\begin{subfigure}[b]{0.48\textwidth}
		\centering
		\includegraphics[width=\textwidth]{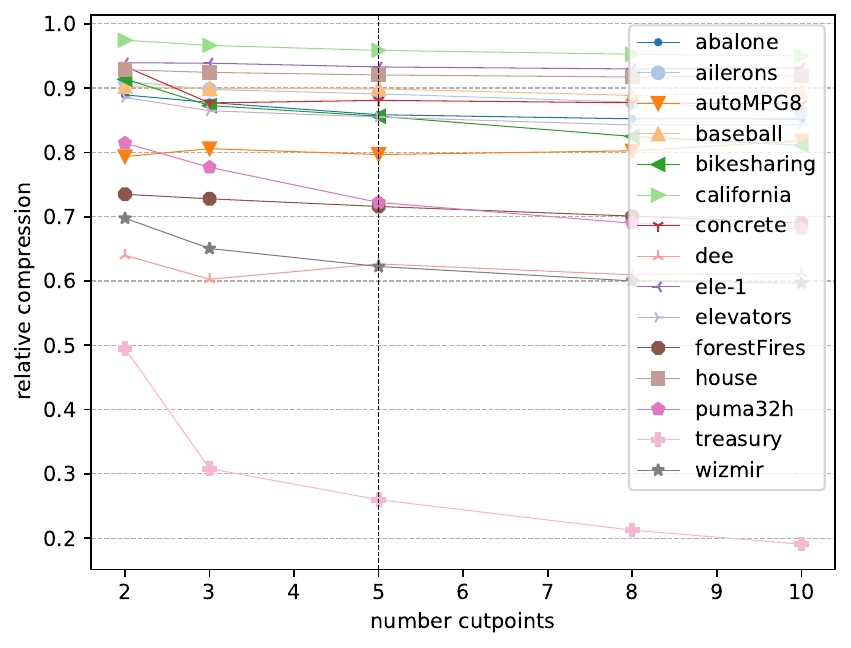}
		\caption{Univariate numeric target}
		\label{fig:ncutpoints_compression_numeric}
	\end{subfigure}
	\caption{Compression ratio obtained by varying the number of cut points and fixing $w_b =100$, $d_{max} =5$ and $\beta =1$ (normalised gain). The black vertical line represents the value used in Experiments section of the paper}
	\label{fig:ncutpoints_compression}
\end{figure}	
\clearpage
	
\end{appendix}

\bibliographystyle{spbasic}      

\end{document}